\documentclass[]{dataflow}

\usepackage[toc,page,header]{appendix}

\usepackage{algorithm}
\usepackage{algorithmic}
\usepackage{microtype}
\usepackage{subcaption}
\usepackage{hyperref}
\usepackage[utf8]{inputenc}
\usepackage[table]{xcolor} 
\usepackage{amsmath}
\usepackage{amssymb}
\usepackage{mathtools}
\usepackage{amsthm}
\usepackage{pifont}
\usepackage[most]{tcolorbox}
\usepackage{etoolbox}
\usepackage{enumitem}
\usepackage{cuted}
\definecolor{verifier_bg}{RGB}{240, 248, 255} % 浅蓝背景，用于 Verifier
\definecolor{policy_bg}{RGB}{255, 255, 255}   % 白色背景，用于 Policy
\definecolor{guide_text}{RGB}{0, 50, 150}     % 深蓝色文字，强调 Guidance
\definecolor{softgreen}{RGB}{34,139,34}
\usepackage[HTML,svgnames]{xcolor}
\usepackage{tcolorbox}
\tcbuselibrary{skins,breakable}

\theoremstyle{plain}

\theoremstyle{definition}

\theoremstyle{remark}

\usepackage{tabularx}

\usepackage{minitoc}
\usepackage{multirow}     % 支持 \multirow
\usepackage{multicol}     % 支持多列排版（部分场景）
\usepackage{booktabs}     % \toprule \midrule \bottomrule
\usepackage{colortbl}     % 单元格或行着色
\usepackage[table]{xcolor} % 颜色支持（含 table 选项）
\usepackage{makecell}     % 允许在表格中换行 / 对齐
\usepackage{graphicx}     % 支持 \resizebox
\usepackage{minted}
\usepackage{fontawesome5}

\usepackage{array}
\definecolor{graybg}{rgb}{0.96, 0.96, 0.96}
\definecolor{bggray}{RGB}{242, 242, 242}
\definecolor{bgblue}{RGB}{220, 235, 255}
\usepackage{float}
\usepackage{wasysym}
\definecolor{aliceblue}{rgb}{0.94, 0.97, 1.0}
\definecolor{lavendermist}{rgb}{0.96, 0.96, 0.99}
\definecolor{lavenderpurple}{rgb}{0.59, 0.48, 0.71}
\definecolor{honeydew}{rgb}{0.98, 1.0, 0.98}
\definecolor{asparagus}{rgb}{0.53, 0.66, 0.42}
\definecolor{conferenceblue}{RGB}{0, 100, 180}  % NeurIPS蓝
\definecolor{conferencered}{RGB}{200, 0, 50}
\definecolor{critique_bg}{RGB}{255, 235, 238}  
\definecolor{critique_frame}{RGB}{211, 47, 47}  
\definecolor{correction_bg}{RGB}{232, 245, 233}  
\definecolor{header_gray}{RGB}{245, 245, 245}

\newtcolorbox{promptbox}[1][]{
  enhanced,
  breakable,
  colback=aliceblue,
  colframe=SteelBlue,
  toptitle=1mm,   
  bottomtitle=1mm,
  boxrule=0.5pt,
  arc=2mm,
  left=8pt,right=8pt,top=7pt,bottom=7pt,
  title=\textbf{SYSTEM PROMPT},
  fonttitle=\normalsize,
  coltitle=white,
  #1
}

\newtcolorbox{critiquepromptbox}[1][]{
  enhanced,
  breakable,
  colback=lavendermist,
  colframe=lavenderpurple,
  toptitle=1mm,   
  bottomtitle=1mm,
  boxrule=0.5pt,
  arc=2mm,
  left=8pt,right=8pt,top=7pt,bottom=7pt,
  title=\textbf{CRITIQUE PROMPT},
  fonttitle=\normalsize,
  coltitle=white,
  #1
}

\newtcolorbox{userpromptbox}[1][]{
  enhanced,
  breakable,
  colback=honeydew,
  colframe=asparagus,
  toptitle=1mm,   
  bottomtitle=1mm,
  boxrule=0.5pt,
  arc=2mm,
  left=8pt,right=8pt,top=7pt,bottom=7pt,
  title=\textbf{USER PROMPT},
  fonttitle=\normalsize,
  coltitle=white,
  #1
}

\newtcolorbox{critiquesystempromptbox}[1][]{
  enhanced,
  breakable,
  colback=lavendermist,
  colframe=lavenderpurple,
  toptitle=1mm,   
  bottomtitle=1mm,
  boxrule=0.5pt,
  arc=2mm,
  left=8pt,right=8pt,top=7pt,bottom=7pt,
  title=\textbf{CRITIQUE SYSTEM PROMPT},
  fonttitle=\normalsize,
  coltitle=white,
  #1
}

\newtcolorbox{critiquesystempromptboxwoimg}[1][]{
  enhanced,
  breakable,
  colback=lavendermist,
  colframe=lavenderpurple,
  toptitle=1mm,   
  bottomtitle=1mm,
  boxrule=0.5pt,
  arc=2mm,
  left=8pt,right=8pt,top=7pt,bottom=7pt,
  title=\textbf{CRITIQUE SYSTEM PROMPT without IMAGE},
  fonttitle=\normalsize,
  coltitle=white,
  #1
}

\newtcolorbox{critiqueuserpromptbox}[1][]{
  enhanced,
  breakable,
  colback=honeydew,
  colframe=asparagus,
  toptitle=1mm,   
  bottomtitle=1mm,
  boxrule=0.5pt,
  arc=2mm,
  left=8pt,right=8pt,top=7pt,bottom=7pt,
  title=\textbf{CRITIQUE USER PROMPT},
  fonttitle=\normalsize,
  coltitle=white,
  #1
}

\title{Canvas-of-Thought: Grounding Reasoning via Mutable Structured States}

\author[*]{Lingzhuang Sun$^{1}$}
\author[*]{Yuxia Zhu$^{2}$}
\author[*]{Ruitong Liu$^{2}$}
\author[]{Hao Liang$^{2}$}
\author[]{Zheng Sun$^{1}$}
\author[]{Caijun Jia$^{1}$}
\author[]{Honghao He$^{1}$}
\author[]{Yuchen Wu$^{3}$}
\author[]{Siyuan Li$^{4}$}
\author[]{Jingxuan Wei$^{1}$}
\author[]{Xiangxiang Zhang$^{1}$}
\author[]{Bihui Yu$^{1}$}
\author[]{Wentao Zhang$^{2}$}

\affiliation[]{$^{1}$University of Chinese Academy of Sciences $^{2}$Peking University\\ $^{3}$New York University  $^{4}$Westlake University}

\abstract{
While Chain-of-Thought (CoT) prompting has significantly advanced the reasoning capabilities of Multimodal Large Language Models (MLLMs), relying solely on linear text sequences remains a bottleneck for complex tasks. We observe that even when auxiliary visual elements are interleaved, they are often treated as static snapshots within a one-dimensional, unstructured reasoning chain. We argue that such approaches treat reasoning history as an immutable stream: correcting a local error necessitates either generating verbose downstream corrections or regenerating the entire context. This forces the model to implicitly maintain and track state updates, significantly increasing token consumption and cognitive load. This limitation is particularly acute in high-dimensional domains, such as geometry and SVG design, where the textual expression of CoT lacks explicit visual guidance, further constraining the model's reasoning precision.
To bridge this gap, we introduce \textbf{Canvas-of-Thought (Canvas-CoT)}. By leveraging a HTML Canvas as an external reasoning substrate, Canvas-CoT empowers the model to perform atomic, DOM-based CRUD operations. This architecture enables in-place state revisions without disrupting the surrounding context, allowing the model to explicitly maintain the "ground truth". Furthermore, we integrate a rendering-based critique loop that serves as a hard constraint validator, providing explicit visual feedback to resolve complex tasks that are difficult to articulate through text alone. Extensive experiments on VCode, RBench-V, and MathVista demonstrate that Canvas-CoT significantly outperforms existing baselines, establishing a new paradigm for context-efficient multimodal reasoning.
}

\date{\today}
\def\emailicon{\raisebox{-1.5pt}{\includegraphics[height=1.05em]{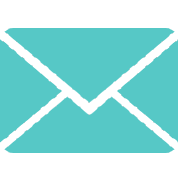}}}
\def\githubicon{\raisebox{-1.5pt}{\includegraphics[height=1.05em]{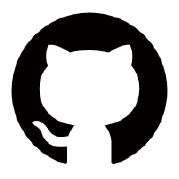}}}

\contribution[*]{Equal Contribution}
\checkdata[ \emailicon \hspace{0.3em} Email ]{\email{sunlinzhuang21@mails.ucas.ac.cn}, \email{Zzzyxiia@outlook.com}, \email{ruitong.jerry@gmail.com}, \\  \email{wentao.zhang@pku.edu.cn}}

\newcommand{\sourcelink}{The code will be released soon.}
\newcommand{\datalink}{https://huggingface.co/datasets/OpenDCAI/dataflow-instruct-10k}
\checkdata[ \githubicon \hspace{0.3em} Source Code ]{ {\sourcelink} }
\begin{document}
\maketitle

% footnote
\renewcommand{\thefootnote}{\fnsymbol{footnote}} 
\setcounter{footnote}{0}

\renewcommand{\thefootnote}{\arabic{footnote}}
\pagestyle{fancy}
\fancyhf{}
\fancyhead[L]{}
\fancyhead[R]{\thepage}

% % Content
% \newpage
% \tableofcontents
% \newpage

\section{Introduction}

\begin{figure*}[t]
    \centering
    \includegraphics[width=0.95\textwidth]{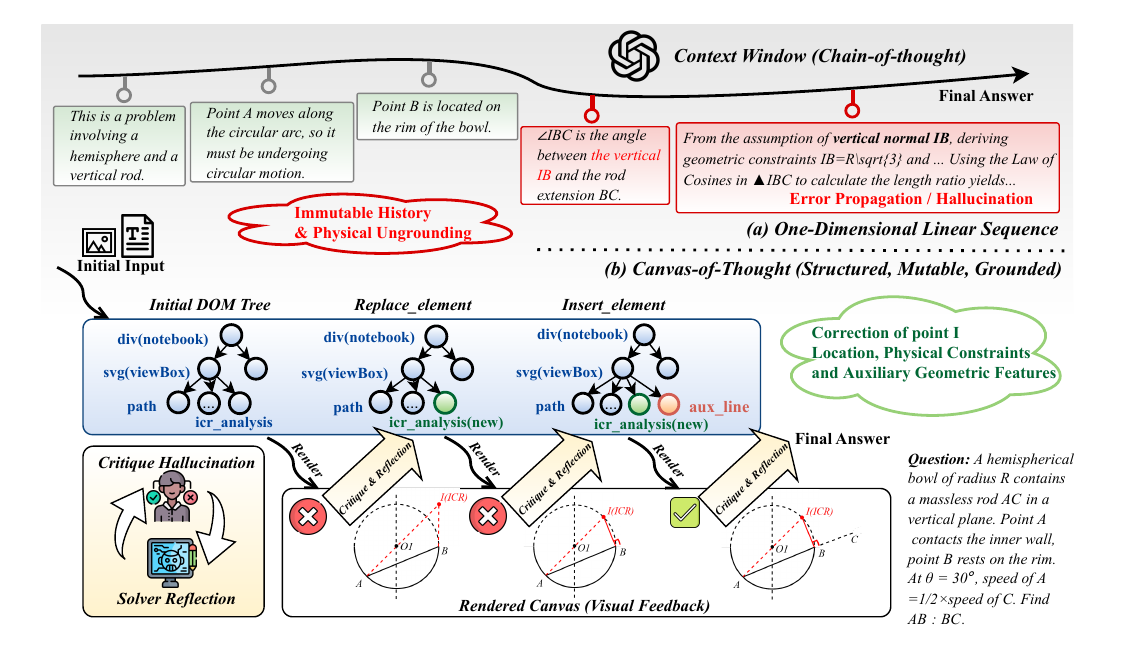}
    \caption{Paradigm Shift: Linear Textual Narration to Stateful Visual Modeling. (a): Text-CoT exhibits latent constraint hallucination. An implicit horizontal sliding prior mislocates the Instantaneous Center of Rotation, deriving a false geometry $IB=R\sqrt{3}$ obscured by linear algebraic steps. (b): Canvas-CoT externalizes this assumption via the DOM. The Rendering-Critique loop flags spatial conflicts, triggering a \texttt{replace\_element} operation to realign the vector, and correct the ICR. This restores the valid geometry $IB=R$, followed by \texttt{insert\_element} to finalize auxiliary lines and point $C$.}
    \label{fig:demo}
\end{figure*}

The integration of Chain-of-Thought (CoT) prompting \cite{wei2022chain} has significantly advanced the reasoning capabilities of Multimodal Large Language Models (MLLMs) \cite{wang2025multimodal}, enabling them to decompose complex queries into intermediate logical steps. Despite these advancements, recent evaluations on graduate-level benchmarks such as RBench \cite{guo2025rbench, guo2025rbenchv} and MathVista \cite{lu2024mathvista} reveal that even frontier models (GPT-5 \cite{openai_gpt5_2025}, Gemini 2.5 \cite{comanici2025gemini} and Gemini 3 \cite{google2025gemini3}) still struggle with complex multimodal reasoning. A fundamental limitation persists: \textcolor{conferencered}{\emph{the structural mismatch between the linear, autoregressive nature of text generation and the mutable, multi-dimensional nature of complex reasoning.}}

Current MLLM paradigms typically treat reasoning as a linear, append-only stream \cite{openai_gpt5_2025,openai2024gpt4o,comanici2025gemini,google2025gemini3,li2024llava,yang2025qwen3,zhu2025internvl3}. In this setting, the reasoning history is effectively immutable; once a thought token is generated, it becomes a fixed artifact. Consequently, correcting a local error, such as misplacing a single geometric coordinate, necessitates either generating verbose downstream corrections or regenerating the entire context. This inherent rigidity forces the model to implicitly maintain and track state updates within its context window, which significantly increases cognitive load and token consumption. Concurrently, empirical studies have identified critical pathologies in such reasoning chains, including hallucination snowballing \cite{he-etal-2025-mmboundary}, an overthinking effect where accuracy declines with reasoning length \cite{anonymous2026when}, and visual neglect of crucial spatial inputs \cite{yang2025look}. These challenges are particularly acute in high-dimensional domains like geometry and vector graphics (SVG), where textual descriptions often lack the precision to represent holistic spatial relationships, leading to accumulated errors that are difficult to rectify through text alone.

To address these challenges, we introduce \textbf{Canvas-of-Thought (Canvas-CoT)}, a framework that augments standard reasoning with an interactive, \textbf{DOM-based external substrate}. Rather than relying solely on linear text sequences, Canvas-CoT utilizes a HTML page to structure reasoning artifacts into discrete, identifiable objects (e.g., $<rect~id="obj\_1">$). This approach introduces two key capabilities that bridge the gap between textual reasoning and spatial execution:

\begin{enumerate}
\item \textbf{Structured State Manipulation:} By leveraging the Document Object Model (DOM), our framework allows the model to perform atomic operations (Insert, Replace, Modify, Delete) on specific elements. Unlike standard CoT, which requires appending new text to describe changes, Canvas-CoT enables \textit{in-place} revisions. The model can target and modify a specific node via its DOM id.
\item \textbf{Visual-Physical Grounding:} We integrate a rendering-based critique loop that serves as a constraint validator. In purely textual reasoning, models may hallucinate plausible-sounding but geometrically impossible configurations (e.g., overlapping solids). By rendering the DOM state, Canvas-CoT externalizes these hidden conflicts into explicit visual feedback. This allows the model to perceive the discrepancy between its intended design and the actual execution, facilitating grounded self-correction.
\end{enumerate}
Our contributions are summarized as follows:
\begin{itemize}
\item We identify the limitations of linear text generation in handling mutable reasoning states and propose \textbf{Canvas-CoT}, a framework that employs the HTML DOM as a structured external state.
\item We introduce an interactive update mechanism that enables targeted state revisions, allowing models to refine specific reasoning steps without the redundancy of full context regeneration.
\item We demonstrate that integrating a rendering-based feedback loop significantly improves performance on complex reasoning tasks. Extensive experiments on VCode, RBench-V, and MathVista show that Canvas-CoT effectively reduces reasoning errors and outperforms existing baselines.
\end{itemize}

\section{Related Work}
Recent MLLM reasoning paradigms bifurcate into linear explicit and linear latent substrates. Explicit methods, ranging from foundational Chain-of-Thought (CoT) \cite{wei2022chain} to search-based (ToT/GoT) \cite{yao2023tree, besta2024graph} and code-centric frameworks (PoT/VCode) \cite{chen2022program, lin2025vcode}, enable deliberate "System 2" planning \cite{xiang2025towards}. However, these approaches remain constrained by an immutable, append-only topology. Because the reasoning trace is treated as a monolithic log, correcting local errors imposes a high serialization tax, necessitating verbose downstream regeneration \cite{DBLP:journals/corr/abs-2501-12948, sui2025stop, yang2025longlive}. Conversely, latent approaches (e.g., Coconut, CODI) \cite{hao2024training, shen2025codi} compress reasoning into continuous vectors to enhance inference density. Yet, this efficiency sacrifices analyzability, rendering internal states opaque and untraceable \cite{yue2025hybrid, shen2025efficient}.

\section{Canvas-CoT Methodology}
\label{sec:method}

\begin{figure}[t]
    \centering
    \includegraphics[width=1.0\textwidth]{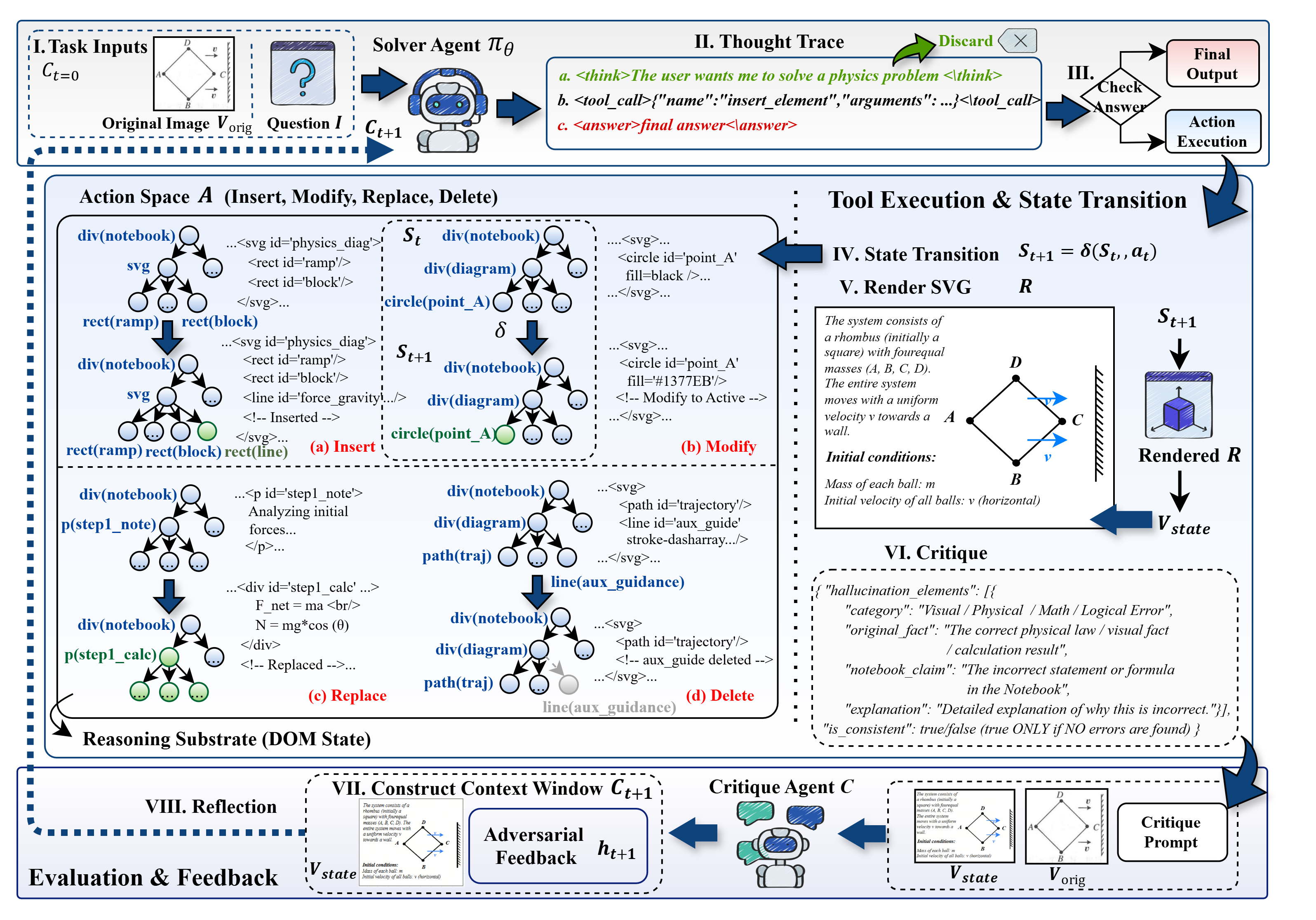}
    \caption{Overview of the Canvas-CoT Pipeline.}
    \label{fig:method}
\end{figure}

In this section, we detail the operational mechanics of the Canvas-CoT. We conceptualize the reasoning process as an iterative optimization over a structured state space, guided by an adversarial feedback loop.

\textbf{Decoupled Architecture:} Unlike monolithic CoT that conflates reasoning logic and state persistence within the KV-cache, Canvas-CoT introduces a decoupled architecture. We transform the LLM from a narrator into a Stateful Controller, externalizing the epistemic state into a structured DOM substrate. This separation enables the model to manage complex constraints through explicit state manipulation rather than probabilistic sequence extension.

\subsection{Preliminaries}
\label{sec:preliminaries}

% \subsection{Formal Problem Definition}
\paragraph{Task Definition}

Let $\mathcal{T}$ be a reasoning task given by a natural language instruction. Instead of generating a flat sequence of tokens $\mathbf{y} = (y_1, y_2, \dots, y_n)$, our framework generates a sequence of states and actions $(\mathcal{S}_0, \mathcal{A}_0, \mathcal{S}_1, \mathcal{A}_1, \dots, \mathcal{S}_T)$, where each $\mathcal{S}_t$ represents a structured representation of the model's current epistemic state.

% \subsection{The Reasoning Substrate: The DOM State}
\paragraph{The Reasoning Substrate: The DOM State}

The core of our framework is the Reasoning Substrate $\mathcal{S}$, defined as a DOM-tree:
\begin{equation}
\label{eq:statedef}
\mathcal{S} = \{ \mathcal{V}, \mathcal{E}, \mathcal{P} \}
\end{equation}
Where:
\begin{itemize}
    \item $\mathcal{V} = \{v_1, v_2, \dots, v_k\}$ is a set of Reasoning Artifacts (e.g., geometric points, circuit components, or logic nodes), represented as HTML elements.
    \item $\mathcal{E}$ represents the Hierarchical Relations (parent-child nodes) and Topological Links (e.g., CSS-defined connections).
    \item $\mathcal{P}$ is the set of Attributes (e.g., coordinates, labels, or physical constraints) bound to each node.
\end{itemize}

% \subsection{The Action Space: CRUD Operations}
\paragraph{The Action Space: CRUD Operations}

The controller interacts with the substrate through a discrete action space $\mathcal{A}$, comprising four atomic CRUD operations:
\begin{equation}
\label{eq:action_space}
\mathcal{A} \in \{ \text{Insert, Replace, Modify, Delete} \}
\end{equation}
A transition function $\delta$ maps the current state and action to a subsequent state:
\begin{equation}
\label{eq:state_transition}
\mathcal{S}_{t+1} = \delta(\mathcal{S}_t, a_t)
\end{equation}
Unlike standard CoT where $\delta$ is a simple concatenation, in our framework, $\delta$ allows for non-monotonic updates, such as the modification or removal of previously established nodes $v_i \in \mathcal{V}$ when a logical contradiction is detected.

% \subsection{The Observation Space: Multi-Modal Feedback}
% After each action $a_t$, the model receives an observation $o_{t+1}$ composed of two modalities:
% (1) Symbolic Feedback ($\mathcal{F}_{sym}$): The updated DOM code and any execution logs (e.g., CSS layout engine warnings). And (2) Visual Feedback ($\mathcal{F}_{vis}$): The rendered canvas image $\mathcal{I} = \mathcal{R}(\mathcal{S})$, where $\mathcal{R}$ is the rendering engine (e.g., a headless Chromium browser).
% The objective of the model is to find a sequence of actions that maximizes the grounding score $\mathcal{G}(\mathcal{S}_T, \mathcal{T})$, ensuring the final state satisfies all constraints of the task.

\subsection{State Initialization}

The reasoning process begins with the initialization of the external memory. Let $\mathcal{S}_0$ denote the initial state of the Blackboard. At $t=0$, $\mathcal{S}_0$ consists of a minimal HTML5 boilerplate with an empty \texttt{<body>} container. The agent receives the task instruction $I$ and the original visual input $V_{orig}$. Unlike standard CoT, which relies solely on the implicit hidden states of the Transformer to maintain context, Canvas-CoT explicitly instantiates an external "World Model" that allows for visual debugging and state persistence.

\subsection{The Iterative Reasoning Cycle}

The core workflow of Canvas-CoT proceeds through an iterative loop that synchronizes the agent's internal thought process with the external structured state. This cycle breaks the linear dependency of autoregressive generation, allowing for what we term Stateful Visual Reasoning. The process at timestep $t$ is defined as follows.

\subsubsection{Atomic Reasoning and Action Generation}

The Solver Agent ($\mathcal{G}$), parameterized by $\theta$, takes the current context window $C_t$ as input (detailed in Section~\ref{sec:Ct_def}). To prevent error propagation common in long-context CoT, we enforce a "One Step Thinking" constraint. The agent generates a dual-output tuple:
\begin{equation}
\label{eq:policy}
(\tau_t, a_t) \sim \pi_\theta(\cdot \mid C_t)
\end{equation}
where:
\begin{itemize}
    \item $\tau_t$ is the Thought Trace (enclosed in \texttt{<think>} tags), representing a discrete unit of reasoning (e.g., "Analyze the color of the object at coordinates $x,y$").
    \item $a_t \in \mathcal{A}$ is the Tool Call (enclosed in $<tool\_call>$ tags), which executes a CRUD operation to update the external state.
For example, if $\tau_t$ deduces that a specific object exists, $a_t$ might correspond to an insert-element operation with specific SVG attributes (coordinates, fill color, ID).
\end{itemize}

\begin{algorithm}[t]
\caption{Canvas-CoT Inference Algorithm}
\label{alg:canvas-cot}
\begin{algorithmic}[1]
\REQUIRE Input Query $I$, Image $V_{\text{orig}}$, LLM $\pi_\theta$, Critic $\mathcal{C}$, Renderer $\mathcal{R}$
\ENSURE Final Answer $A$
\STATE Initialize $\mathcal{S}_0 \leftarrow \text{Empty DOM}$
\STATE Initialize history $H \leftarrow \emptyset$
\WHILE{not converged}
    \STATE $V_{\text{render}} \leftarrow \mathcal{R}(\mathcal{S}_t)$
    \STATE Construct prompt $P \leftarrow (I, V_{\text{orig}}, V_{\text{render}}, H)$
    \STATE Generate $(\tau_t, a_t) \sim \pi_\theta(P)$
    \IF{$\tau_t$ contains {\tt <answer>}}
        \STATE \textbf{return} ExtractAnswer($\tau_t$)
    \ENDIF
    \STATE $\mathcal{S}_{t+1} \leftarrow \delta(\mathcal{S}_t, a_t)$
    \STATE $V'_{\text{state}} \leftarrow \mathcal{R}(\mathcal{S}_{t+1})$
    \STATE $h \leftarrow \mathcal{C}(V_{\text{orig}}, V'_{\text{state}})$
    \STATE $H \leftarrow H \cup \{a_t, h\}$ \COMMENT{$\tau_t$ is discarded}
    \STATE $t \leftarrow t + 1$
\ENDWHILE
\end{algorithmic}
% \vspace{-10pt}
\end{algorithm}

\subsubsection{Non-monotonic State Transition}

Upon generating action $a_t$, the deterministic environment executes the operation on the DOM tree. The state transition is governed by the function $\delta$:
$$\mathcal{S}_{t+1} = \delta(\mathcal{S}_t, a_t)$$
Crucially, because $\delta$ supports modify and delete operations, the reasoning trajectory is non-monotonic. If the agent recognizes a previous error (e.g., an incorrect bounding box size), it updates the existing DOM node rather than appending a correction text to the end of the sequence. This ensures that $\mathcal{S}_{t+1}$ always represents the agent's current best belief of the world state, rather than a history of trial-and-error.  To rigorously quantify the efficiency of this non-monotonic transition, we analyze the cost of error correction.

% % edit复杂度分析，偏理论分析而非证明
% \begin{theorem}[Correction Complexity]
% \label{the:correction}
% For a reasoning trajectory of length $N$, the \textbf{expected token generation cost} to correct a random error scales linearly as $O(N)$ for autoregressive models, whereas Canvas-CoT achieves constant $O(1)$ complexity via discrete mutable updates.
% \end{theorem}
% \emph{Significance:} This explicitly eliminates the "serialization tax," theoretically decoupling the error correction cost from the trajectory length. \emph{Proof.} See Appendix~\ref{proof:correction}.

\subsection{Visual Grounding and Adversarial Critique}

A fundamental limitation of text-based CoT is the lack of physical grounding—a model can assert logical contradictions (e.g., overlapping objects that shouldn't overlap) without internal detection. We address this via a Rendering-Critique Loop, framing visual grounding as a closed-loop feedback control problem.
After the state transition, a headless browser engine $\mathcal{R}$ renders the DOM tree into a raster image $V_{state}$:
\begin{equation}
\label{eq:v_state}
V_{\text{state}} = \mathcal{R}(\mathcal{S}_{t+1})
\end{equation}
We introduce a secondary Critique Agent ($\mathcal{C}$), prompting it to act as an objective discriminator. The critic compares the ground truth image $V_{orig}$ with the rendered state $V_{state}$ to identify hallucinations:
\begin{equation}
\label{eq:ht+1}
h_{t+1} = \mathcal{C}(V_{\text{orig}}, V_{\text{state}}, I)
\end{equation}
where $h_{t+1}$ is a structured JSON object categorizing discrepancies into Attribute Errors, False Existences, or Spatial Conflicts. Rather than a passive suggestion, the feedback $h_{t+1}$ functions as a symbolic residual signal. It serves as a "visual gradient", explicitly informing the Generator Agent of the mismatch between its internal world model and reality.

%%%%既然 $h_{t+1}$ 是建议，那么 Theorem 3.2 的证明逻辑应建立在“只要反馈存在且准确，生成器就有能力根据反馈缩小偏差”这一假设上-->“单调减少误差”的过程

% \begin{theorem}[Convergence of Visual Grounding]
% \label{the:visual_grounding}
% Let $L$ be a semantic visual discrepancy metric. Assuming an $\epsilon$-consistent critique agent defined over the finite discrete state space of DOM trees, the Canvas-CoT process monotonically reduces $L(\mathcal{S}_t)$ and converges to the $\epsilon$-neighborhood of the ground truth in finite steps.
% \end{theorem}
% \emph{Significance:} This theoretical guarantee ensures that the adversarial loop eliminates the risk of infinite oscillation or error accumulation, strictly outperforming open-loop generation by bounding the maximum reasoning steps required. \emph{Proof.} See Appendix~\ref{proof:Convergence_Analysis}.

\subsection{Recurrent Context Optimization}
\label{sec:Ct_def}

To address the constraints of finite context windows and mitigate the accumulation of historical noise, we implement a Context Pruning strategy. At the end of each iteration $t$, the verbose thought trace $\tau_t$ is discarded, as its logical essence is already distilled into the persistent state. The context for the next step, $C_{t+1}$, is constructed not by concatenating history, but by aggregating the current persistent state and the critique feedback:
\begin{equation}
\label{eq:context_window}
C_{t+1} = \{ I, \text{ActionHistory}_{0 \dots t}, V_{\text{state}}, h_{t+1} \}
\end{equation}
This creates a Markovian-like property where the decision at $t+1$ depends primarily on the current structured state $\mathcal{S}_{t+1}$ (visualized via $V_{state}$) and the immediate feedback, rather than the potentially flawed textual history of previous reasoning steps.

\subsection{Termination}

The iterative process continues until the Generator Agent outputs a termination token (enclosed in $<answer>$), signifying that the structured state $\mathcal{S}$ sufficiently resolves the query $I$.

\section{DOM Tree Manipulation}

% The core of Canvas-CoT is the \textit{ID-addressable} reasoning substrate, which externalizes the model's epistemic state into a structured, mutable environment. By treating the reasoning history as a manageable DOM tree rather than an immutable text stream, the framework enables surgical updates to logic nodes without linear re-generation.
The core of Canvas-CoT is the \textit{ID-addressable} reasoning substrate. By treating the reasoning history as a manageable DOM tree, the framework enables in-place updates with the rendering of html canvas.

\subsection{ID-Addressable Lookup and Random Access}

Following the foundational definitions in Section \ref{sec:preliminaries}, we represent the reasoning substrate $\mathcal{S}$ as a structured DOM tree. To facilitate non-monotonic updates, we implement a hash-based address map $\mathcal{H}$:
\begin{equation}
    \mathcal{H}: \mathcal{ID} \to \text{NodePointer}
\end{equation}
This mechanism enables the system to perform a direct lookup $v_i = \mathcal{H}(ID_{target})$ to obtain the memory pointer for any vertex in $\mathcal{V}$, bypassing the need for a sequential scan over the reasoning history. This effectively decouples the cost of state access from the historical reasoning depth $N$.

\subsection{Deterministic Parsing Layer}

The transition from unstructured generation to structured state updates is mediated by a deterministic parsing function:
\begin{equation}
    \text{parse}(f) : \text{String} \to \mathcal{S}_{subtree}
\end{equation}
This function converts raw HTML/SVG fragments $f$ into memory-resident objects. It ensures \textit{operational verifiability} by rejecting malformed tags or missing attributes to maintain renderable states, and guarantees \textit{atomic consistency} by treating updates as discrete mutations. This prevents the redundant or contradictory descriptions typical of linear CoT, ensuring the substrate remains in a globally consistent state.
%structured sub-graphs. Crucially, $\text{parse}(f)$ enforces structural integrity; if a fragment is malformed or violates schema constraints, the system prevents the state transition, ensuring the substrate remains in a verifiable and consistent state.

\begin{table*}[htbp]
  \small
  \centering
  \setlength{\tabcolsep}{5pt}
  \caption{\textbf{Main results on VCode.} Accuracy (\%) comparisons. \colorbox{bggray}{Gray background} indicates base performance, and \colorbox{bgblue}{blue background} highlights superior results.}
  \resizebox{\textwidth}{!}{%
    \begin{tabular}{l|c|ccccccc|c|ccc}
    \toprule
    \multicolumn{1}{c|}{\multirow{2}[2]{*}{\textbf{Model Name}}} & \multirow{2}[2]{*}{\textbf{Methods}} & \multicolumn{7}{c|}{\textbf{MM-Vet}}                 & \textbf{MMMU} & \multicolumn{3}{c}{\textbf{CV-Bench}} \\
          &       & \textbf{Rec} & \textbf{Ocr} & \textbf{Know} & \textbf{Gen} & \textbf{Spat} & \textbf{Math} & \textbf{Avg.} & \textbf{Avg.} & \textbf{2D} & \textbf{3D} & \textbf{Avg.} \\
    \midrule
    Claude-4-Opus & Iterative Reflection & 30.4  & 52.3  & 13.9  & 18.5  & 49.5  & 50.4  & 37.5  & 42.5  & 41.6  & 58.3  & 49.9 \\
    OpenAI GPT-o3 & Iterative Reflection & 31.3  & 55.2  & 17.7  & 19.7  & 48.5  & 61.5  & 39.5  & 39.0  & 47.4  & 56.7  & 52.1 \\
    OpenAI GPT-4o & Iterative Reflection & 23.1  & 58.4  & 12.7  & 17.0  & 51.3  & 60.4  & 35.0  & 44.5  & 29.3  & 50.0  & 39.6 \\
    Seed-1.6-Thinking & Iterative Reflection & 18.9  & 46.5  & 8.1   & 11.9  & 44.1  & 38.5  & 28.7  & 43.2  & 45.3  & 51.7  & 48.5 \\
    \midrule
    Gemini2.5-pro & Chain-of-thought & \cellcolor{bggray}35.3  & \cellcolor{bggray}57.3  & \cellcolor{bggray}19.6  & \cellcolor{bggray}24.2  & \cellcolor{bggray}58.1  & \cellcolor{bggray}60.8  & \cellcolor{bggray}42.7  & \cellcolor{bggray}44.5  & \cellcolor{bggray}45.9  & \cellcolor{bggray}56.7  & \cellcolor{bggray}51.3 \\
    Gemini2.5-pro & Tree-of-thought & 32.7  & 59.3  & 20.2  & 25.7  & 57.3  & 57.7  & 43.8  & 45.2  & 47.4  & 60.0  & 53.7 \\
    Gemini2.5-pro & Program-of-thought & 32.0  & 56.3  & 16.7  & 22.0  & 54.7  & 46.2  & 40.6  & 45.8  & 50.0  & 61.7  & 55.9 \\
    Gemini2.5-pro & Iterative Reflection & 28.9  & 57.8  & 20.0  & 22.9  & 47.9  & 68.5  & 39.1  & 45.2  & 56.1  & 56.7  & 56.4 \\
    Gemini2.5-pro & Canvas-of-thought & 34.0  & \cellcolor{bgblue}62.2  & \cellcolor{bgblue}22.4  & \cellcolor{bgblue}26.1  & \cellcolor{bgblue}59.2  & 60.4  & \cellcolor{bgblue}44.2  & \cellcolor{bgblue}47.3  & 52.6  & \cellcolor{bgblue}68.3  & \cellcolor{bgblue}60.5 \\
    \midrule
    OpenAI GPT-5 & Chain-of-thought & \cellcolor{bggray}31.5  & \cellcolor{bggray}65.3  & \cellcolor{bggray}18.7  & \cellcolor{bggray}22.0  & \cellcolor{bggray}63.6  & \cellcolor{bggray}71.5  & \cellcolor{bggray}42.8  & \cellcolor{bggray}44.5  & \cellcolor{bggray}47.4  & \cellcolor{bggray}63.3  & \cellcolor{bggray}55.4 \\
    OpenAI GPT-5 & Tree-of-thought & 33.3  & 67.7  & 17.7  & 22.9  & 65.3  & 73.1  & 47.0  & 41.8  & 47.4  & 66.7  & 57.1 \\
    OpenAI GPT-5 & Program-of-thought & 33.3  & 62.5  & 15.5  & 20.0  & 60.0  & 71.5  & 44.0  & 43.1  & 42.5  & 65.0  & 53.8 \\
    OpenAI GPT-5 & Iterative Reflection & 34.0  & 64.9  & 20.5  & 23.8  & 60.5  & 65.4  & 43.9  & 42.5  & 51.8  & 66.7  & 59.2 \\
    OpenAI GPT-5 & Canvas-of-thought & \cellcolor{bgblue}36.1  & \cellcolor{bgblue}75.2  & 18.5  & \cellcolor{bgblue}24.9  & \cellcolor{bgblue}73.1  & \cellcolor{bgblue}78.5  & \cellcolor{bgblue}49.8  & \cellcolor{bgblue}46.6  & \cellcolor{bgblue}52.5  & \cellcolor{bgblue}70.0  & \cellcolor{bgblue}61.2 \\
    \bottomrule
    \end{tabular}%
    }
    % \vspace{-3mm}
  \label{tab:exp_vcode}%
\end{table*}%

\subsection{Formal Operational Mechanics}

The state transition function $S_{t+1} = \delta(S_t, a_t)$ is implemented through four atomic CRUD operations. These operations act directly on the memory objects identified by $\mathcal{H}$:

\subsubsection{Insert}
The system appends new reasoning artifacts into the existing hierarchy. Here, $ID_{root}$ specifies the \textbf{parent node} to which the new fragment is attached:
\begin{equation}
\begin{aligned}
    & \delta(S_t, \text{Insert}(f, ID_{root})) \\
    & \implies \mathcal{V}_{t+1} = \mathcal{V}_t \cup \{ \text{parse}(f) \}
\end{aligned}
\end{equation}
The artifact is mounted onto the branch identified by $\mathcal{H}(ID_{root})$, with an optional $\mathcal(ID_{before})$ parameter to control the topological sequence and rendering order relative to existing siblings. This allows the substrate to grow while maintaining precise logical and spatial links within the reasoning chain.

\subsubsection{Modify}
The system performs in-place attribute updates, where $ID_{target}$ identifies the \textbf{specific node} being revised:
\begin{equation}
\begin{aligned}
    & \delta(S_t, \text{Modify}(ID_{target}, \Delta P)) \\
    & \implies \mathcal{P}_{v_i} \leftarrow \mathcal{P}_{v_i} \oplus \Delta P
\end{aligned}
\end{equation}
where $v_i = \mathcal{H}(ID_{target})$ and $\oplus$ denotes the attribute overwriting operator. This enables surgical corrections to specific nodes.

\subsubsection{Replace}
The system executes an atomic swap of node references, where $ID_{target}$ refers to the obsolete node to be substituted:
\begin{equation}
\begin{aligned}
    &\delta(S_t, \text{Replace}(ID_{target}, f)) \implies \\
    &\quad \mathcal{V}_{t+1} = (\mathcal{V}_t \setminus \{v_{old}\}) \cup \{v_{new} \mid \text{parse}(f)\}
\end{aligned}
\end{equation}
By preserving the parent-child edges of the original $ID_{target}$, the new node $v_{new}$ inherits its predecessor's logical position, facilitating non-monotonic correction of localized hallucinations.

\subsubsection{Delete}
The system prunes the substrate, with $ID_{target}$ specifying the root of the subtree to be removed:
\begin{equation}
\begin{aligned}
    \delta(S_t, \text{Delete}(ID_{target})) \implies \\
    S_{t+1} = S_t \setminus \text{Subtree}(\mathcal{H}(ID_{target}))
\end{aligned}
\end{equation}
This recursively removes the designated artifact and all its descendants, purging invalidated reasoning paths and preventing conflicting hypotheses from persisting in the state.

By utilizing atomic updates via $\mathcal{H}$, Canvas-CoT reduces the re-generation cost of standard CoT to an mutation cost, rendering error correction complexity independent of historical trajectory depth.
\section{Experiments}
In this section, we empirically validate the Canvas-CoT framework through the following research questions:

\begin{itemize}
    \item \textbf{RQ1:} Does the structured state manipulation improve performance on tasks requiring complex output generation (e.g., SVG coding)? 
    % Table~\ref{tab:exp_vcode}
    \item \textbf{RQ2:} Can Canvas-CoT enhance multi-step reasoning capabilities? 
    % Table~\ref{tab:exp_rbenchv}, Table~\ref{tab:exp_mathvista}
    \item \textbf{RQ3:} How does the mutable state architecture impact output token efficiency?
    % Figure~\ref{fig:vcode_token}, Figure~\ref{fig:rbench_token}
    \item \textbf{RQ4:} What is the role of the rendering-critique loop in mitigating hallucinations? 
    % Figure~\ref{fig:case}
\end{itemize}

% Table generated by Excel2LaTeX from sheet 'Sheet1'
\begin{table*}[t]
  \centering
  \small
  % \tiny
  \setlength{\tabcolsep}{4pt}
  \caption{\textbf{Main results on RBench-V.} \colorbox{bggray}{Gray background} indicates base performance (CoT, Budget=1), and \colorbox{bgblue}{blue background} highlights the best results within each model group.}
    \resizebox{\textwidth}{!}{%
    \begin{tabular}{l|c|cc|cc|cc|cc|cc}
    \toprule
    \multicolumn{1}{c|}{\multirow{2}[2]{*}{\textbf{Model Name}}} & \multirow{2}[2]{*}{\textbf{Method}} & \multicolumn{2}{c|}{\textbf{Overall}} & \multicolumn{2}{c|}{\textbf{Math}} & \multicolumn{2}{c|}{\textbf{Physics}} & \multicolumn{2}{c|}{\textbf{Counting}} & \multicolumn{2}{c}{\textbf{Game}} \\
          &       & \textbf{Score} & \textbf{Token} & \textbf{Score} & \textbf{Token} & \textbf{Score} & \textbf{Token} & \textbf{Score} & \textbf{Token} & \textbf{Score} & \textbf{Token} \\
    \midrule
    OpenAI GPT-4.5 & Chain-of-thought & 12.6  & -     & 18.2  & -     & 2.5   & -     & 11.8  & -     & 15.3  & - \\
    Doubao-1.5-vision-pro & Chain-of-thought & 15.6  & -     & 30.1  & -     & 8.9   & -     & 12.8  & -     & 12.0  & - \\
    GPT-5-mini & Chain-of-thought & 25.2  & -     & 49.4  & -     & 19.1  & -     & 16.9  & -     & 15.3  & - \\
    \midrule
    Gemini2.5-pro & Chain-of-thought & \cellcolor{bggray}30.0  & \cellcolor{bggray}1.3k  & \cellcolor{bggray}47.7  & \cellcolor{bggray}1.4k  & \cellcolor{bggray}46.5  & \cellcolor{bggray}1.6k  & \cellcolor{bggray}25.1  & \cellcolor{bggray}0.9k  & \cellcolor{bggray}12.7  & \cellcolor{bggray}1.3k \\
    Gemini2.5-pro & Tree-of-thought & 30.4  & 8.2k  & 48.3  & 9.1k  & 45.9  & 13.7k & 25.6  & 7.3k  & 13.5  & 5.2k \\
    Gemini2.5-pro & Program-of-thought & 31.8  & 7.4k  & 48.9  & 8.8k  & 47.8  & 11.7k & 26.2  & 6.9k  & 15.6  & 4.5k \\
    Gemini2.5-pro & Iterative Reflection & 25.8  & 7.3k  & 42.1  & 8.5k  & 40.1  & 11.6k & 20.5  & 6.7k  & 10.9  & 4.4k \\
    Gemini2.5-pro & Canvas-of-thought & 31.5  & 7.2k  & \cellcolor{bgblue}49.4  & 8.5k  & \cellcolor{bgblue}48.3  & 9.4k  & \cellcolor{bgblue}26.6  & 6.2k  & \cellcolor{bgblue}18.9  & 5.7k \\
    \midrule
    OpenAI GPT-5 & Chain-of-thought & \cellcolor{bggray}17.4  & \cellcolor{bggray}0.7k  & \cellcolor{bggray}30.7  & \cellcolor{bggray}0.8k  & \cellcolor{bggray}15.3  & \cellcolor{bggray}1.5k  & \cellcolor{bggray}12.3  & \cellcolor{bggray}0.5k  & \cellcolor{bggray}13.8  & \cellcolor{bggray}0.3k \\
    OpenAI GPT-5 & Tree-of-thought & 20.2  & 6.1k  & 34.1  & 6.5k  & 17.2  & 8.7k  & 16.4  & 5.1k  & 15.6  & 5.0k \\
    OpenAI GPT-5 & Program-of-thought & 23.5  & 2.8k  & 42.1  & 3.4k  & 19.8  & 4.8k  & 20.5  & 2.2k  & 16.0  & 1.8k \\
    OpenAI GPT-5 & Iterative Reflection & 12.7  & 2.6k  & 22.7  & 3.0k  & 12.1  & 4.5k  & 8.7   & 2.1k  & 9.5   & 1.7k \\
    OpenAI GPT-5 & Canvas-of-thought & \cellcolor{bgblue}32.4  & 1.0k  & \cellcolor{bgblue}48.9  & 1.1k  & \cellcolor{bgblue}47.1  & 1.4k  & \cellcolor{bgblue}27.2  & 0.9k  & \cellcolor{bgblue}16.7  & 0.9k \\
    \bottomrule
    \end{tabular}%
    }
    % \vspace{-4mm}
  \label{tab:exp_rbenchv}%
\end{table*}%

\subsection{Experiment Setup}

\subsubsection{Baseline Models.}

We compare Canvas-CoT based on \textbf{Gemini-2.5-pro} and \textbf{GPT-5} separately against four representative paradigms:
\textbf{(1) Standard CoT}: The foundational method that prompts the model to generate intermediate reasoning steps before arriving at a final answer. \textbf{(2) Tree-of-thought}: Inspired by~\citet{sun2024beats}, this framework incorporates three specific actions: Question Disambiguation, One Step Forward, and Giving Answer. The model navigates the reasoning space using a Breadth-First Search (BFS) algorithm. \textbf{(3) Program-of-thought}: Inspired by~\citet{lin2025vcode}, this approach provides the model with external tools and a ReAct paradigm to get the final answer. \textbf{(4) Iterative Reflection}: We adopt a multi-turn reasoning loop where the model undergoes a "Think, Solve, Reflect, and Refine" pipeline. In this setup, the critic model is GPT-5 or Gemini-2.5-Pro, following the configuration of Canvas-CoT.

It is important to note that such baselines are inherently text-centric. Their reasoning processes operate within an implicit semantic space, lacking an explicit, renderable external substrate (e.g., a DOM tree). Consequently, these paradigms are structurally incapable of supporting a grounded visual critique loop, relying instead solely on the LLM's internal parametric knowledge for self-correction.

% We utilize \textbf{Gemini-2.5-pro} and \textbf{GPT-5} as backbone models to demonstrate that our framework's benefits are agnostic to the underlying MLLM.

\subsubsection{Settings and Inference Budget.}
To ensure a fair comparison of computational resources, we introduce the concept of an Inference $Budget$. (1) For Standard CoT, we set $Budget=1$, representing a direct generation. (2) For PoT, Iterative Reflection and Canvas-CoT, we set the $Budget=6$, allowing the model to sample up to $6$ reasoning paths or perform $6$ rounds of revision. (3) for ToT, we set the $Budget=10$.

\subsubsection{Benchmarks.}

\textbf{VCode}: Visual Coding Task. Evaluates the generation of SVG code from visual prompts. This tests the model's ability to translate visual intentions into structured DOM representations.

\textbf{RBench-V}: A suite focusing on physical reasoning, counting, mathematical reasoning and game reasoning, where purely textual reasoning often fails due to lack of spatial grounding.

\textbf{MathVista}: A multimodal mathematical reasoning dataset requiring precise visual parsing and multi-step logical deduction, which covers algebraic, geometry problem solving, visual QA and textbook QA.

\begin{figure*}[t]
    \centering
    % 左侧图片
    \begin{minipage}{0.48\textwidth}
        \centering
        \includegraphics[width=\linewidth]{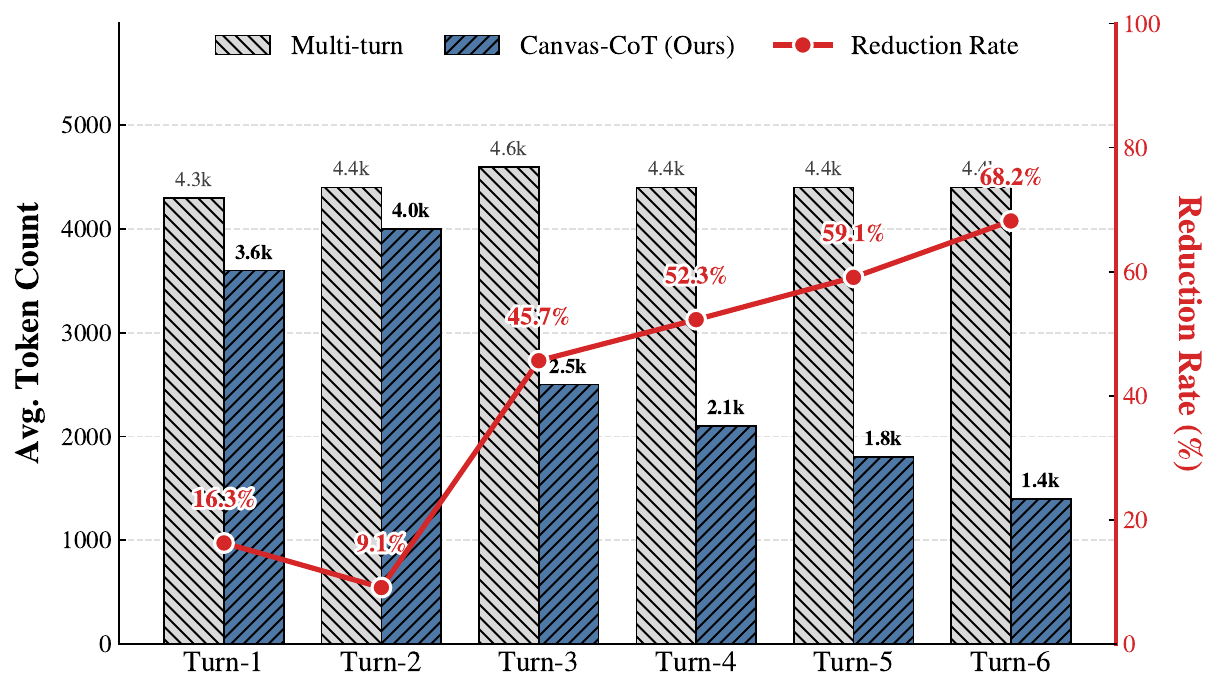}
        \caption{VCode Bench Token.}
        \label{fig:vcode_token}
    \end{minipage}
    \hfill % 在两个 minipage 之间填满空白，实现左右对齐
    % 右侧图片
    \begin{minipage}{0.48\textwidth}
        \centering
        \includegraphics[width=\linewidth]{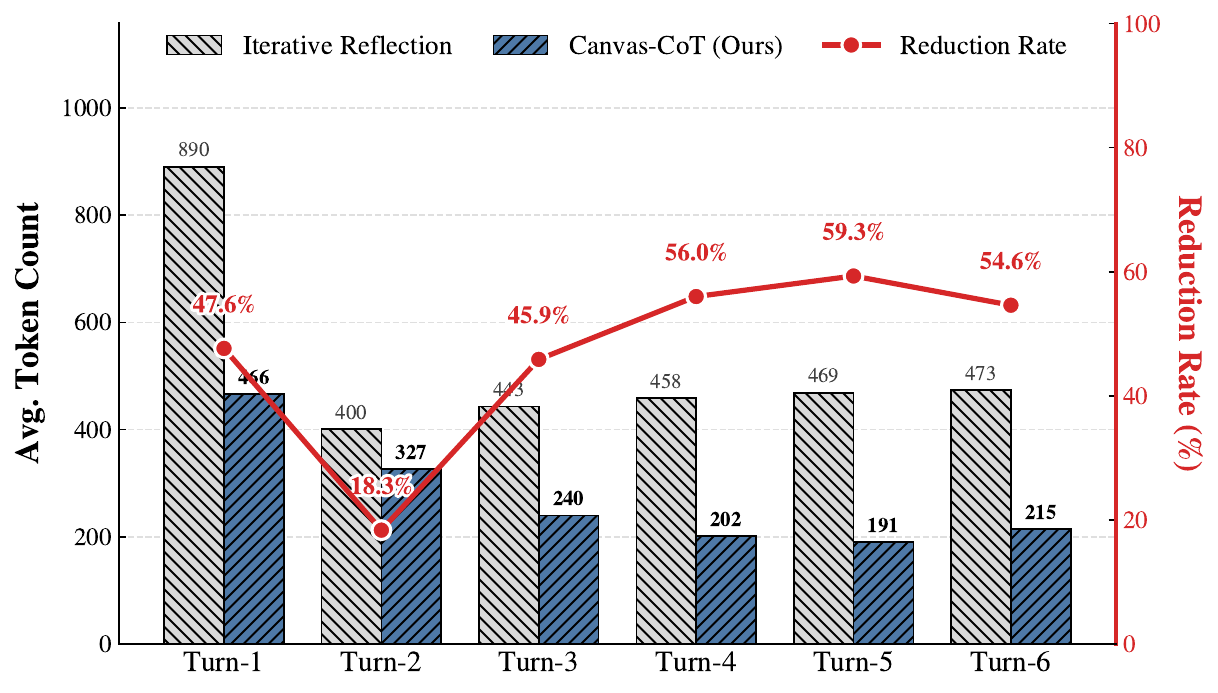}
        \caption{RBench-V Token.}
        \label{fig:rbench_token}
    \end{minipage}
    % \vspace{-10pt} % 如果需要整体缩小与下方正文的距离
\end{figure*}

\begin{table*}[t]
  \centering
  \small
  \setlength{\tabcolsep}{2.5pt}
  \caption{\textbf{Main results on MathVista.} \colorbox{bggray}{Gray background} indicates base performance (CoT), \colorbox{bgblue}{blue background} highlights best results, and \underline{underline} marks second-best.}
  \resizebox{\textwidth}{!}{%
    \begin{tabular}{c|c|c|cc|cc|cc|cc}
    \toprule
    \multirow{2}[2]{*}{\textbf{Model Name}} & \multirow{2}[2]{*}{\textbf{Methods}} & \multirow{2}[2]{*}{\textbf{\# Budget}} & \multicolumn{2}{c|}{\textbf{ALG}} & \multicolumn{2}{c|}{\textbf{GPS}} & \multicolumn{2}{c|}{\textbf{VQA}} & \multicolumn{2}{c}{\textbf{FQA}} \\
          &       &       & \textbf{pass @ 1} & \textbf{pass @ 2} & \textbf{pass @ 1} & \textbf{pass @ 2} & \textbf{pass @ 1} & \textbf{pass @ 2} & \textbf{pass @ 1} & \textbf{pass @ 2} \\
    \midrule
    \multirow{5}[2]{*}{GPT-5} & Chain-of-thought & 1     & \cellcolor{bggray}82.9  & \cellcolor{bggray}86.5  & \cellcolor{bggray}83.2  & \cellcolor{bggray}86.5  & \cellcolor{bggray}54.8  & \cellcolor{bggray}57.5  & \cellcolor{bggray}72.1  & \cellcolor{bggray}75.8 \\
          & Tree-of-thought & 10     & 82.8  & 86.4  & 83.1  & 86.8  & 58.4  & 62.9  & 72.2  & 76.2 \\
          & Program-of-thought & 6     & 82.2  & 86.3  & 83.0  & 86.8  & 57.8  & 61.1  & 72.4  & 76.5 \\
          & Iterative Reflection & 6     & 82.4  & 86.3  & 82.8  & 86.9  & 59.1  & 63.3  & 72.3  & 76.9 \\
          & Canvas-of-thought & 6     & \cellcolor{bgblue}83.9  & \cellcolor{bgblue}87.1  & 82.2  & \cellcolor{bgblue}87.0  & \cellcolor{bgblue}62.0  & \cellcolor{bgblue}66.5  & \cellcolor{bgblue}72.5  & \cellcolor{bgblue}77.7 \\
    \midrule
    \multirow{5}[2]{*}{Gemini-2.5-pro} & Chain-of-thought & 1     & \cellcolor{bggray}84.7  & \cellcolor{bggray}86.3  & \cellcolor{bggray}86.1  & \cellcolor{bggray}87.6  & \cellcolor{bggray}69.3  & \cellcolor{bggray}70.9  & \cellcolor{bggray}84.4  & \cellcolor{bggray}86.4 \\
          & Tree-of-thought & 10     & 84.5  & 86.2  & 86.0  & 87.7  & 69.9  & 71.4  & 84.5  & 86.6 \\
          & Program-of-thought & 6     & 83.9  & 85.8  & 85.8  & 87.6  & 70.2  & 71.0  & 84.7  & 86.8 \\
          & Iterative Reflection & 6     & 84.2  & 86.0  & 85.7  & 87.6  & 70.8  & 71.2  & 84.6  & 87.0 \\
          & Canvas-of-thought & 6     & \cellcolor{bgblue}86.1  & \cellcolor{bgblue}87.6  & \cellcolor{bgblue}89.9  & \cellcolor{bgblue}92.3  & \cellcolor{bgblue}70.9  & \cellcolor{bgblue}72.1  & 84.6  & \cellcolor{bgblue}87.3 \\
    \bottomrule
    \end{tabular}%
    }
    % \vspace{-4mm}
  \label{tab:exp_mathvista}%
\end{table*}%

\subsection{RQ1: Effectiveness on Complex Visual Generation}

% We first evaluate the model's capability in generating complex, structured visual code using the VCode benchmark. A fundamental failure mode in standard MLLMs is structural collapse, where as the length of the generated SVG code increases, the model often loses track of nested hierarchies, leading to unclosed tags or syntax errors. Canvas-CoT effectively eliminates this bottleneck through its ID-addressable architecture. Instead of predicting a monolithic code block in a single autoregressive pass, our method treats the SVG as a collection of discrete, addressable objects. By utilizing unique DOM IDs, the model can target and refine specific substructures without regenerating the surrounding syntax. Table~\ref{tab:exp_vcode} demonstrates this advantage. Canvas-CoT achieves a 61.2\% Average Score with GPT-5, significantly outperforming the Iterative Reflection baseline (59.2\%). This confirms that decoupling object manipulation from code serialization prevents structural degradation, allowing the model to maintain high precision even in complex generation tasks.

Evaluation on the VCode benchmark demonstrates that Canvas-CoT effectively eliminates the structural collapse common in long-sequence autoregressive generation. By treating SVG elements as discrete, ID-addressable objects rather than a monolithic stream, our method prevents the syntactic degradation typical of standard MLLMs. As shown in Table~\ref{tab:exp_vcode}, Canvas-CoT (GPT-5) achieves a higher average score, significantly outperforming the Iterative Reflection baseline by maintaining structural precision throughout the generation process.

% \begin{figure}[htbp]
%     \includegraphics[width=0.47\textwidth]{imags/pdf/vcode_token.pdf}
%     \caption{VCode Bench Token.}
%     \label{fig:vcode_token}
%     \vspace{-20pt}
% \end{figure}

% \begin{figure}[htbp]
%     \includegraphics[width=0.47\textwidth]{imags/pdf/rbench_token.pdf}
%     \caption{RBench Token.}
%     \label{fig:rbench_token}
%     \vspace{-20pt}
% \end{figure}

\subsection{RQ2: Effectiveness on Complex Reasoning Processes}

% To assess generalization beyond code generation, we evaluate performance on RBench-V and MathVista.
% \textbf{(1) RBench-V Results.} As shown in Table~\ref{tab:exp_rbenchv}, Canvas-CoT establishes a new state-of-the-art. On the Gemini-2.5-pro backbone, it achieves an Overall Score of 47.7, surpassing the strong Tree-of-Thought baseline (30.4) and Program-of-Thought (31.8). The improvement is particularly pronounced in the Math Score (48.9 vs. 31.5) and Physics Score (42.1 vs. 25.8).
% \textbf{(2) MathVista Results.} Table 3 corroborates these findings. Canvas-CoT achieves 68.3\% accuracy on the complex reasoning split, outperforming CoT (60.5\%) and Iterative Reflection (66.7\%)

On multi-step reasoning tasks such as RBench-V (Table~\ref{tab:exp_rbenchv}) and MathVista (Table~\ref{tab:exp_mathvista}), Canvas-CoT addresses the challenge of state tracking by utilizing the DOM substrate as an external state. Key physical properties and geometric relationships are explicitly written into the canvas attributes (e.g., updating a coordinate from x=10 to x=12 based on velocity). As shown in Table~\ref{tab:exp_rbenchv}, this explicit state persistence leads to superior performance. Using the GPT-5 backbone, Canvas-CoT achieves an Overall Score of 32.4, significantly surpassing the Tree-of-Thought (20.2) and Program-of-Thought (23.5) baselines. Notably, our method shows substantial gains in Physics and Math sub-tasks that demand rigorous state tracking. This evidence suggests that offloading the burden of implicit state maintenance to an external canvas significantly reduces the cognitive load on the MLLM, thereby mitigating error accumulation in multi-step reasoning.

% \begin{figure*}[t]
%     \centering
%     \includegraphics[width=0.9\textwidth]{imags/rbench_token.png}
%     \caption{Token expenses on RBench-V benchmark. Stacked bars show Input (Bottom) + Output (Top)}
%     \label{fig:rbench_token}
% \end{figure*}

% \begin{itemize}
%     \item Linear Baseline: When an error occurs (e.g., incorrect coordinates), standard CoT must regenerate the entire reasoning chain or append verbose textual corrections, accumulating "historical debt."

%     \item Canvas-CoT: The agent utilizes Modify or Delete operations to rectify the DOM node directly.
% \end{itemize}

% \paragraph{Correction Success Rate.}
% Further analysis reveals that in 35\% of successful trajectories, the model performed at least one Update operation to correct an intermediate state. This confirms that the efficiency gains stem from the model's ability to maintain a pristine "current state" ($S_t$) rather than a cluttered history of trial-and-error.

\subsection{RQ3: Token Efficiency of Canvas-CoT}

As illustrated in Figure~\ref{fig:vcode_token}, the immutable nature of traditional CoT imposes a significant redundancy overhead: each reflection step necessitates the regeneration of the entire SVG code block. This inefficiency is particularly pronounced in high-dimensional tasks such as RBench-V. In Figure~\ref{fig:rbench_token}, although the final output in these tasks is often a concise scalar, correctly deriving the solution requires maintaining and manipulating a auxiliary visual reasoning state. Consequently, standard Iterative Reflection baselines are struggle to allocate more token budgets to describe these spatial configurations through text.
In contrast, Canvas-CoT overcomes this bottleneck by leveraging atomic operations to perform surgical updates (e.g., insert, delete, modify) on specific DOM elements. This decouples correction costs from context length. As reasoning converges, the requisite state updates diminish, resulting in a decrease in token consumption. While integrating a rendering loop introduces a constant visual encoding overhead, it significantly mitigates the prohibitive serialization cost of autoregressive regeneration.

% A fundamental limitation of purely CoT-based MLLMs is their lack of physical grounding, which often allows for the hallucination of logically or physically impossible scenarios (e.g., floating objects or overlapping solids) without detection. We hypothesized that our Rendering-Critique Loop serves as a hard constraint validator to filter these hallucinations.

% \paragraph{Physical Grounding Improvements}
% Our results on the RBench-V benchmark (Table~\ref{tab:exp_rbenchv}), which specifically targets physical and spatial reasoning, provide strong evidence for this hypothesis. In the Physics split, the GPT-5 model equipped with Canvas-CoT achieves a score of 36.9, a dramatic improvement over the Standard CoT baseline of 19.7. This suggests that the adversarial feedback from the rendering engine effectively acts as a "visual gradient," informing the agent of discrepancies between its internal world model and physical reality.

% \paragraph{Visual Semantic Consistency}
% To further quantify the reduction in hallucinations, we analyzed the visual consistency between the generated code and the task description using SigLIP similarity scores. As shown in Figure~\ref{fig:siglip}, the Canvas-CoT architecture consistently outperforms the Native baselines across all benchmarks, raising the overall SigLIP score from 68.01 to 69.27. This metric confirms that the generated visual states are not only syntactically correct but also semantically aligned with the ground truth constraints, effectively minimizing the "logical fragility" observed in text-only reasoning.

\subsection{RQ4: Ablation Study on the Critique Loop}

\begin{table}[htbp]
  % \tiny
  \small
  \centering
  \setlength{\tabcolsep}{3pt}
  \caption{Ablation Study on the Rendering-Critique Loop.}
  \resizebox{0.5\textwidth}{!}{%
    \begin{tabular}{l|c|ccc}
    \toprule
    \multirow{2}[2]{*}{\textbf{Methods}} & \multirow{2}[2]{*}{\textbf{Rbench-V}} & \multicolumn{3}{c}{\textbf{Vcode}} \\
          &       & \textbf{MM-Vet} & \textbf{MMMU} & \textbf{CV-Becnh} \\
    \midrule
    \multicolumn{5}{c}{\textit{GPT-5}} \\
    \midrule
    Iterative Reflection    & 17.4  & 42.8  & 44.5  & 55.4 \\
    + DOM Canvas & 26.9  & 44.6  & 45.2  & 59.7 \\
    \textbf{+ Critical} & 32.4  & 49.8  & 46.6  & 61.2 \\
    \midrule
    \multicolumn{5}{c}{\textit{Gemini-2.5-Pro}} \\
    \midrule
    Iterative Reflection    & 30.0  & 42.7  & 44.5  & 51.3 \\
    + DOM Canvas & 30.8  & 43.8  & 45.3  & 56.1 \\
    \textbf{+ Critical} & 31.5  & 44.2  & 47.3  & 60.5 \\
    \bottomrule
    \end{tabular}%
    }
    % \vspace{-8pt}
  \label{tab:ablation}%
\end{table}%

To decouple the contribution of the structured DOM from the visual feedback mechanism, we conducted an ablation study on the GPT-5 and Gemini-2.5-Pro backbones. There are three configurations: (1) Iterative Reflection: The baseline without external state. (2) + DOM Canvas: The model uses the HTML substrate for state tracking but lacks the critical feedback. (3) + Critical: The full Canvas-CoT framework, enabling the Rendering-Critique loop to generate visual gradients. As shown in Table \ref{tab:ablation}, introducing the DOM Canvas alone yields notable improvements over the baseline. Crucially, the full integration of the Critical module triggers a second performance surge.

\section{Conclusion}

In this paper, we introduced Canvas-CoT, a novel framework that fundamentally shifts multimodal reasoning from a linear, immutable text generation process to a mutable, structured state manipulation task. By decoupling reasoning logic from state persistence, we demonstrated that utilizing an ID-addressable DOM substrate enables models to perform atomic CRUD operations, thereby allowing for surgical revisions without regenerating full contexts.
Furthermore, we addressed the critical issue of visual-physical ungrounding in standard CoT approaches by integrating a rendering-based critique loop. This mechanism externalizes hidden conflicts into explicit visual feedback, enabling the model to autonomously detect and correct geometric and spatial hallucinations.
Extensive evaluations across VCode, RBench-V, and MathVista confirm that Canvas-CoT significantly outperforms existing text-centric baselines, particularly in tasks requiring high-dimensional spatial reasoning.

\clearpage

\bibliographystyle{plainnat}
\bibliography{main}

@article{wei2022chain,
  title={Chain-of-thought prompting elicits reasoning in large language models},
  author={Wei, Jason and Wang, Xuezhi and Schuurmans, Dale and Bosma, Maarten and Xia, Fei and Chi, Ed and Le, Quoc V and Zhou, Denny and others},
  journal={Advances in neural information processing systems},
  volume={35},
  pages={24824--24837},
  year={2022}
}

@article{xiang2025towards,
  title={Towards system 2 reasoning in llms: Learning how to think with meta chain-of-thought},
  author={Xiang, Violet and Snell, Charlie and Gandhi, Kanishk and Albalak, Alon and Singh, Anikait and Blagden, Chase and Phung, Duy and Rafailov, Rafael and Lile, Nathan and Mahan, Dakota and others},
  journal={arXiv preprint arXiv:2501.04682},
  year={2025}
}

@article{yao2023tree,
  title={Tree of thoughts: Deliberate problem solving with large language models},
  author={Yao, Shunyu and Yu, Dian and Zhao, Jeffrey and Shafran, Izhak and Griffiths, Tom and Cao, Yuan and Narasimhan, Karthik},
  journal={Advances in neural information processing systems},
  volume={36},
  pages={11809--11822},
  year={2023}
}

@inproceedings{besta2024graph,
  title={Graph of thoughts: Solving elaborate problems with large language models},
  author={Besta, Maciej and Blach, Nils and Kubicek, Ales and Gerstenberger, Robert and Podstawski, Michal and Gianinazzi, Lukas and Gajda, Joanna and Lehmann, Tomasz and Niewiadomski, Hubert and Nyczyk, Piotr and others},
  booktitle={Proceedings of the AAAI conference on artificial intelligence},
  volume={38},
  number={16},
  pages={17682--17690},
  year={2024}
}

@article{DBLP:journals/corr/abs-2501-12948,
  publtype={informal},
  author={DeepSeek-AI and Daya Guo and Dejian Yang and Haowei Zhang and Junxiao Song and Ruoyu Zhang and Runxin Xu and Qihao Zhu and Shirong Ma and Peiyi Wang and Xiao Bi and Xiaokang Zhang and Xingkai Yu and Yu Wu and Z. F. Wu and Zhibin Gou and Zhihong Shao and Zhuoshu Li and Ziyi Gao and Aixin Liu and Bing Xue and Bingxuan Wang and Bochao Wu and Bei Feng and Chengda Lu and Chenggang Zhao and Chengqi Deng and Chenyu Zhang and Chong Ruan and Damai Dai and Deli Chen and Dongjie Ji and Erhang Li and Fangyun Lin and Fucong Dai and Fuli Luo and Guangbo Hao and Guanting Chen and Guowei Li and H. Zhang and Han Bao and Hanwei Xu and Haocheng Wang and Honghui Ding and Huajian Xin and Huazuo Gao and Hui Qu and Hui Li and Jianzhong Guo and Jiashi Li and Jiawei Wang and Jingchang Chen and Jingyang Yuan and Junjie Qiu and Junlong Li and J. L. Cai and Jiaqi Ni and Jian Liang and Jin Chen and Kai Dong and Kai Hu and Kaige Gao and Kang Guan and Kexin Huang and Kuai Yu and Lean Wang and Lecong Zhang and Liang Zhao and Litong Wang and Liyue Zhang and Lei Xu and Leyi Xia and Mingchuan Zhang and Minghua Zhang and Minghui Tang and Meng Li and Miaojun Wang and Mingming Li and Ning Tian and Panpan Huang and Peng Zhang and Qiancheng Wang and Qinyu Chen and Qiushi Du and Ruiqi Ge and Ruisong Zhang and Ruizhe Pan and Runji Wang and R. J. Chen and R. L. Jin and Ruyi Chen and Shanghao Lu and Shangyan Zhou and Shanhuang Chen and Shengfeng Ye and Shiyu Wang and Shuiping Yu and Shunfeng Zhou and Shuting Pan and S. S. Li},
  title={DeepSeek-R1: Incentivizing Reasoning Capability in LLMs via Reinforcement Learning},
  year={2025},
  month={January},
  cdate={1735689600000},
  journal={CoRR},
  volume={abs/2501.12948},
  url={https://doi.org/10.48550/arXiv.2501.12948}
}

@article{sui2025stop,
  title={Stop overthinking: A survey on efficient reasoning for large language models},
  author={Sui, Yang and Chuang, Yu-Neng and Wang, Guanchu and Zhang, Jiamu and Zhang, Tianyi and Yuan, Jiayi and Liu, Hongyi and Wen, Andrew and Zhong, Shaochen and Zou, Na and others},
  journal={arXiv preprint arXiv:2503.16419},
  year={2025}
}

@article{yang2025longlive,
  title={Longlive: Real-time interactive long video generation},
  author={Yang, Shuai and Huang, Wei and Chu, Ruihang and Xiao, Yicheng and Zhao, Yuyang and Wang, Xianbang and Li, Muyang and Xie, Enze and Chen, Yingcong and Lu, Yao and others},
  journal={arXiv preprint arXiv:2509.22622},
  year={2025}
}

@article{hao2024training,
  title={Training large language models to reason in a continuous latent space},
  author={Hao, Shibo and Sukhbaatar, Sainbayar and Su, DiJia and Li, Xian and Hu, Zhiting and Weston, Jason and Tian, Yuandong},
  journal={arXiv preprint arXiv:2412.06769},
  year={2024}
}

@article{shen2025codi,
  title={Codi: Compressing chain-of-thought into continuous space via self-distillation},
  author={Shen, Zhenyi and Yan, Hanqi and Zhang, Linhai and Hu, Zhanghao and Du, Yali and He, Yulan},
  journal={arXiv preprint arXiv:2502.21074},
  year={2025}
}

@article{chen2022program,
  title={Program of thoughts prompting: Disentangling computation from reasoning for numerical reasoning tasks},
  author={Chen, Wenhu and Ma, Xueguang and Wang, Xinyi and Cohen, William W},
  journal={arXiv preprint arXiv:2211.12588},
  year={2022}
}

@article{tan2025think,
  title={Think Silently, Think Fast: Dynamic Latent Compression of LLM Reasoning Chains},
  author={Tan, Wenhui and Li, Jiaze and Ju, Jianzhong and Luo, Zhenbo and Luan, Jian and Song, Ruihua},
  journal={arXiv preprint arXiv:2505.16552},
  year={2025}
}

@article{yue2025hybrid,
  title={Hybrid Latent Reasoning via Reinforcement Learning},
  author={Yue, Zhenrui and Jin, Bowen and Zeng, Huimin and Zhuang, Honglei and Qin, Zhen and Yoon, Jinsung and Shang, Lanyu and Han, Jiawei and Wang, Dong},
  journal={arXiv preprint arXiv:2505.18454},
  year={2025}
}

@article{shen2025efficient,
  title={Efficient reasoning with hidden thinking},
  author={Shen, Xuan and Wang, Yizhou and Shi, Xiangxi and Wang, Yanzhi and Zhao, Pu and Gu, Jiuxiang},
  journal={arXiv preprint arXiv:2501.19201},
  year={2025}
}

@inproceedings{yao2022react,
  title={React: Synergizing reasoning and acting in language models},
  author={Yao, Shunyu and Zhao, Jeffrey and Yu, Dian and Du, Nan and Shafran, Izhak and Narasimhan, Karthik R and Cao, Yuan},
  booktitle={The eleventh international conference on learning representations},
  year={2022}
}

@article{lin2025vcode,
  title={VCode: a Multimodal Coding Benchmark with SVG as Symbolic Visual Representation},
  author={Lin, Kevin Qinghong and Zheng, Yuhao and Ran, Hangyu and Zhu, Dantong and Mao, Dongxing and Li, Linjie and Torr, Philip and Wang, Alex Jinpeng},
  journal={arXiv preprint arXiv:2511.02778},
  year={2025}
}

@article{wang2025multimodal,
  title={Multimodal chain-of-thought reasoning: A comprehensive survey},
  author={Wang, Yaoting and Wu, Shengqiong and Zhang, Yuecheng and Yan, Shuicheng and Liu, Ziwei and Luo, Jiebo and Fei, Hao},
  journal={arXiv preprint arXiv:2503.12605},
  year={2025}
}

@inproceedings{
anonymous2026when,
title={When More is Less: Understanding Chain-of-Thought Length in {LLM}s},
author={Anonymous},
booktitle={The Fourteenth International Conference on Learning Representations},
year={2026},
url={https://openreview.net/forum?id=6QDFsYxtI1}
}

@article{yang2025look,
  title={Look-back: Implicit visual re-focusing in mllm reasoning},
  author={Yang, Shuo and Niu, Yuwei and Liu, Yuyang and Ye, Yang and Lin, Bin and Yuan, Li},
  journal={arXiv preprint arXiv:2507.03019},
  year={2025}
}

@inproceedings{he-etal-2025-mmboundary,
    title = "{MMB}oundary: Advancing {MLLM} Knowledge Boundary Awareness through Reasoning Step Confidence Calibration",
    author = "He, Zhitao  and
      Polisetty, Sandeep  and
      Fan, Zhiyuan  and
      Huang, Yuchen  and
      Wu, Shujin  and
      Fung, Yi R.",
    editor = "Che, Wanxiang  and
      Nabende, Joyce  and
      Shutova, Ekaterina  and
      Pilehvar, Mohammad Taher",
    booktitle = "Proceedings of the 63rd Annual Meeting of the Association for Computational Linguistics (Volume 1: Long Papers)",
    month = jul,
    year = "2025",
    address = "Vienna, Austria",
    publisher = "Association for Computational Linguistics",
    url = "https://aclanthology.org/2025.acl-long.802/",
    doi = "10.18653/v1/2025.acl-long.802",
    pages = "16427--16444",
    ISBN = "979-8-89176-251-0"
}

@inproceedings{
guo2025rbench,
title={{RB}ench: Graduate-level Multi-disciplinary Benchmarks for {LLM} \& {MLLM} Complex Reasoning Evaluation},
author={Meng-Hao Guo and Jiajun Xu and Yi Zhang and Jiaxi Song and Haoyang Peng and Yi-Xuan Deng and Xinzhi Dong and Kiyohiro Nakayama and Zhengyang Geng and Chen Wang and Bolin Ni and Guo-Wei Yang and Yongming Rao and Houwen Peng and Han Hu and Gordon Wetzstein and Shi-min Hu},
booktitle={Forty-second International Conference on Machine Learning},
year={2025},
url={https://openreview.net/forum?id=BeoXADC9PW}
}

@inproceedings{
guo2025rbenchv,
title={{RB}ench-V: A Primary Assessment for Visual Reasoning Models with Multimodal Outputs},
author={Meng-Hao Guo and Xuanyu Chu and Qianrui Yang and Zhe-Han Mo and Yiqing Shen and Pei-lin Li and Xinjie Lin and Jinnian Zhang and Xin-Sheng Chen and Yi Zhang and Kiyohiro Nakayama and Zhengyang Geng and Houwen Peng and Han Hu and Shi-min Hu},
booktitle={The Thirty-ninth Annual Conference on Neural Information Processing Systems Datasets and Benchmarks Track},
year={2025},
url={https://openreview.net/forum?id=n6GVwmGvZ0}
}

@inproceedings{
lu2024mathvista,
title={MathVista: Evaluating Mathematical Reasoning of Foundation Models in Visual Contexts},
author={Pan Lu and Hritik Bansal and Tony Xia and Jiacheng Liu and Chunyuan Li and Hannaneh Hajishirzi and Hao Cheng and Kai-Wei Chang and Michel Galley and Jianfeng Gao},
booktitle={The Twelfth International Conference on Learning Representations},
year={2024},
url={https://openreview.net/forum?id=KUNzEQMWU7}
}

@article{comanici2025gemini,
  title={Gemini 2.5: Pushing the frontier with advanced reasoning, multimodality, long context, and next generation agentic capabilities},
  author={Comanici, Gheorghe and Bieber, Eric and Schaekermann, Mike and Pasupat, Ice and Sachdeva, Noveen and Dhillon, Inderjit and Blistein, Marcel and Ram, Ori and Zhang, Dan and Rosen, Evan and others},
  journal={arXiv preprint arXiv:2507.06261},
  year={2025}
}

@misc{google2025gemini3,
  author = {{Google DeepMind}},
  title = {A New Era of Intelligence with {Gemini 3}},
  howpublished = {\url{https://blog.google/products-and-platforms/products/gemini/gemini-3/}},
  year = {2025},
  month = {November},
  note = {Official Blog. Accessed: 2026-01-28}
}

@misc{openai_gpt5_2025,
  title = {Introducing GPT-5},
  author = {OpenAI},
  year = {2025},
  howpublished = {\url{https://openai.com/index/introducing-gpt-5}}
}

@article{li2024llava,
  title={Llava-onevision: Easy visual task transfer},
  author={Li, Bo and Zhang, Yuanhan and Guo, Dong and Zhang, Renrui and Li, Feng and Zhang, Hao and Zhang, Kaichen and Zhang, Peiyuan and Li, Yanwei and Liu, Ziwei and others},
  journal={arXiv preprint arXiv:2408.03326},
  year={2024}
}

@misc{openai2024gpt4o,
  author = {OpenAI},
  title = {{GPT-4o System Card}},
  institution = {OpenAI},
  year = {2024},
  month = {August},
  url = {https://openai.com/index/gpt-4o-system-card/},
  note = {Technical Report}
}

@article{yang2025qwen3,
  title={Qwen3 technical report},
  author={Yang, An and Li, Anfeng and Yang, Baosong and Zhang, Beichen and Hui, Binyuan and Zheng, Bo and Yu, Bowen and Gao, Chang and Huang, Chengen and Lv, Chenxu and others},
  journal={arXiv preprint arXiv:2505.09388},
  year={2025}
}

@article{zhu2025internvl3,
  title={Internvl3: Exploring advanced training and test-time recipes for open-source multimodal models},
  author={Zhu, Jinguo and Wang, Weiyun and Chen, Zhe and Liu, Zhaoyang and Ye, Shenglong and Gu, Lixin and Tian, Hao and Duan, Yuchen and Su, Weijie and Shao, Jie and others},
  journal={arXiv preprint arXiv:2504.10479},
  year={2025}
}

@article{sun2024beats,
  title={Beats: Optimizing llm mathematical capabilities with backverify and adaptive disambiguate based efficient tree search},
  author={Sun, Linzhuang and Liang, Hao and Wei, Jingxuan and Yu, Bihui and He, Conghui and Zhou, Zenan and Zhang, Wentao},
  journal={arXiv preprint arXiv:2409.17972},
  year={2024}
}

\clearpage

\beginappendix

\begin{center}
    {\LARGE \textbf{Appendix Contents}}
\end{center}
\vspace{2em}

\begin{itemize}[leftmargin=1.5em, label={}]
    \setlength{\itemsep}{0.4em}
    
    % \item \textbf{\hyperref[proof:proof]{A. Theoretical Proofs}} \dotfill \pageref{proof:proof}
    % \begin{itemize}[leftmargin=1.5em, label={}]
    %     \item \hyperref[the:correction]{A.1. Proof of Theorem \ref{the:correction}} \dotfill \pageref{proof:correction}
    %     \item \hyperref[the:visual_grounding]{A.2. Proof of Theorem \ref{the:visual_grounding}} \dotfill \pageref{proof:Convergence_Analysis}
    % \end{itemize}
    
    \vspace{0.8em}
    \item \textbf{\hyperref[sec:more_exp]{A. Implementation Details}} \dotfill \pageref{sec:more_exp}
    \begin{itemize}[leftmargin=1.5em, label={}]
        \item \hyperref[sec:mc]{A.1. Model Configurations} \dotfill \pageref{sec:mc}
        \item \hyperref[sec:rsr]{A.2. Reasoning Substrate and Rendering} \dotfill \pageref{sec:rsr}
    \end{itemize}

    \vspace{0.8em}
    \item \textbf{\hyperref[sec:siglip]{B. SigLip Score on VCode Benchmark}} \dotfill \pageref{sec:siglip}

    \vspace{0.8em}
    \item \textbf{\hyperref[sec:related_work]{C. Detailed Related Work}} \dotfill \pageref{sec:related_work}

    \vspace{0.8em}
    \item \textbf{\hyperref[sec:benchmark_intro]{D. Detailed Benchmark Introduction}} \dotfill \pageref{sec:benchmark_intro}
    \begin{itemize}[leftmargin=1.5em, label={}]
        \item \hyperref[sec:vcodebench]{D.1. VCode} \dotfill \pageref{sec:vcodebench}
        \item \hyperref[sec:rbenchbench]{D.2. RBench-V} \dotfill \pageref{sec:rbenchbench}
        \item \hyperref[sec:mathvistabench]{D.3. MathVista} \dotfill \pageref{sec:mathvistabench}
    \end{itemize}
    
    \vspace{0.8em}
    \item \textbf{\hyperref[sec: qualitative_case_studies]{E. Qualitative Case Studies}} \dotfill \pageref{sec: qualitative_case_studies}
    
    \vspace{0.8em}
    \item \textbf{\hyperref[sec:detailed_case_ana]{F. Detailed Case Analysis}} \dotfill \pageref{sec:detailed_case_ana}
    \begin{itemize}[leftmargin=1.5em, label={}]
        \item \hyperref[sec:vcodecase]{F.1. VCode Cases} \dotfill \pageref{sec:vcodecase}
        \begin{itemize}[leftmargin=1.5em, label={}]
            \item \hyperref[sec:imagecase]{F.1.1. Final Image Rendered by SVG} \dotfill \pageref{sec:imagecase}
            \item \hyperref[sec:traj]{F.1.2. CRUD-Action Trajectory of Canvas-CoT} \dotfill \pageref{sec:traj}
        \end{itemize}
        \item \hyperref[sec:rbenchcase]{F.2. RBench-V Cases} \dotfill \pageref{sec:rbenchcase}
    \end{itemize}

    \vspace{0.8em}
    \item \textbf{\hyperref[sec:prompts]{G. Prompts}} \dotfill \pageref{sec:prompts}
    % \begin{itemize}[leftmargin=1.5em, label={}]
    %     \item \hyperref[sec:vcodeprompts]{G.1. VCode Prompts} \dotfill \pageref{sec:vcodeprompts}
    %     \item \hyperref[sec:rbenchprompts]{G.2. RBench-V \& MathVista Prompts} \dotfill \pageref{sec:rbenchprompts}
    % \end{itemize}

    % ... 以此类推添加其他项
\end{itemize}

\clearpage
\section{Implementation Details}
\label{sec:more_exp}

\subsection{Model Configurations}
\label{sec:mc}
We evaluate Canvas-CoT using two primary Multimodal Large Language Model (MLLM) backbones: GPT-5 and Gemini 2.5 Pro. To ensure reproducibility and minimize variance in the reasoning trajectory, we set the sampling temperature to 0 for all distinct inference runs. The framework is model-agnostic, interacting with the backend purely through standard chat completion APIs.

\subsection{Reasoning Substrate and Rendering} 
\label{sec:rsr}
The Reasoning Substrate $S$ is implemented as a lightweight HTML5 environment managed by a headless browser engine. 
\begin{itemize} 
    \item Initialization: At timestep t=0, the substrate is initialized with a minimal HTML boilerplate containing an empty <body> tag.
    \item Parsing Layer: We implement a deterministic parsing function that validates raw HTML/SVG fragments generated by the model. This layer filters malformed tags and ensures that all operations (Insert, Replace, Modify) result in a valid, renderable DOM tree before state transition. 
    \item Rendering: For the visual critique loop, the headless browser renders the current DOM state into a high-resolution raster image (Vstate), which is then downsampled to match the resolution of the original input image (Vorig) for pixel-aligned comparison. 
\end{itemize}

\section{SigLip Score on VCode Benchmark}
\label{sec:siglip}

To further validate the quality of the generated SVG code beyond structural correctness, we evaluate the semantic alignment between the rendered outputs and the ground truth visual concepts using the SigLip metric. Figure 6 presents a comparative analysis of Gemini 2.5 Pro across three settings: standard Chain-of-Thought (CoT), Iterative Reflection, and our proposed Canvas-CoT.

The most significant performance gain is observed on the CV-Bench subset, where Canvas-CoT achieves a SigLip score of 72.28, representing a substantial +8.1\% improvement over the CoT baseline (66.88) and outperforming Iterative Reflection (69.50).

On the MM-Vet subset, we observe a notable pathology in the text-only Iterative Reflection baseline, where performance actually degrades compared to standard CoT (68.91 → 68.11). This phenomenon aligns with the "hallucination snowballing" effect discussed in Section 1, where purely verbal self-correction introduces noise. In contrast, Canvas-CoT reverses this trend, achieving a score of 69.74 (+1.2\%). This demonstrates that the Rendering-Critique loop acts as a hard constraint validator, preventing the semantic drift common in ungrounded reasoning chains.

% \begin{figure*}[t]
%     \centering
%     \includegraphics[width=1.0\textwidth]{imags/pdf/appendix/svg_token.pdf}
%     \caption{Trade off between token expenses and accuracy on VCode benchmark.}
%     \label{fig:svg_token}
% \end{figure*}

\begin{figure}[t]
    \centering
    \includegraphics[width=0.47\textwidth]{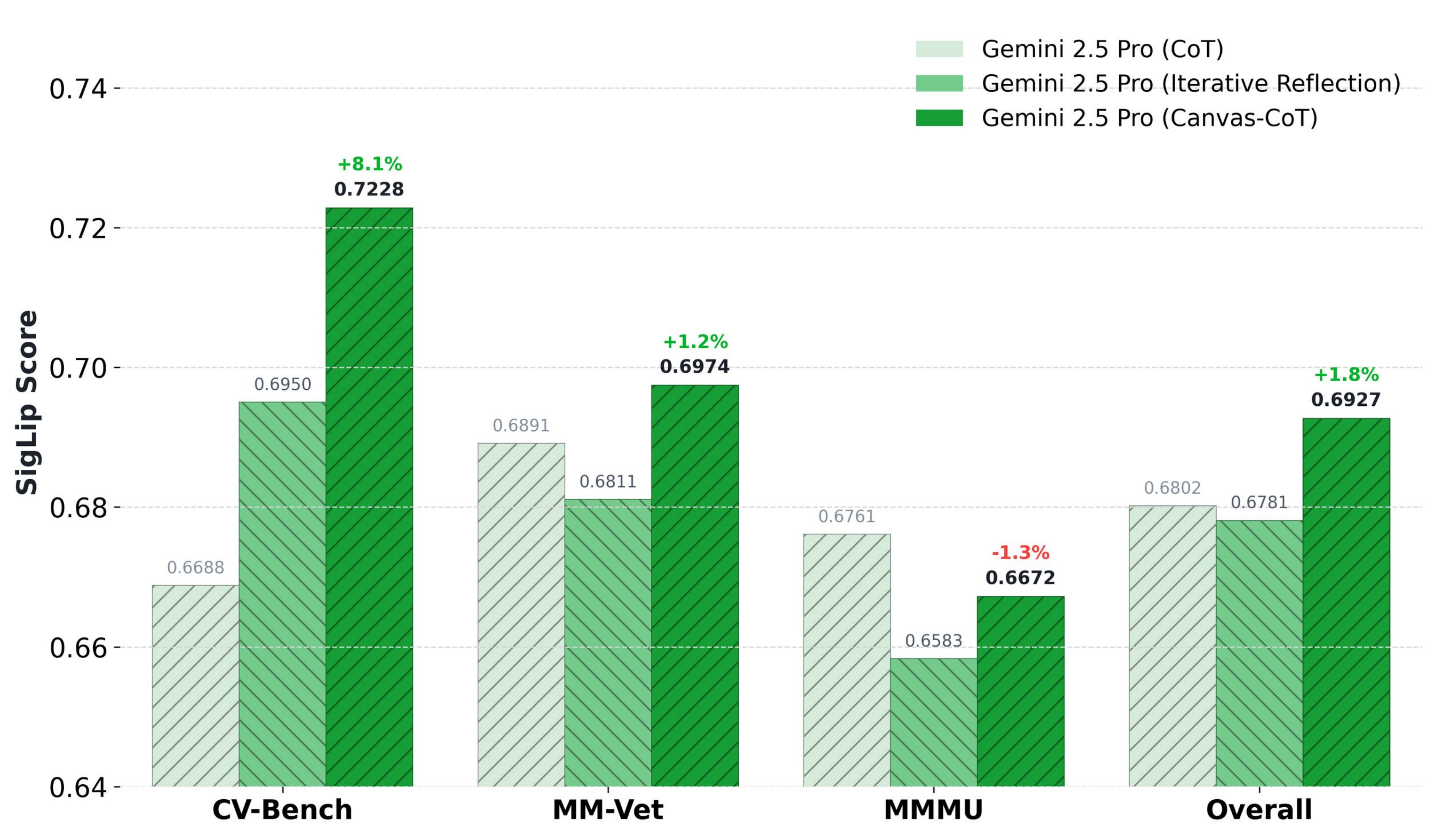}
    \caption{SigLip Score of Gemini on VCode benchmark.}
    \label{fig:siglip}
\end{figure}

% \section{Analysis of SVG Code Token Efficiency and Generation Quality}

% To further evaluate the computational efficiency of our proposed method, we analyze the trade-off between the average number of generated SVG tokens and the resulting Score.

\section{Detailed Related Work}
\label{sec:related_work}

The evolution of MLLM reasoning can be characterized by two orthogonal dimensions: the analyzability of the reasoning process (explicit vs. latent) and the topology of the reasoning substrate (append-only vs. structurally addressable).

\textbf{Linear Explicit Substrates.} Foundational paradigms, established by Chain-of-Thought \cite{wei2022chain} and extended by search-based methods like Tree-of-Thoughts \cite{yao2023tree} and Graph-of-Thoughts \cite{besta2024graph}, rely on explicit natural language tokens. A separate lineage introduces structure through executable code (PoT \cite{chen2022program}), tool-calling (ReAct \cite{yao2022react}), or iterative symbolic refinement (VCode \cite{lin2025vcode}). While these methods move toward a "System 2" \cite{xiang2025towards} cognitive mode by enabling deliberate planning, they remain restricted by an append-only and immutable substrate. Because the reasoning trace is treated as a monolithic linear log rather than a structurally addressable state, correcting a local error requires regenerating the entire downstream context. This imposes a prohibitive serialization tax as trajectories scale in the test-time compute era \cite{DBLP:journals/corr/abs-2501-12948, sui2025stop, yang2025longlive}.

\textbf{Linear Latent Substrates.} To improve inference density and bypass the verbosity of natural language, latent-space frameworks like Coconut \cite{hao2024training}, CODI \cite{shen2025codi}, and CoLaR \cite{tan2025think} compress reasoning paths into continuous vectors.However, this density sacrifices analyzability, transforming internal world model states into an opaque latent space \cite{yue2025hybrid, shen2025efficient} where states cannot be explicitly traced or intervened upon.

Canvas-CoT replaces linear, append-only reasoning with a structural, white-box DOM substrate. By decoupling reasoning logic from state persistence and employing ID-addressable nodes, the framework enables atomic CRUD operations. This shift significantly reduces the token generation overhead required for error correction, allowing precise, in-place updates without downstream context disruption. Finally, a Rendering-Critique loop provides grounded feedback, enabling the model to perform autonomous self-correction.

\section{Detailed Benchmark Introduction}
\label{sec:benchmark_intro}

\subsection{VCode}
\label{sec:vcodebench}
The VCode benchmark consists of 464 image-question pairs across three domains: general common sense (MM-Vet), professional disciplines (MMMU), and vision-centric perception (CV-Bench). It constructs tasks of varying dimensions and difficulty levels to evaluate a model's capability in transforming images into SVG code. VCode employs two primary evaluation metrics: the SigLIP score and CodeVQA. These metrics assess the semantic consistency between the rendered SVG and the original image, and the accuracy of the model's answers based on the rendered SVG, respectively. 

\subsection{RBench-V}
\label{sec:rbenchbench}
The RBench-V benchmark focuses on assessing the reasoning robustness of multimodal models in visual tasks. The dataset contains 803 questions covering core dimensions such as spatial relations, attribute comparison, and logical reasoning. Its evaluation metrics go beyond final answer accuracy by introducing a reasoning consistency indicator. By testing different linguistic expressions of the same visual logic, it evaluates whether the model’s reasoning remains stable under perturbations or varying prompts, thereby effectively identifying potential "hallucinations" in the model. 

\subsection{MathVista}
\label{sec:mathvistabench}
MathVista is a comprehensive benchmark designed to evaluate mathematical reasoning capabilities within visual contexts and is currently regarded as one of the most authoritative benchmarks in this field. The dataset integrates 6,141 samples from 28 existing benchmarks along with three newly constructed datasets (IQTest, FunctionQA, and PaperQA). It covers seven categories of mathematical reasoning tasks, including elementary mathematics, competition-level math, and scientific chart understanding (e.g., algebraic reasoning, geometric problems, and statistical chart analysis). The primary evaluation metric is Accuracy, with fine-grained scoring based on task difficulty and knowledge domains, providing a comprehensive measurement of a model's overall proficiency in solving complex visual mathematical problems.

\section{Qualitative Case Studies}
\label{sec: qualitative_case_studies}

\begin{figure*}[t]
    \centering
    \includegraphics[width=1.0\textwidth]{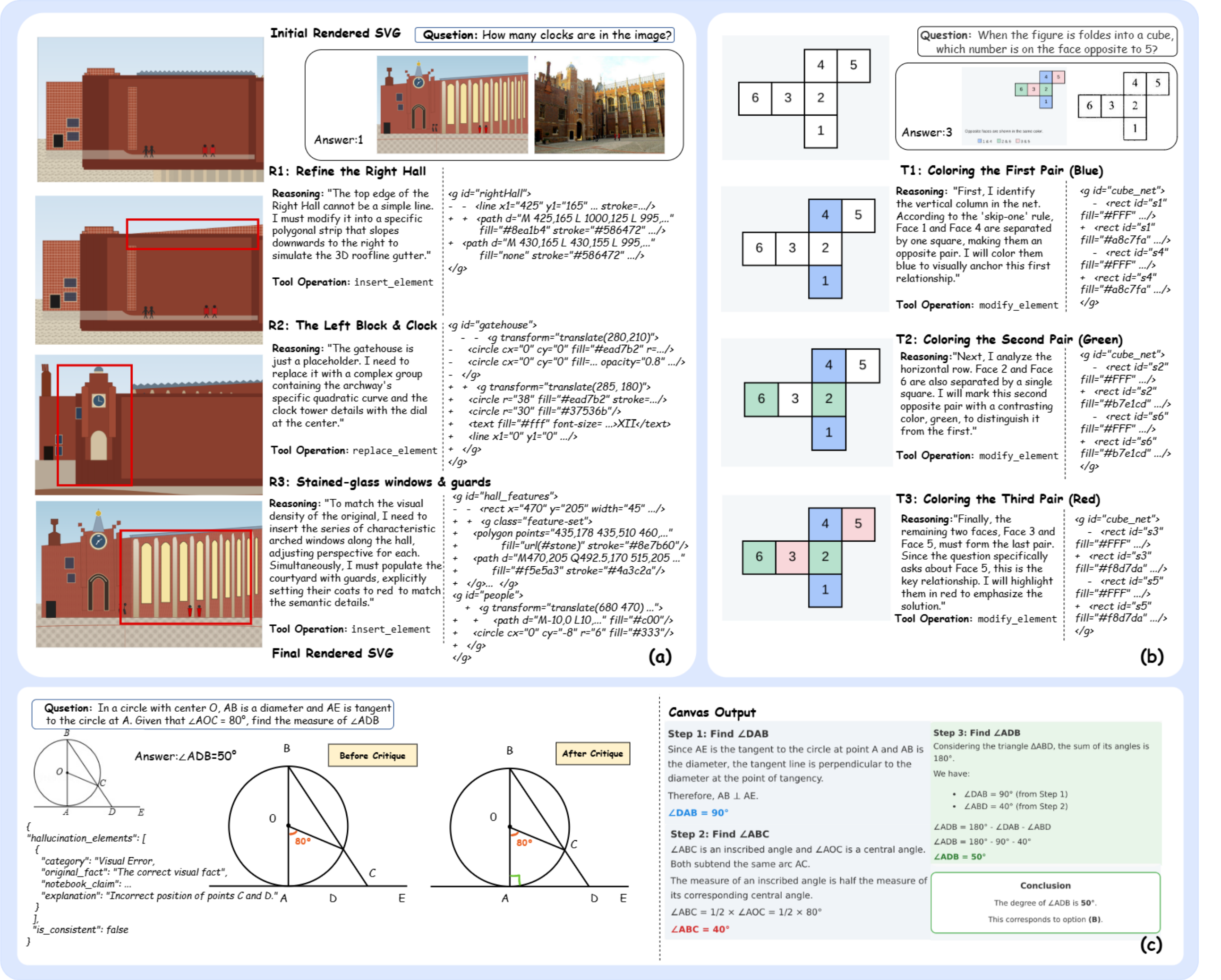}
    \caption{Example demonstration of three core capabilities. (a): Precise Editing of Complex SVG. (b): Effective Representation of Spatial. (c): Grounded Reasoning via Canvas.}
    \label{fig:case}
\end{figure*}

% To visualize the epistemic difference, we compare failure modes in Figure X (select a case from RBench-V/MathVista).
To concretize the epistemic advantages of Canvas-of-Thought, we conduct a qualitative analysis of three representative reasoning modes illustrated in Figure~\ref{fig:case}. Each case demonstrates how our framework transcends the limitations of linear text generation by leveraging the structured, mutable, and grounded nature of the DOM substrate.

\paragraph{Precision in Structural Generation via ID-Addressable Reasoning.} 
As shown in Figure~\ref{fig:case}(a), in the Image-to-SVG task, our method excels at refining complex architectural geometries after an initial draft. Standard CoT and Iterative Reflection often struggle here, as correcting a local error such as a roofline's perspective or clock face details, typically requires regenerating the entire code block, increasing the risk of cascading hallucinations and raising token consumption. In contrast, Figure~\ref{fig:case}(a) shows Canvas-CoT leveraging ID-Addressable Reasoning by assigning unique identifiers to semantic components like \texttt{<g id="rightHall">} and \texttt{<g id="gatehouse">}, effectively creating a discrete memory bank. For instance, when upgrading the \texttt{gatehouse} from a rough placeholder to a detailed structure, the Solver Agent executes a targeted \texttt{replace\_element} operation solely on the \texttt{gatehouse} node to insert the specific dial geometry and arched masonry. Similarly, it applies precise \texttt{insert\_element} operations to the \texttt{rightHall} to correct the 3D roofline strip. This bypasses the ``append-only'' limitation, enabling direct indexing and editing of specific reasoning states without disturbing the global context.

\paragraph{Efficiency in Spatial Logic via Atomic State Mutability.} Figure~\ref{fig:case}(b) presents a Cube Net Folding problem, requiring the identification of opposite faces. Solving this via text-only CoT demands maintaining a high mental load to track spatial permutations, often leading to ``lost-in-the-middle'' errors. Canvas-CoT addresses this through Atomic State Mutability. As the agent identifies face pairs like faces 1 and 4, it does not merely output text; it executes \texttt{modify\_element} calls to update their visual attributes like \texttt{fill="blue"}. These $\mathcal{O}(1)$ atomic updates offload the cognitive burden onto the canvas. The visual state thus serves as a dynamic external working memory, where the logical validity of ``opposite faces'' is maintained through persistent state mutations rather than transient tokens, ensuring stability in multi-step spatial reasoning.

\paragraph{Error Correction via Visual-Physical Grounding.} Figure~\ref{fig:case}(c) depicts a geometry problem from MathVista requiring the calculation of $\angle ADB$. A common failure mode in text-based reasoning is visual alignment error, where the model retrieves correct geometric properties but fails to map them accurately to visual entities. Initially, the agent incorrectly plotted the positions of points $C$ and $D$, generating a visual state where their coordinate labels are swapped. While a text-only model might hallucinate a solution based on this erroneous topology, Canvas-CoT addresses this issue by introducing a rendering-critique loop. The critique module compares the rendered diagram against the original visual evidence, immediately flagging the misalignment of labels $C$ and $D$. This feedback triggers a corrective \texttt{modify\_element} operation to swap the coordinates. By grounding the reasoning process in a verifiable visual state, the model ensures geometric consistency, enabling the correct derivation of the angle properties.

%Appendix D
\section{Detailed Case Analysis}
\label{sec:detailed_case_ana}
\subsection{VCode Cases}
\label{sec:vcodecase}

In this section, we present a qualitative comparison on the VCode benchmark to demonstrate the efficacy of Canvas-CoT in structured image generation, MM-VET, MMMU, CV-Bench are this benchmark's subsets. In \ref{sec:imagecase}, we selected sample cases to compare Canvas-CoT against the same base models employing 5-round Iterative Reflection on several key points. Beyond final outputs rendered by SVG, we also illustrate the prompts applied for this task, and the CURD action trajectory of Canvas-CoT to achieve precise refinement of SVG structures in \ref{sec:traj}. The corresponding sequence of rendered images is also presented in Figure~\ref{fig:logcase}. 

In the following figures, "IR" denotes Iterative Reflection.

\clearpage
\subsubsection{Final Image Rendered by SVG}
\label{sec:imagecase}
% \paragraph{MMVET}

% %%%%%%%%%%%%%%%%%%v1_36%%%%%%%%%%%%%%%%%%%
% 居中且垂直居中的列类型
\newcolumntype{C}[1]{>{\centering\arraybackslash}m{#1}}

% 1. 绿色实心圆 (全对)
\newcommand{\cFull}{\textcolor{green!80!black}{\CIRCLE}} 
% 2. 黄色半实心圆 (半对)
\newcommand{\cHalf}{\textcolor{yellow!80!red}{\LEFTcircle}}
% 3. 红色空心圆 (错误)
\newcommand{\cEmpty}{\textcolor{red}{\Circle}}

\begin{table}[htbp]
\centering
\small
\resizebox{\textwidth}{!}{%
\begin{tabular}{C{3.2cm} C{3.2cm} C{3.2cm} C{3.2cm} C{3.2cm}}
\toprule

% --- line1 ---
\multicolumn{2}{c}{\textbf{Question \& Answer}} & 
\multicolumn{3}{c}{\textbf{Key Points}} \\ 
\cmidrule(r){1-2} 

\cmidrule(l){3-5}

\multicolumn{2}{c}{
    \begin{tabular}{l} 
        \textbf{Q:} What should we add in the third step? \\ 
        \textbf{A:} milk %Phytoplankton\texttt{<AND>}Seaweed 
    \end{tabular}  
} &
\multicolumn{3}{c}{
    \begin{tabular}{l}
    1. Third step item count: milk, whisk, bowl. \\
    2. Realism of object shape representation. \\
    3. Overall layout fidelity
    \end{tabular}
}\\

\midrule

% --- line2 ---
\textbf{Original Image} & 
\textbf{Gemini-2.5-pro (ours)} & 
\textbf{GPT-5 (ours)} & 
\textbf{Gemini-2.5-pro (IR)} & 
\textbf{GPT-5 (IR)} \\ 

% --- line3 ---
\includegraphics[width=3.1cm]{imags/pdf/appendix/v1_36a.pdf} & 
\includegraphics[width=3.1cm]{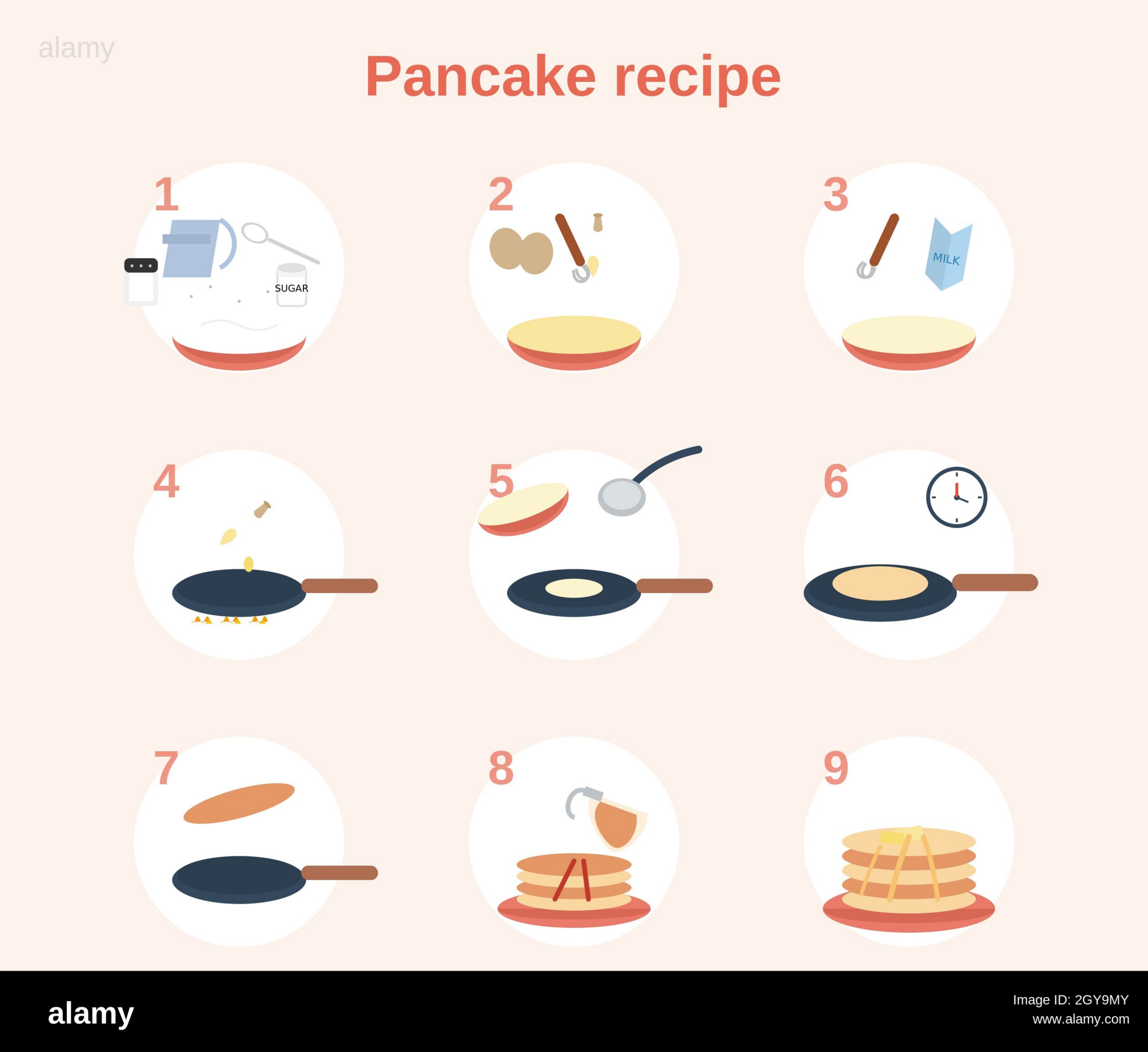} & 
\includegraphics[width=3.1cm]{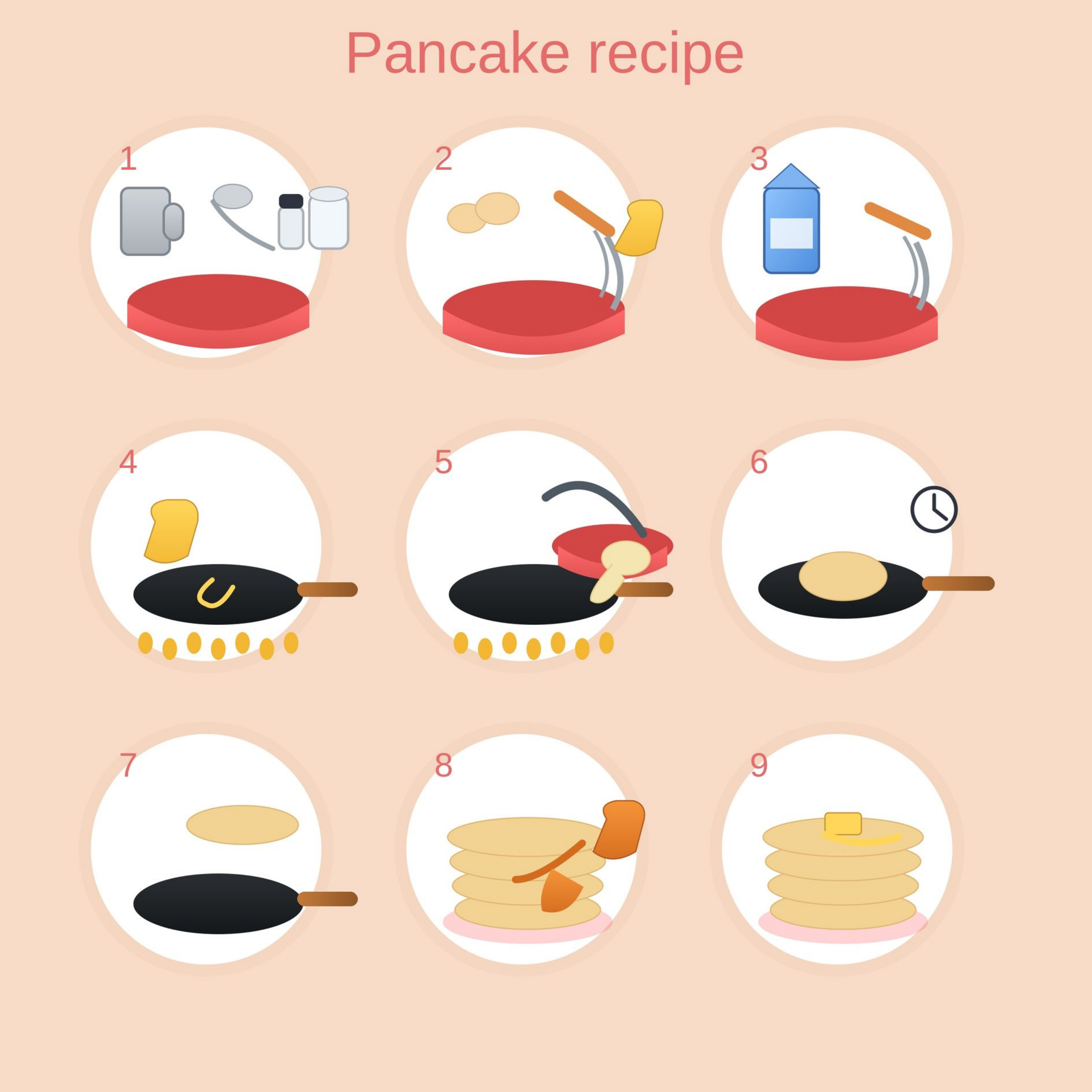} & 
\includegraphics[width=3.1cm]{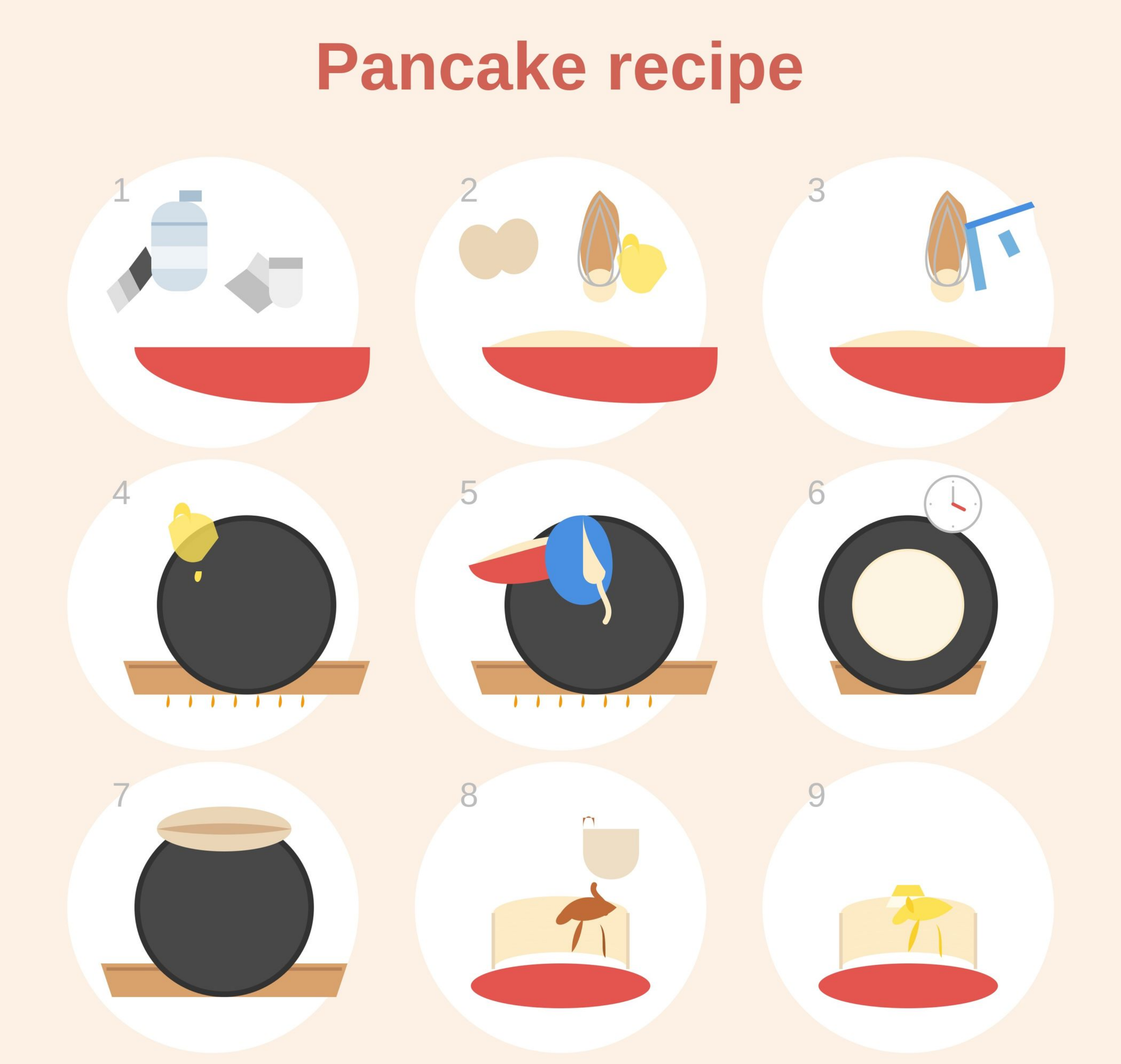} & 
\includegraphics[width=3.1cm]{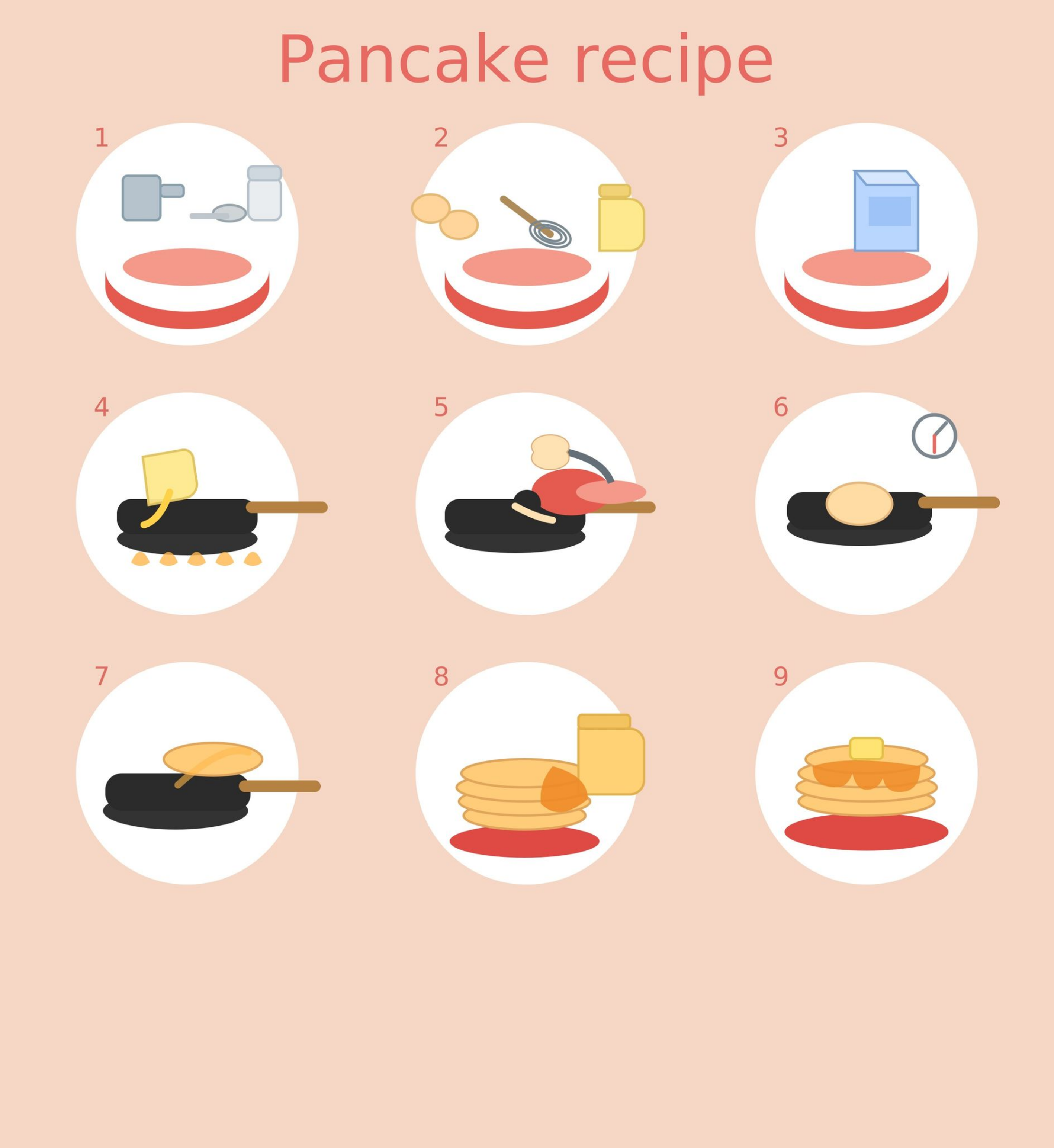} \\ 

% --- line4 ---
\cFull \quad \cFull \quad \cFull & 
\cFull \quad \cFull \quad \cFull & 
\cFull \quad \cFull \quad \cFull & 
\cFull \quad \cHalf \quad \cHalf & 
\cEmpty \quad \cHalf \quad \cHalf \\
\bottomrule
\end{tabular}
}
\vspace{0.5em}
\caption{Case v1\_36}

\end{table}

%%%%%%%%%%%%%%%%%%v1_184%%%%%%%%%%%%%%%%%%%
\begin{table}[htbp]
\centering
\small
\resizebox{\textwidth}{!}{%
\begin{tabular}{C{3.2cm} C{3.2cm} C{3.2cm} C{3.2cm} C{3.2cm}}
\toprule

% --- line1 ---
\multicolumn{3}{c}{\textbf{Question \& Answer}} & 
\multicolumn{2}{c}{\textbf{Key Points}} \\ 
\cmidrule(r){1-3} 
\cmidrule(l){4-5}

\multicolumn{3}{c}{
    \begin{tabular}{p{9.0cm}} 
        \textbf{Q:} Can you give a short introduction to this painting? \\ 
        \textbf{A:} The Arnolfini Portrait (or other titles) is a 1434 oil painting on oak panel by the Early Netherlandish painter Jan van Eyck. ... The painting was bought by the National Gallery in London in 1842. 
    \end{tabular}  
} &
\multicolumn{2}{c}{
    \begin{tabular}{l}
    1. Aspect ratio of the image. \\
    2. Fidelity of object shape restoration. \\
    3. Color fidelity.\\
    4. Image detail fidelity.
    \end{tabular} 
} \\

\midrule

% --- line2 ---
\textbf{Original Image} & 
\textbf{Gemini-2.5-pro (ours)} & 
\textbf{GPT-5 (ours)} & 
\textbf{Gemini-2.5-pro (IR)} & 
\textbf{GPT-5 (IR)} \\ 

% --- line3 ---
\includegraphics[height=3.1cm]{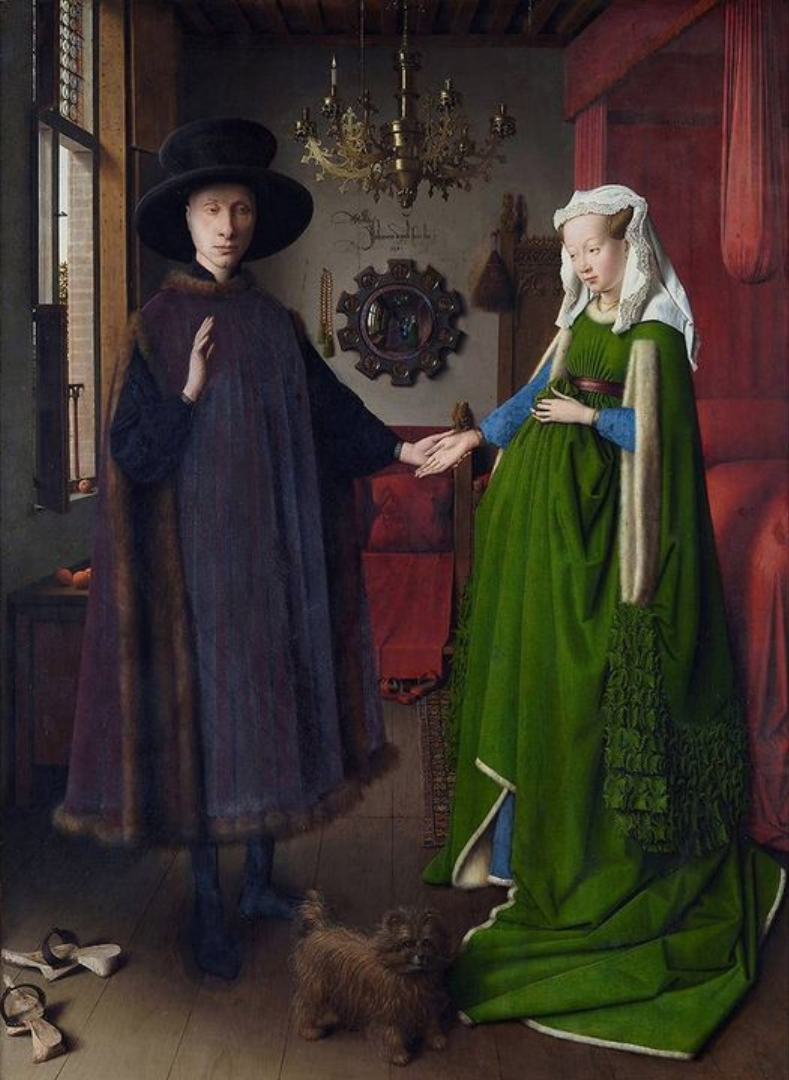} & 
\includegraphics[height=3.1cm]{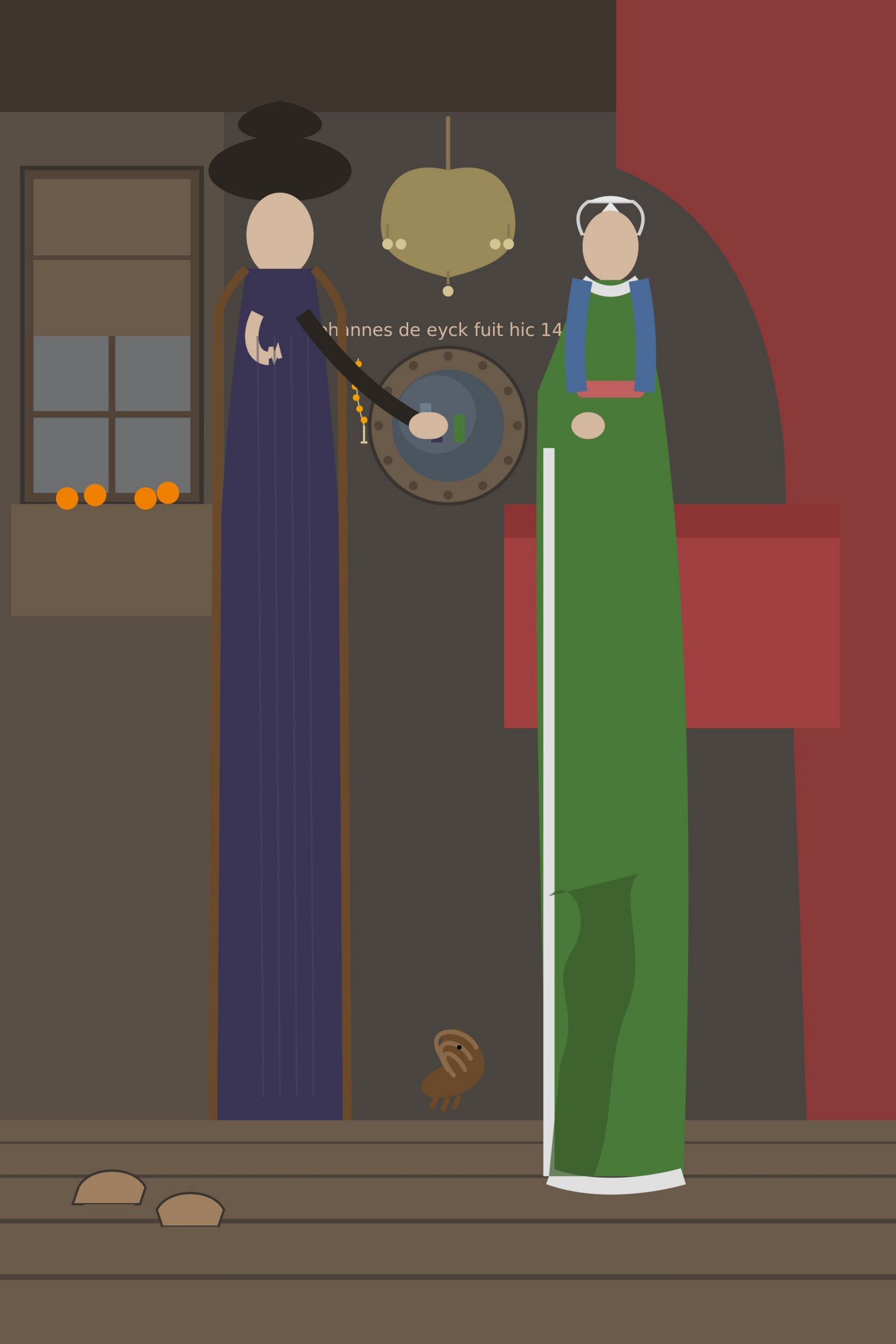} & 
\includegraphics[height=3.1cm]{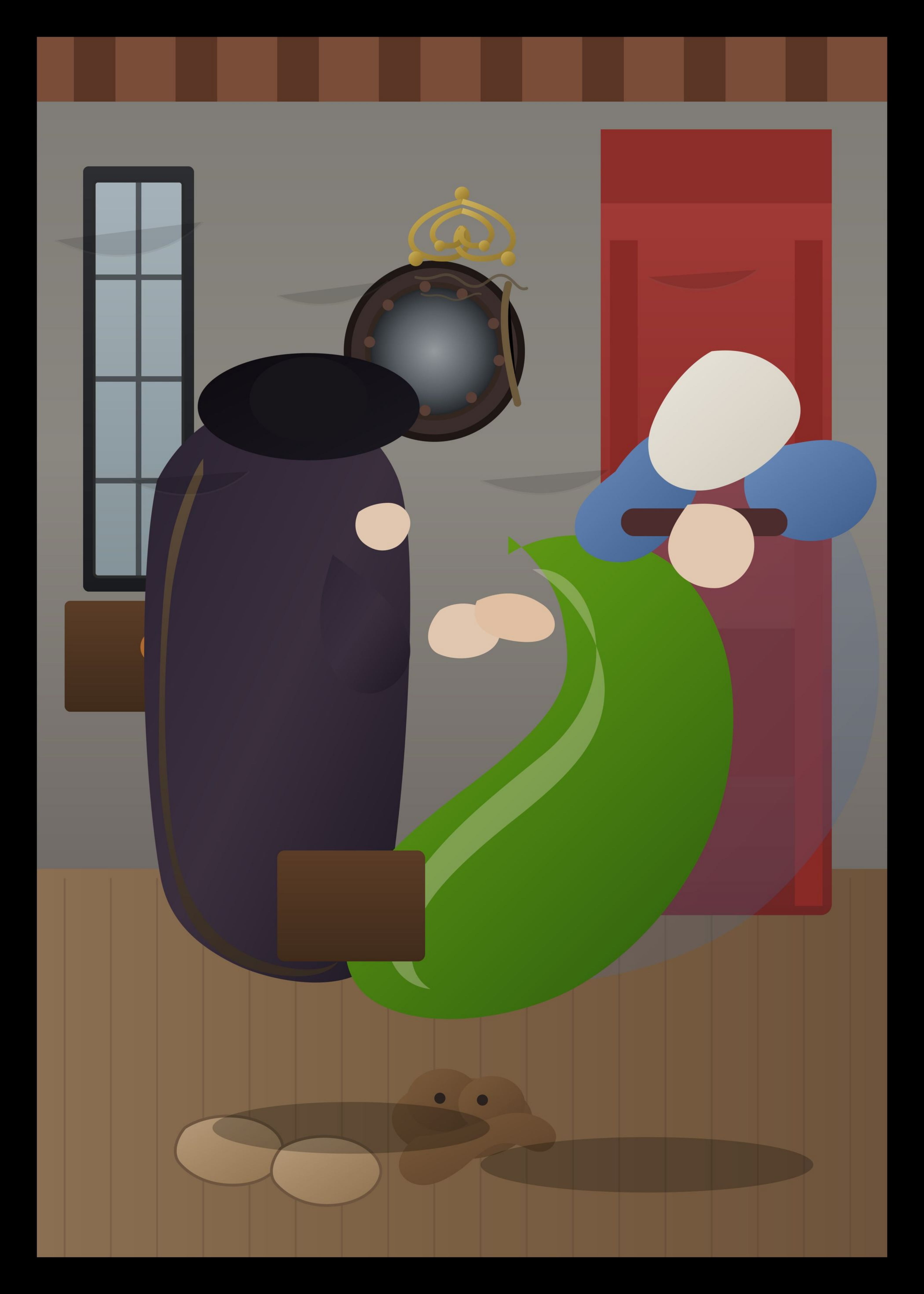} & 
\includegraphics[width=3.2cm]{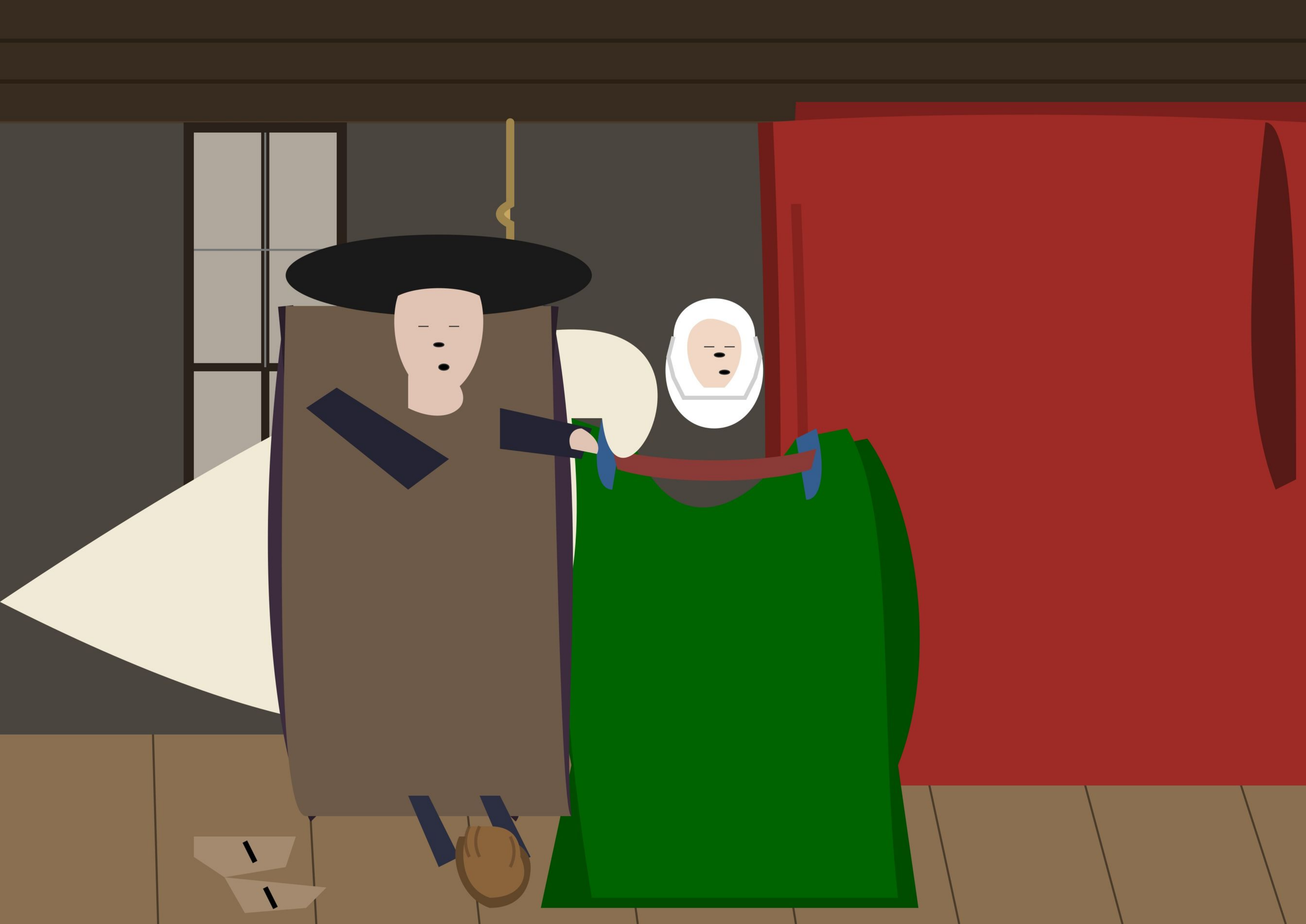} & 
\includegraphics[width=3.2cm]{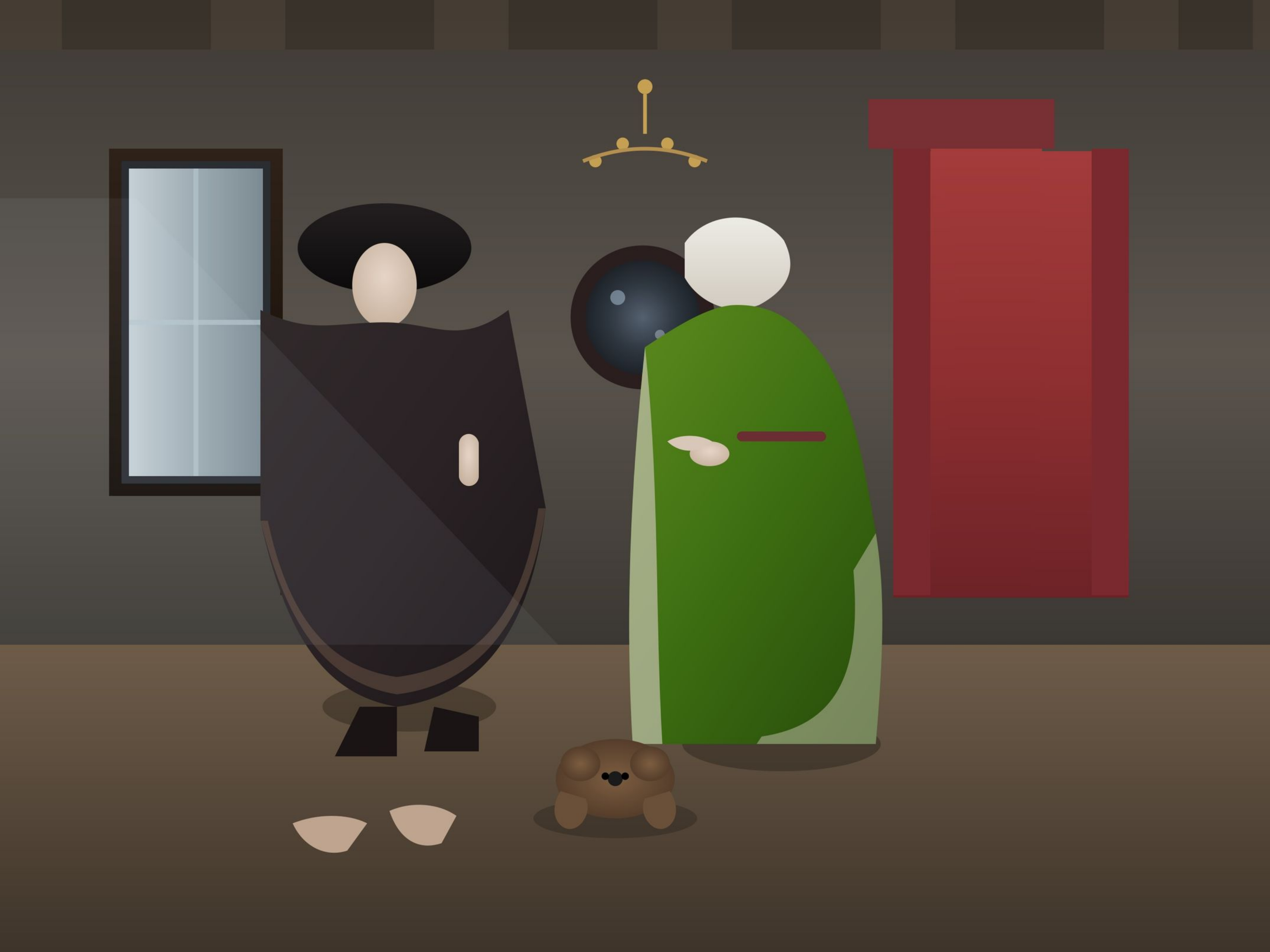} \\ 

% --- line4 ---
\cFull \quad \cFull \quad \cFull \quad \cFull & 
\cFull \quad \cFull \quad \cFull \quad \cHalf& 
\cFull \quad \cHalf \quad \cFull \quad \cHalf& 
\cEmpty \quad \cHalf \quad \cFull \quad \cEmpty& 
\cEmpty \quad \cHalf \quad \cFull \quad \cEmpty\\
\bottomrule
\end{tabular}
}
\vspace{0.5em}
\caption{Case v1\_184}
\end{table}

%%%%%%%%%%%%%%%%%%v1_23%%%%%%%%%%%%%%%%%%%
\begin{table}[!htbp]
\centering
\small
\resizebox{\textwidth}{!}{%
\begin{tabular}{C{3.2cm} C{3.2cm} C{3.2cm} C{3.2cm} C{3.2cm}}
\toprule

% --- line1 ---
% \textbf{ Original Image } & 
\multicolumn{2}{c}{\textbf{Question \& Answer}} & 
\multicolumn{3}{c}{\textbf{Key Points}} \\ 
\cmidrule(r){1-2} 
%\cmidrule(lr){2-3} 
\cmidrule(l){3-5}

\multicolumn{2}{c}{
    \begin{tabular}{l} 
        \textbf{Q:} What is the price for tomatoes? \\ 
        \textbf{A:} \text{eight}\texttt{<OR>}8.0 %Phytoplankton\texttt{<AND>}Seaweed 
    \end{tabular}  
} &
\multicolumn{3}{c}{
    \begin{tabular}{l}
    1. Whether the tomato is present.\\
    2. Relative positions between the tomato and its price tag.\\
    3. Scene restoration fidelity.\\
    4. Fruit detail restoration accuracy.
    \end{tabular} 
}\\

\midrule

% --- line2 ---
\textbf{Original Image} & 
\textbf{Gemini-2.5-pro (ours)} & 
\textbf{GPT-5 (ours)} & 
\textbf{Gemini-2.5-pro (IR)} & 
\textbf{GPT-5 (IR)} \\ 
% \cmidrule(r){1-1} 
% \cmidrule(lr){2-2} 
% \cmidrule(lr){3-3} 
% \cmidrule(lr){4-4} 
% \cmidrule(l){5-5}
%\addlinespace[3pt]

% --- line3 ---
\includegraphics[width=3.2cm]{imags/pdf/appendix/v1_23a.pdf} & 
\includegraphics[width=3.2cm]{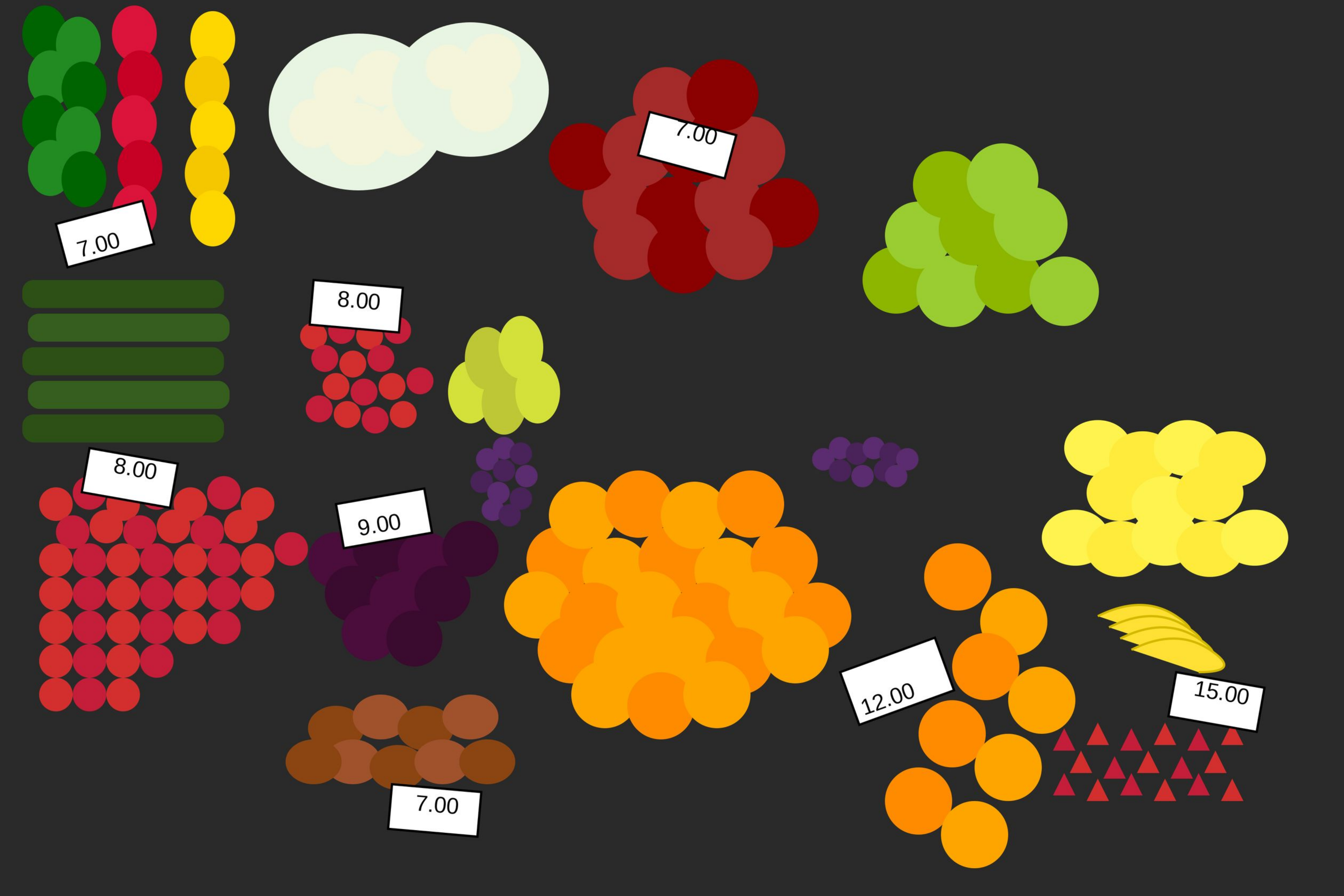} & 
\includegraphics[width=3.2cm]{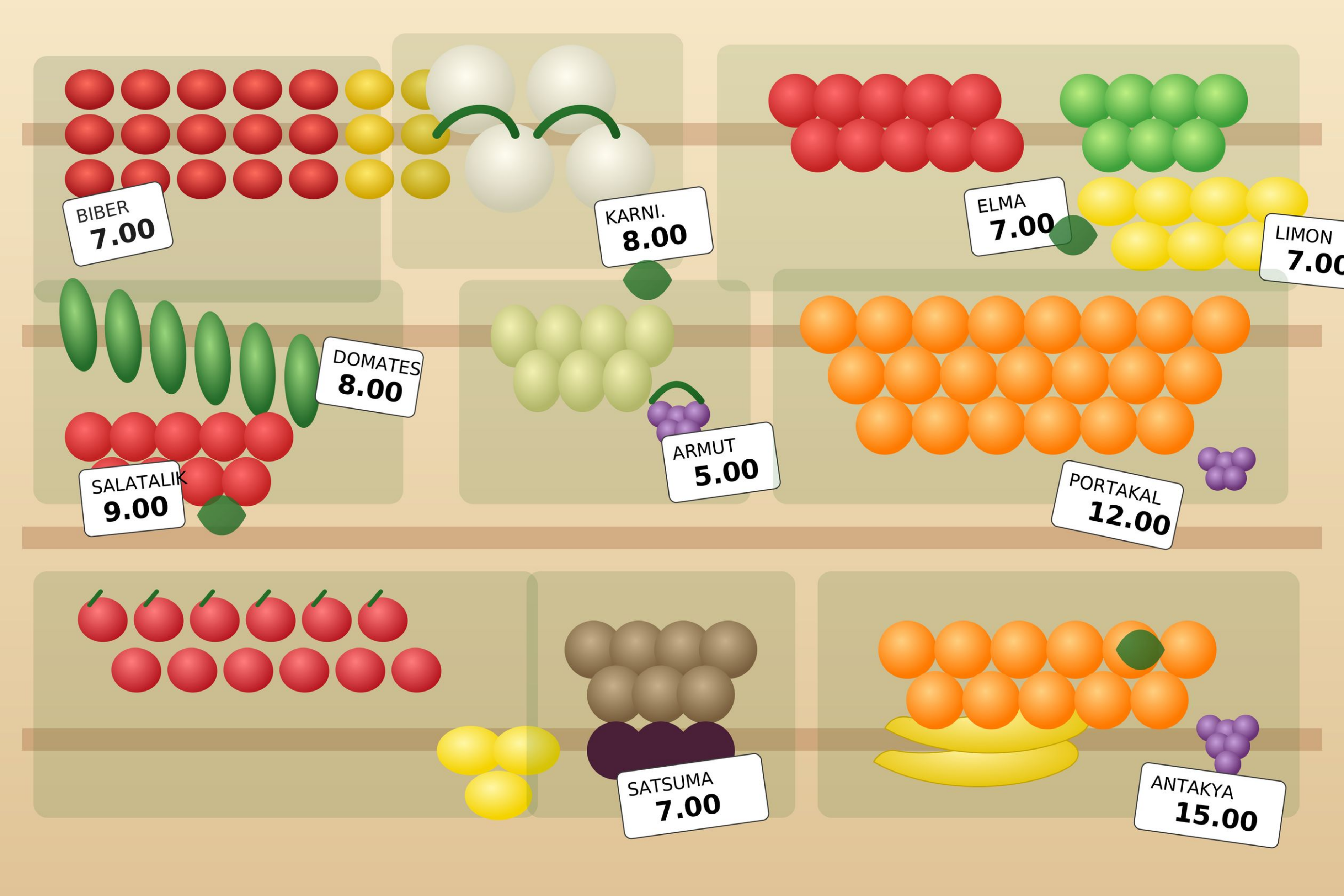} & 
\includegraphics[width=3.2cm]{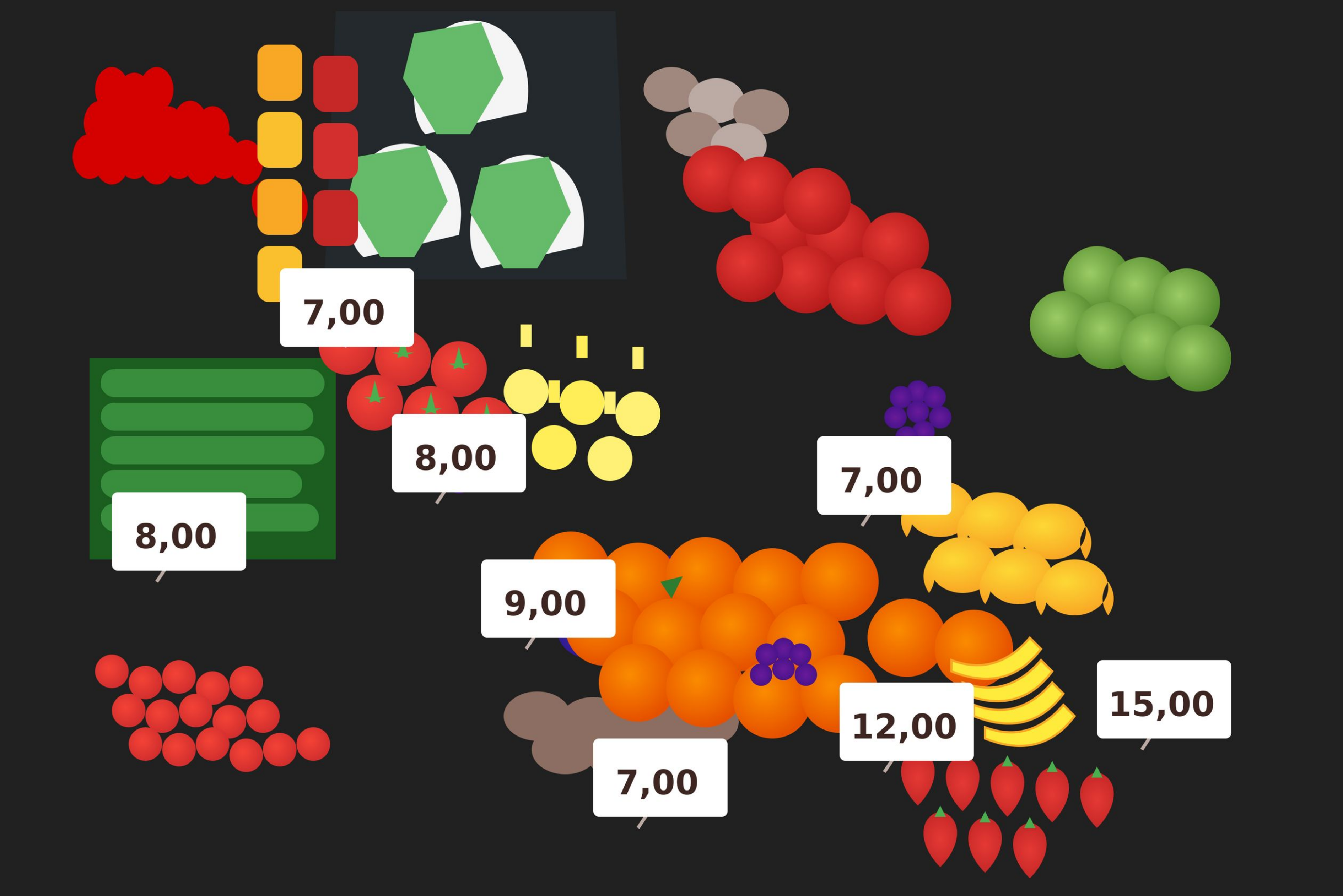} & 
\includegraphics[width=3.2cm]{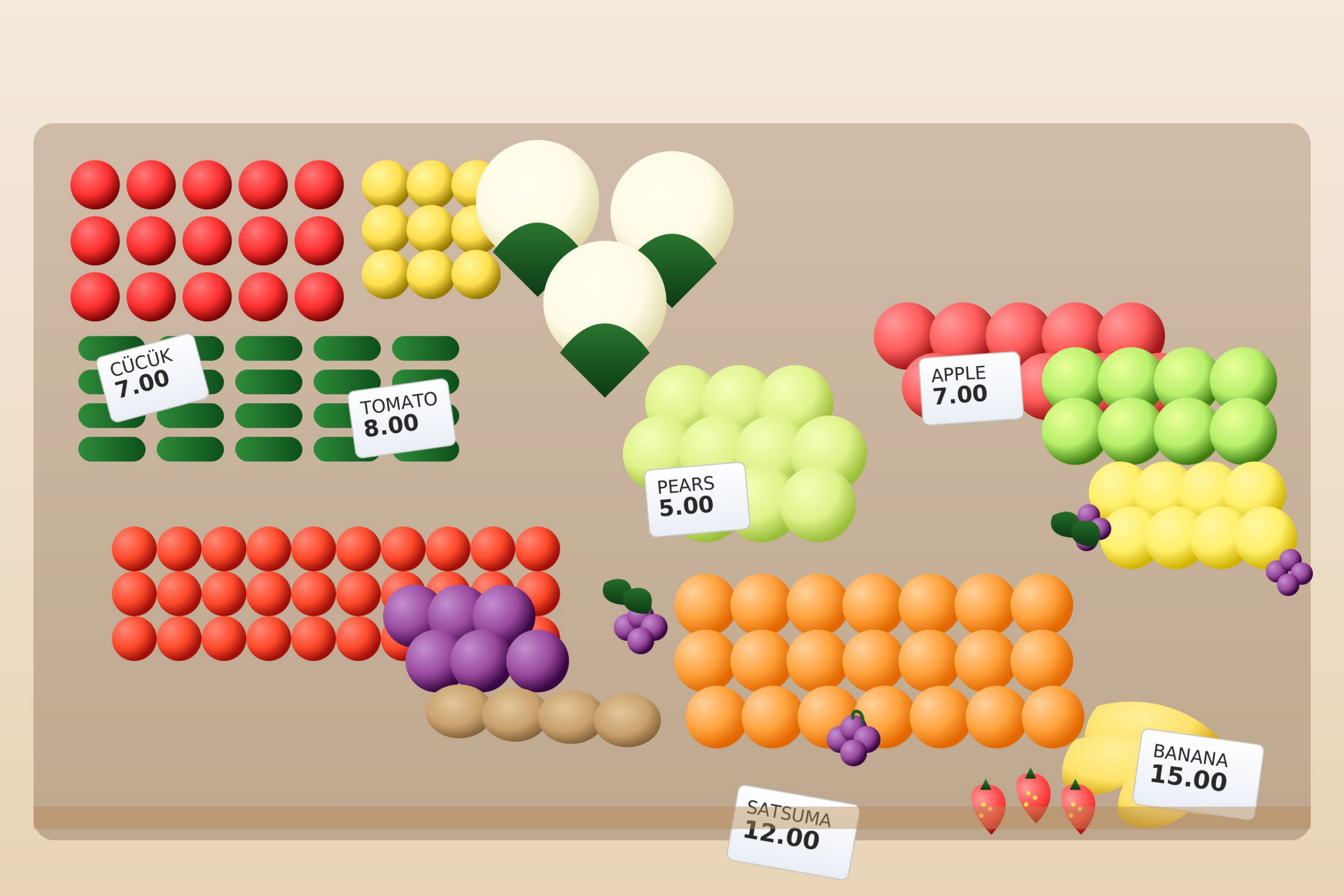} \\ 

% --- line4 ---
\cFull \quad \cFull \quad \cFull \quad \cFull & 
\cFull \quad \cFull \quad \cHalf \quad \cHalf& 
\cFull \quad \cEmpty \quad \cFull \quad \cHalf& 
\cFull \quad \cFull \quad \cHalf \quad \cHalf& 
\cEmpty \quad \cEmpty \quad \cHalf \quad \cHalf\\
\bottomrule
\end{tabular}
}
\vspace{0.5em}
\caption{Case v1\_23}

\end{table}

% \clearpage
% \paragraph{MMMU}
%%%%%%%%%%%%%%%%%%dev_Physics_5_img1%%%%%%%%%%%%%%%%%%%
\begin{table}[htbp]
\centering
\small
\resizebox{\textwidth}{!}{%
\begin{tabular}{C{3.2cm} C{3.2cm} C{3.2cm} C{3.2cm} C{3.2cm}}
\toprule

% --- line1 ---
% \textbf{ Original Image } & 
\multicolumn{3}{c}{\textbf{Question \& Answer}} & 
\multicolumn{2}{c}{\textbf{Key Points}} \\ 
\cmidrule(r){1-3} 
%\cmidrule(lr){2-3} 
\cmidrule(l){4-5}

\multicolumn{3}{c}{
    \begin{tabular}{p{10.0cm}} 
        \textbf{Q:} A battery, an ammeter, three resistors, and a switch are connected to form the simple circuit shown above. When the switch is closed what would happen to the potential difference across the 15 ohm resistor?(A) it would equal the potential difference across the 20 ohm resistor (B) it would be twice the potential difference across the 30 ohm resistor (C) it would equal the potential difference across the 30 ohm resistor (D) it would be half the potential difference across the 30 ohm resistor  \\ 
        \textbf{A:} (C)
    \end{tabular}  
} &
\multicolumn{2}{c}{
    \begin{tabular}{p{6.0cm}}
    1. Switch S open/closed status. \\
    2. Correct representation of fundamental circuit components. \\
    3. Clear annotation.
    \end{tabular} 
}\\

\midrule

% --- line2 ---
\textbf{Original Image} & 
\textbf{Gemini-2.5-pro (ours)} & 
\textbf{GPT-5 (ours)} & 
\textbf{Gemini-2.5-pro (IR)} & 
\textbf{GPT-5 (IR)} \\ 
% \cmidrule(r){1-1} 
% \cmidrule(lr){2-2} 
% \cmidrule(lr){3-3} 
% \cmidrule(lr){4-4} 
% \cmidrule(l){5-5}
%\addlinespace[3pt]

% --- line3 ---
\includegraphics[width=3.2cm]{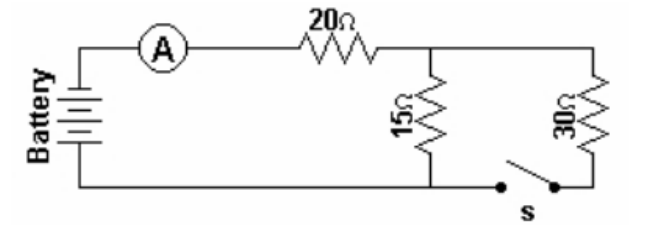} & 
\includegraphics[width=3.2cm]{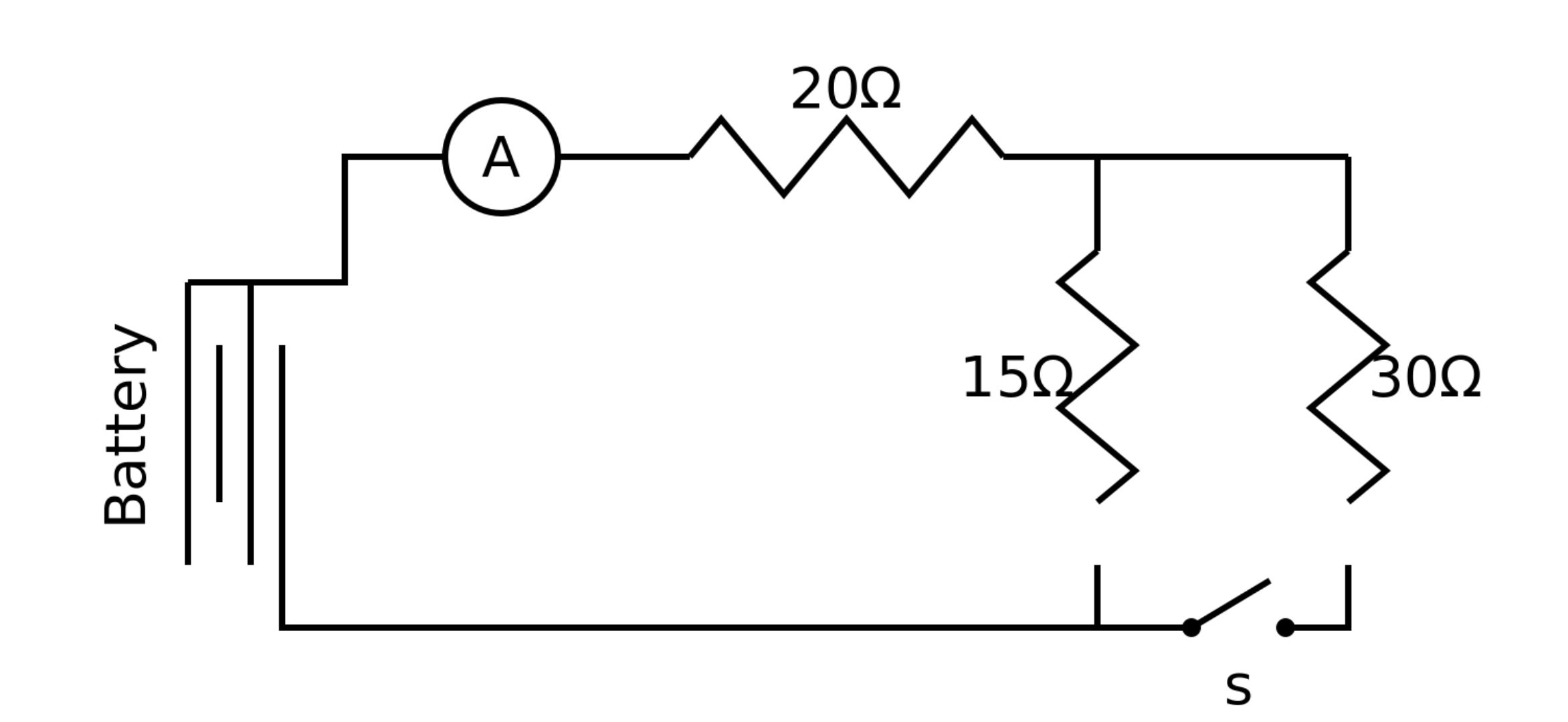} & 
\includegraphics[width=3.2cm]{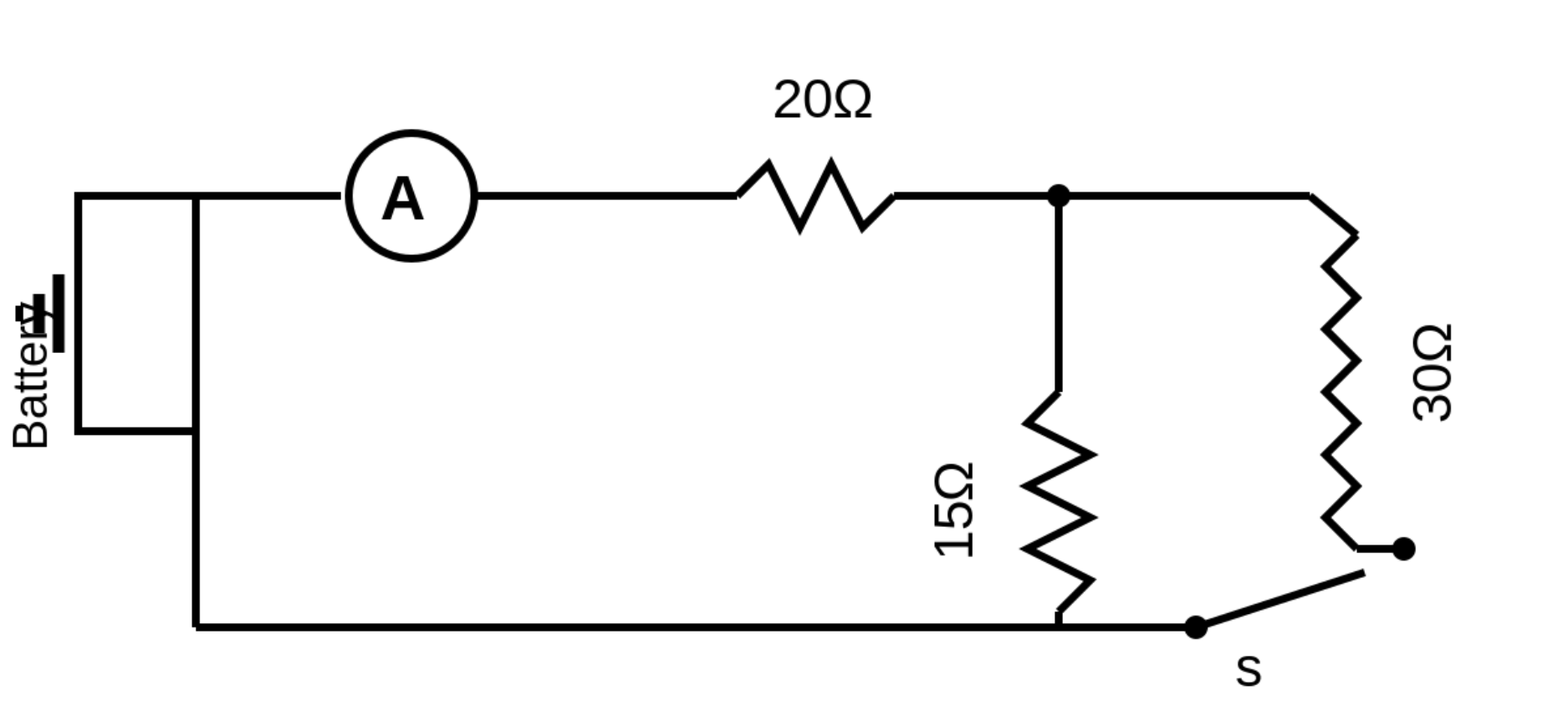} & 
\includegraphics[width=3.2cm]{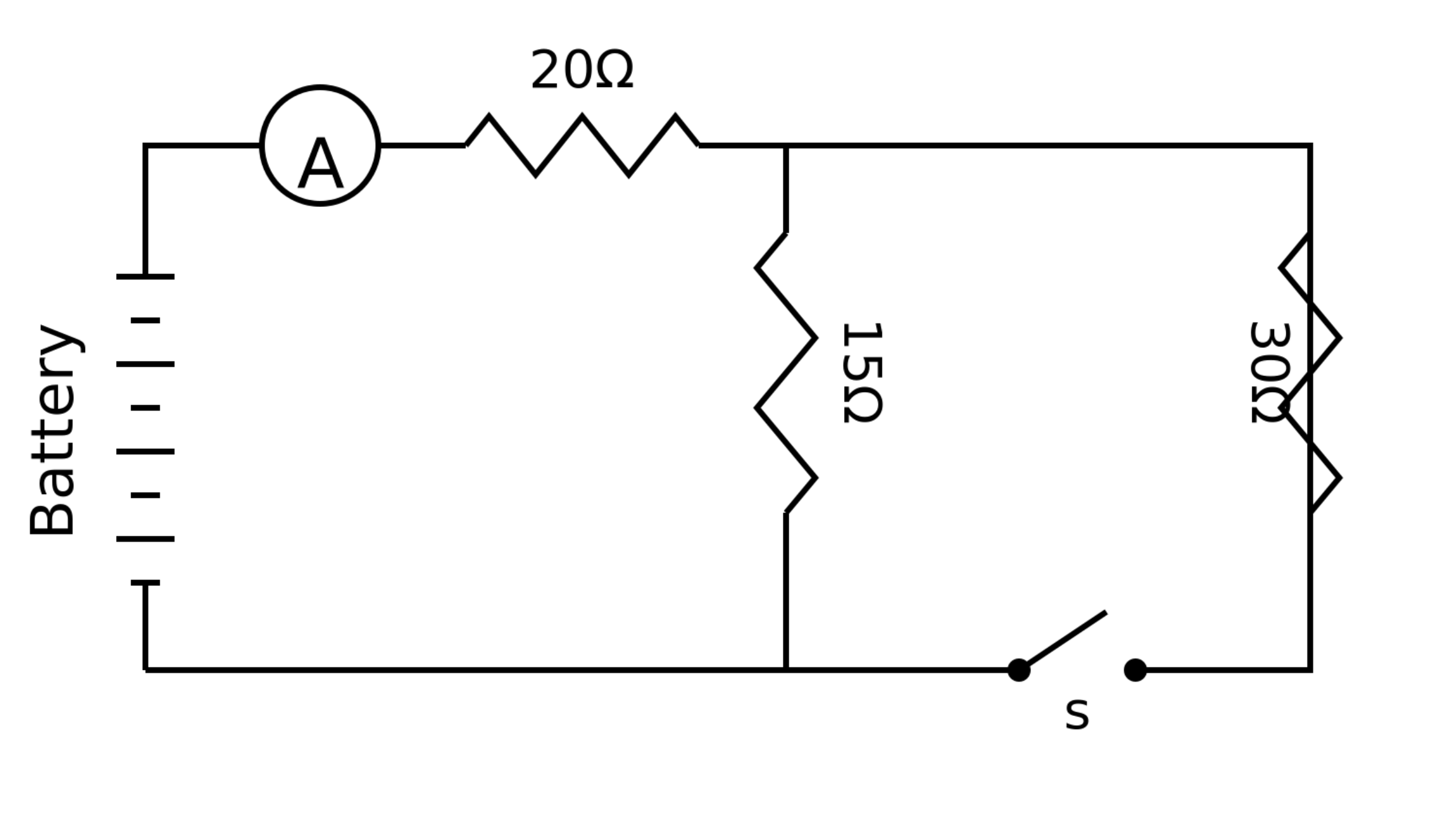} & 
\includegraphics[width=3.2cm]{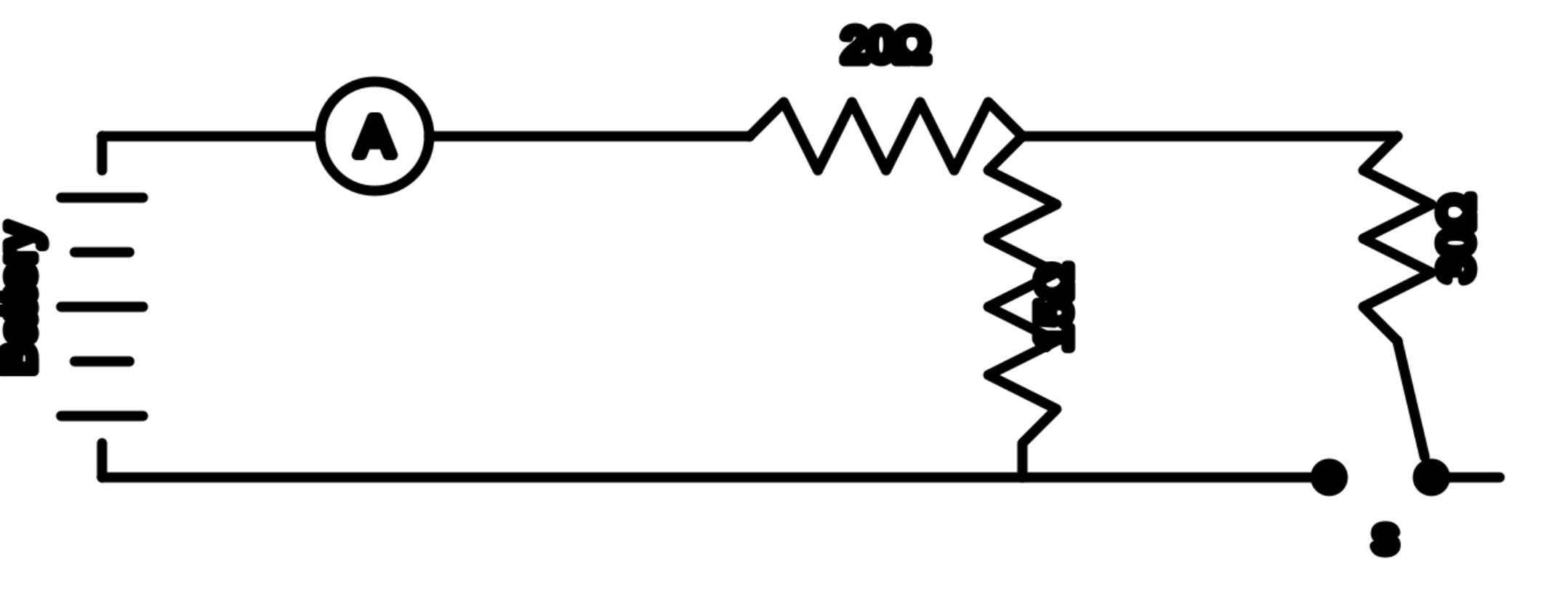} \\ 

% --- line4 ---
\cFull \quad \cFull \quad \cFull & 
\cFull \quad \cFull \quad \cFull & 
\cFull \quad \cHalf \quad \cFull & 
\cFull \quad \cHalf \quad \cFull & 
\cEmpty \quad \cHalf \quad \cEmpty\\
\bottomrule
\end{tabular}
}
\vspace{0.5em}
\caption{Case dev\_Physics\_5\_img1}

\end{table}

%%%%%%%%%%%%%%%%%%dev_Accounting_4_img1%%%%%%%%%%%%%%%%%%%
\begin{table}[htbp]
\centering
\small
\resizebox{\textwidth}{!}{%
\begin{tabular}{C{3.2cm} C{3.2cm} C{3.2cm} C{3.2cm} C{3.2cm}}
\toprule

% --- line1 ---
% \textbf{ Original Image } & 
\multicolumn{3}{c}{\textbf{Question \& Answer}} & 
\multicolumn{2}{c}{\textbf{Key Points}} \\ 
\cmidrule(r){1-3} 
%\cmidrule(lr){2-3} 
\cmidrule(l){4-5}

\multicolumn{3}{c}{
    \begin{tabular}{p{10.0cm}} 
        \textbf{Q:}  Paper Submarine Manufacturing is investigating a lockbox system to reduce its collection time. It has determined the following: The total collection time will be reduced by three days if the lockbox system is adopted. What is the net cash flow per check from adopting? (A) \$.02 (B) \$7.79 (C) \$8.65  \\ 
        \textbf{A:} (A)
    \end{tabular}  
} &
\multicolumn{2}{c}{
    \begin{tabular}{l}
    1. Text is clear without overlapping. \\
    2. Content is completely identical. \\
    3. Proportional layout is reasonable.
    \end{tabular} 
}\\

\midrule

% --- line2 ---
\textbf{Original Image} & 
\textbf{Gemini-2.5-pro (ours)} & 
\textbf{GPT-5 (ours)} & 
\textbf{Gemini-2.5-pro (IR)} & 
\textbf{GPT-5 (IR)} \\ 
% \cmidrule(r){1-1} 
% \cmidrule(lr){2-2} 
% \cmidrule(lr){3-3} 
% \cmidrule(lr){4-4} 
% \cmidrule(l){5-5}
%\addlinespace[3pt]

% --- line3 ---
\includegraphics[width=3.2cm]{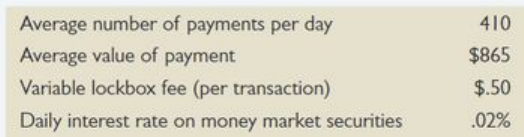} & 
\includegraphics[width=3.2cm]{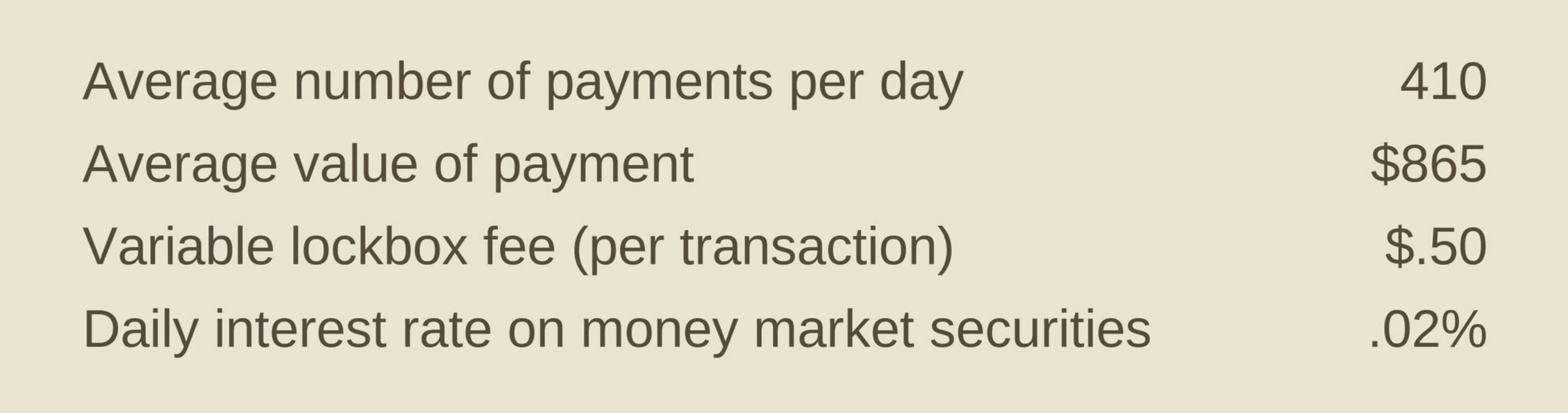} & 
\includegraphics[width=3.2cm]{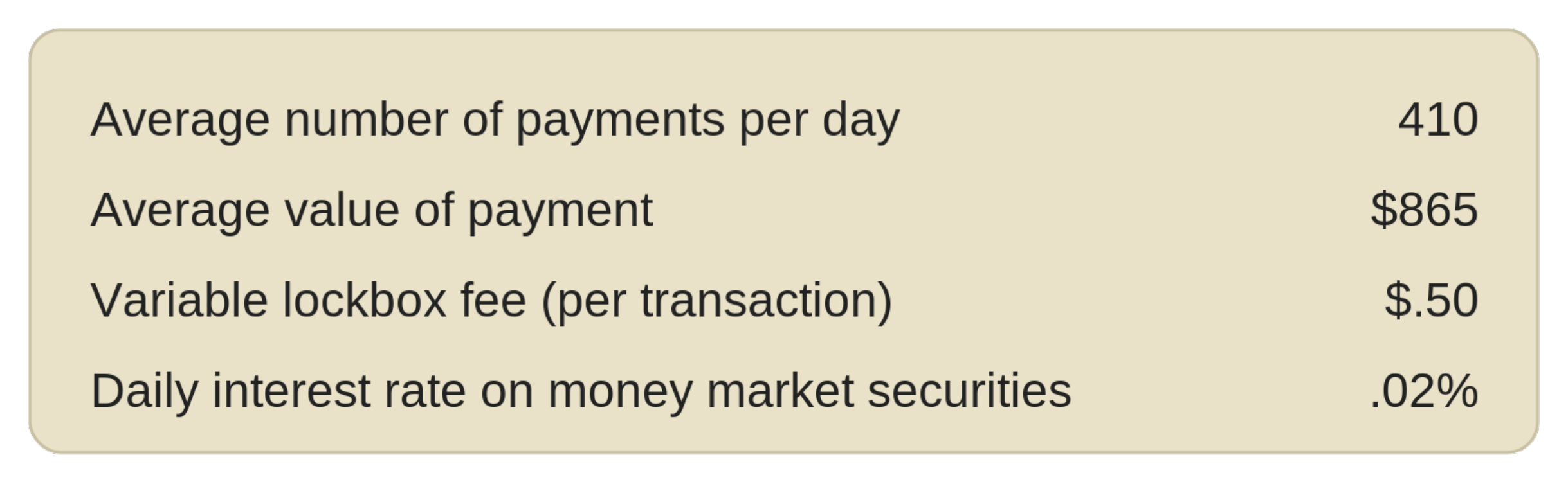} & 
\includegraphics[width=3.2cm]{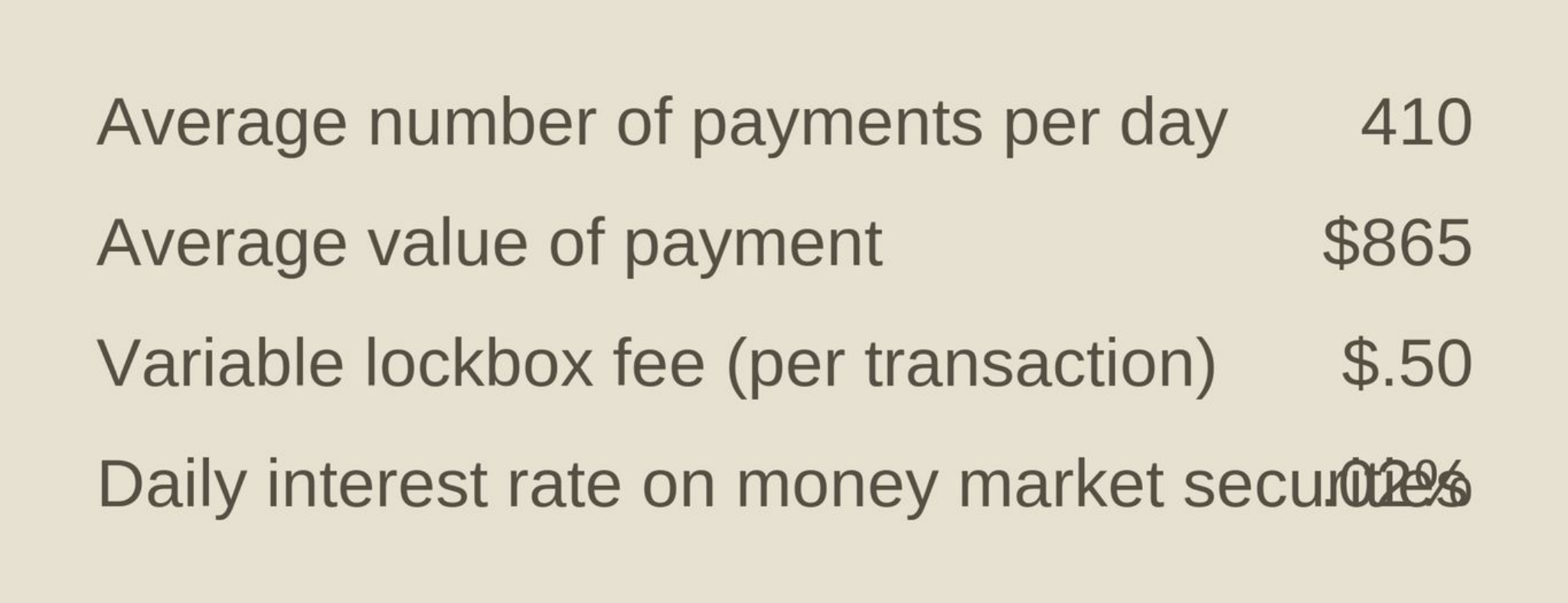} & 
\includegraphics[width=3.2cm]{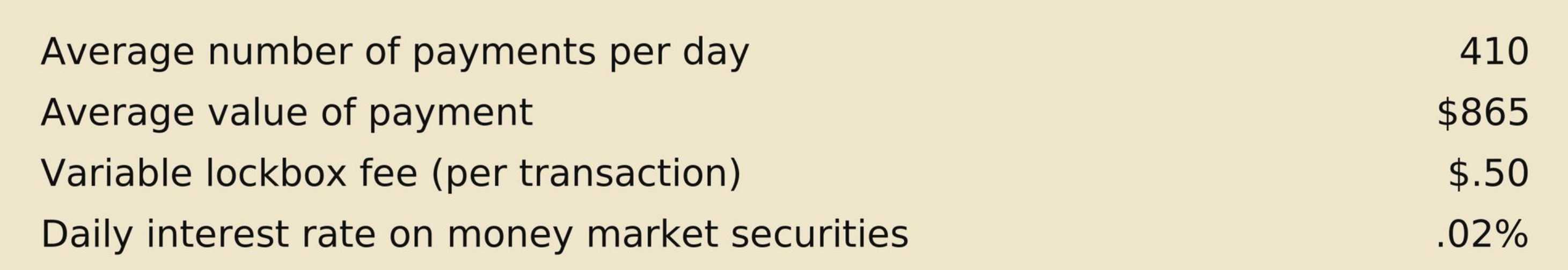} \\ 

% --- line4 ---
\cFull \quad \cFull \quad \cFull & 
\cFull \quad \cFull \quad \cFull & 
\cFull \quad \cFull \quad \cFull & 
\cEmpty \quad \cHalf \quad \cHalf & 
\cFull \quad \cFull \quad \cHalf\\
\bottomrule
\end{tabular}
}
\vspace{0.5em}
\caption{Case dev\_Physics\_5\_img1}

\end{table}

% \paragraph{CV-Bench}

%%%%%%%%%%%%%%%%%%coco_63%%%%%%%%%%%%%%%%%%%
\begin{table}[!htbp]
\centering
\small
\resizebox{\textwidth}{!}{%
\begin{tabular}{C{3.2cm} C{3.2cm} C{3.2cm} C{3.2cm} C{3.2cm}}
\toprule

% --- line1 ---
% \textbf{ Original Image } & 
\multicolumn{3}{c}{\textbf{Question \& Answer}} & 
\multicolumn{2}{c}{\textbf{Key Points}} \\ 
\cmidrule(r){1-3} 
%\cmidrule(lr){2-3} 
\cmidrule(l){4-5}

\multicolumn{3}{c}{
    \begin{tabular}{p{10.0cm}} 
        \textbf{Q:} How many clocks are in the image? Select from the following choices. (A) 2 (B) 1 (C) 0 (D) 3 Answer with the option's letter from the given choices directly. \\ 
        \textbf{A:} B  
    \end{tabular}  
} &
\multicolumn{2}{c}{
    \begin{tabular}{l}
    1. There is a clock on the clock tower.\\
    2. Two guards and two tourists in the square.\\
    3. Right perspective relationship of the building.\\
    4. Windows on the right wall.
    \end{tabular} 
}\\

\midrule

% --- line2 ---
\textbf{Original Image} & 
\textbf{Gemini-2.5-pro (ours)} & 
\textbf{GPT-5 (ours)} & 
\textbf{Gemini-2.5-pro (IR)} & 
\textbf{GPT-5 (IR)} \\ 
% \cmidrule(r){1-1} 
% \cmidrule(lr){2-2} 
% \cmidrule(lr){3-3} 
% \cmidrule(lr){4-4} 
% \cmidrule(l){5-5}
%\addlinespace[3pt]

% --- line3 ---
\includegraphics[width=3.2cm]{imags/pdf/appendix/coco_63a.pdf} & 
\includegraphics[width=3.2cm]{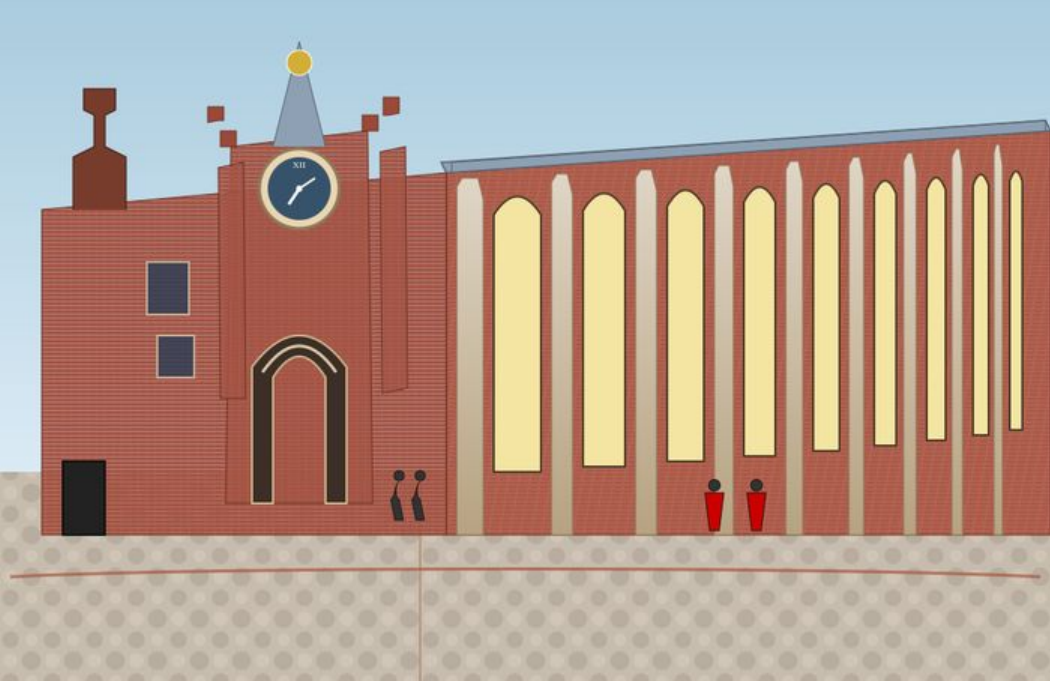} & 
\includegraphics[width=3.2cm]{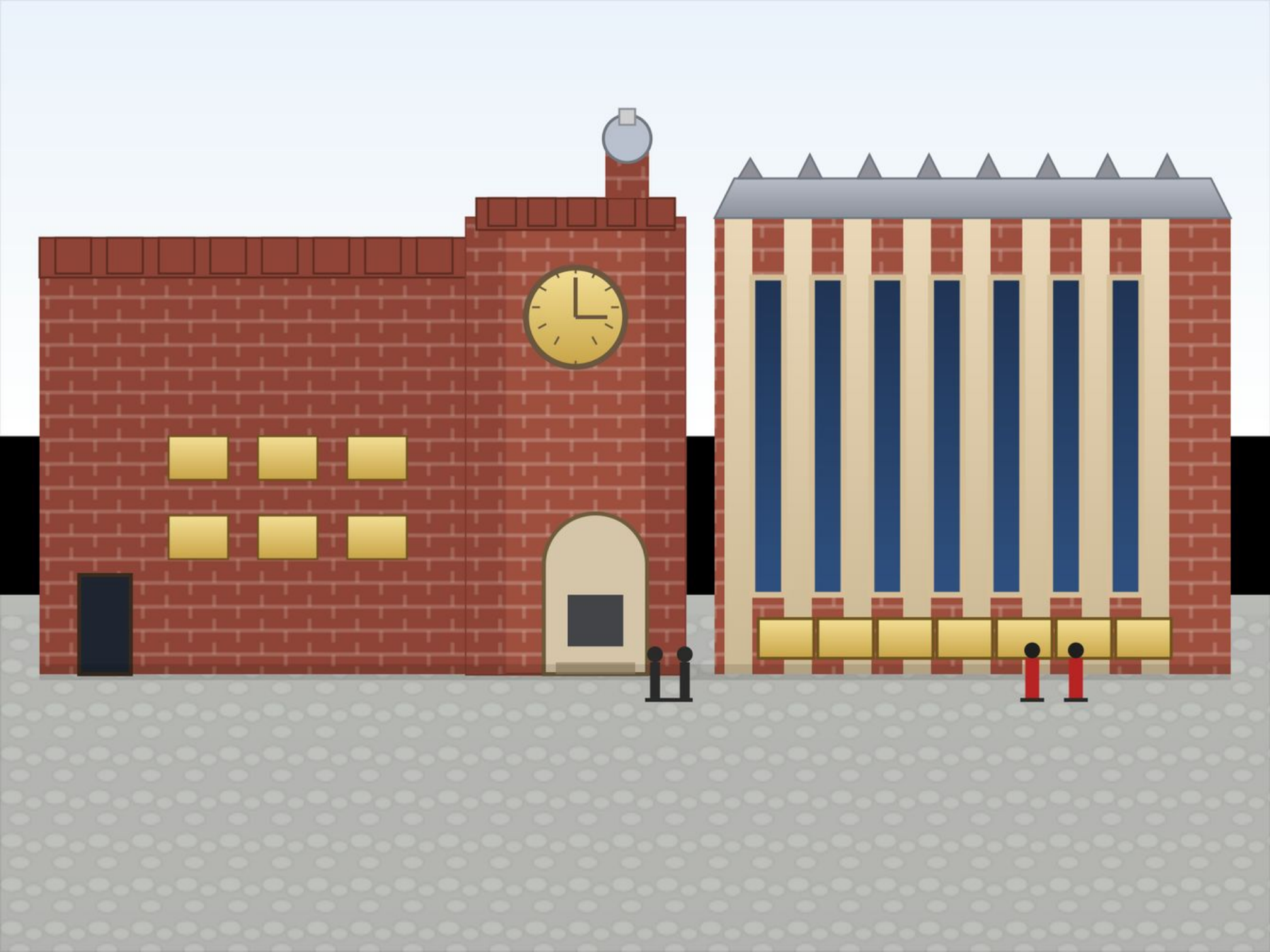} & 
\includegraphics[width=3.2cm]{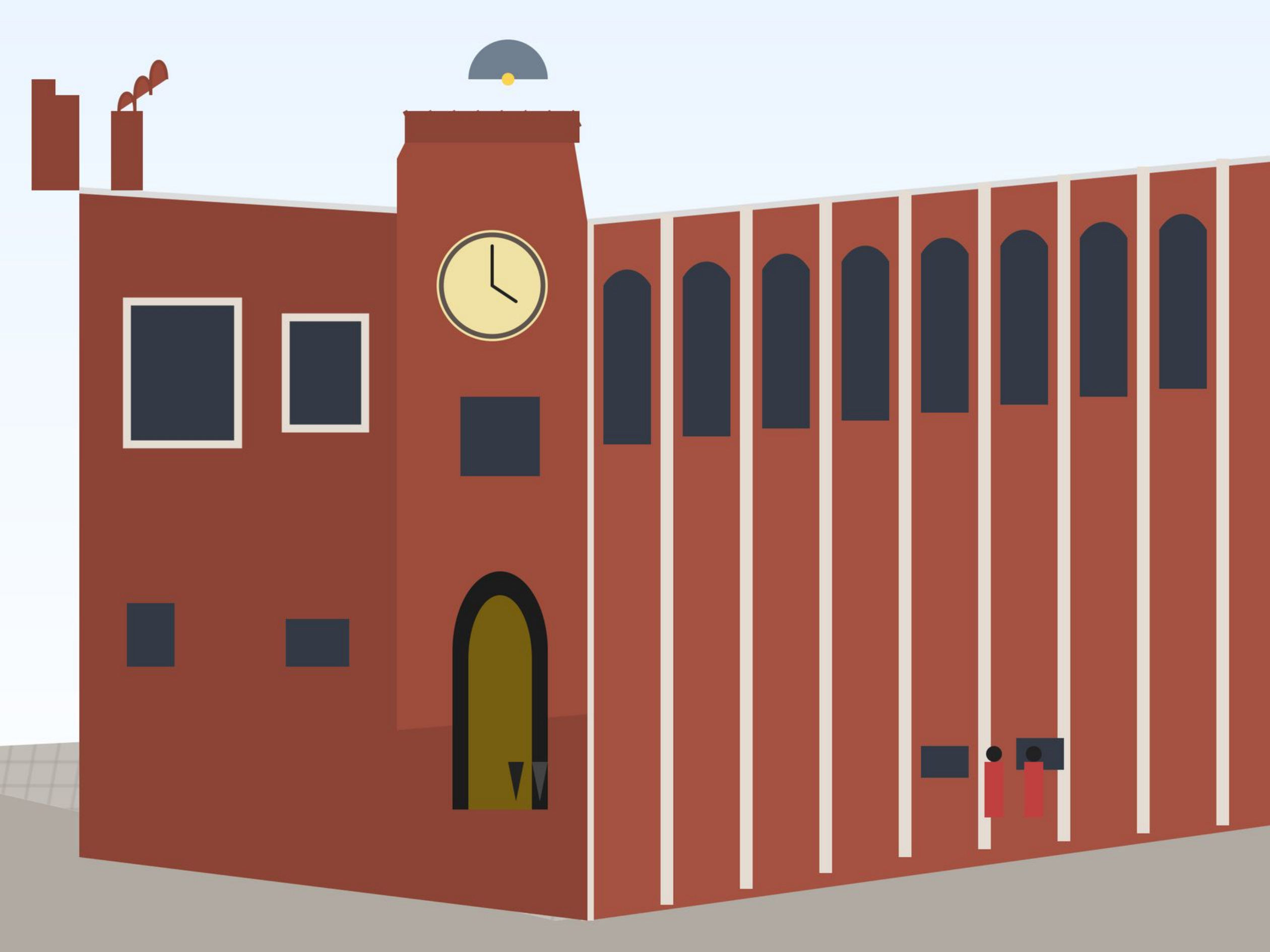} & 
\includegraphics[width=3.2cm]{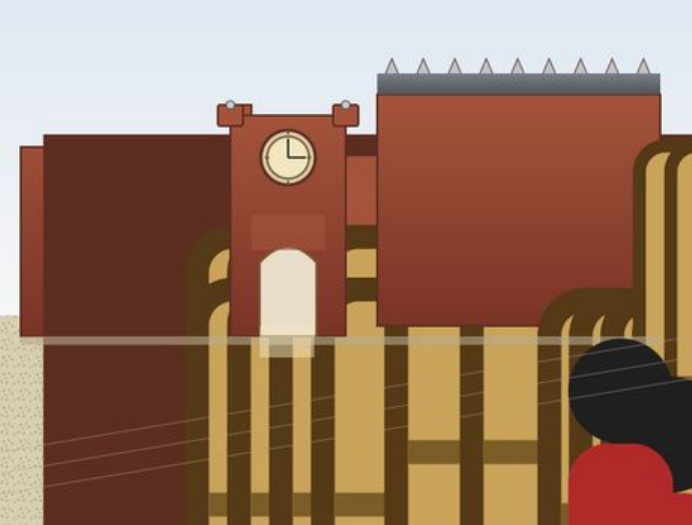} \\ 

% --- line4 ---
\cFull \quad \cFull \quad \cFull \quad \cFull & 
\cFull \quad \cFull \quad \cHalf \quad \cFull& 
\cFull \quad \cFull \quad \cHalf \quad \cFull& 
\cFull \quad \cFull \quad \cHalf \quad \cHalf& 
\cFull \quad \cEmpty \quad \cEmpty \quad \cHalf\\
\bottomrule
\end{tabular}
}
\vspace{0.5em}
\caption{Case coco\_63}

\end{table}

%%%%%%%%%%%%%%%%%%omni3d_hypersim_140%%%%%%%%%%%%%%%%%%%
\begin{table}[!htbp]
\centering
\small
\resizebox{\textwidth}{!}{%
\begin{tabular}{C{3.2cm} C{3.2cm} C{3.2cm} C{3.2cm} C{3.2cm}}
\toprule

% --- line1 ---
% \textbf{ Original Image } & 
\multicolumn{3}{c}{\textbf{Question \& Answer}} & 
\multicolumn{2}{c}{\textbf{Key Points}} \\ 
\cmidrule(r){1-3} 
%\cmidrule(lr){2-3} 
\cmidrule(l){4-5}

\multicolumn{3}{c}{
    \begin{tabular}{p{10.0cm}} 
        \textbf{Q:} Estimate the real-world distances between objects in this image. Which object is closer to the desk (highlighted by a red box), the books (highlighted by a blue box) or the shelves (highlighted by a green box)? (A) books (B) shelves. Answer with the option's letter from the given choices directly. \\ 
        \textbf{A:} B  
    \end{tabular}  
} &
\multicolumn{2}{c}{
    \begin{tabular}{p{6.0cm}}
    1. Three bounding boxes are accurately calibrated.\\
    2. The objects represented by the bounding box annotations are accurate.\\
    3. The spatial orientation of the stairs is correct.\\
    4. Environmental detail representation.
    \end{tabular} 
}\\

\midrule

% --- line2 ---
\textbf{Original Image} & 
\textbf{Gemini-2.5-pro (ours)} & 
\textbf{GPT-5 (ours)} & 
\textbf{Gemini-2.5-pro (IR)} & 
\textbf{GPT-5 (IR)} \\ 
% \cmidrule(r){1-1} 
% \cmidrule(lr){2-2} 
% \cmidrule(lr){3-3} 
% \cmidrule(lr){4-4} 
% \cmidrule(l){5-5}
%\addlinespace[3pt]

% --- line3 ---
\includegraphics[width=3.2cm]{imags/pdf/appendix/omni3d_hypersim_140a.pdf} & 
\includegraphics[width=3.2cm]{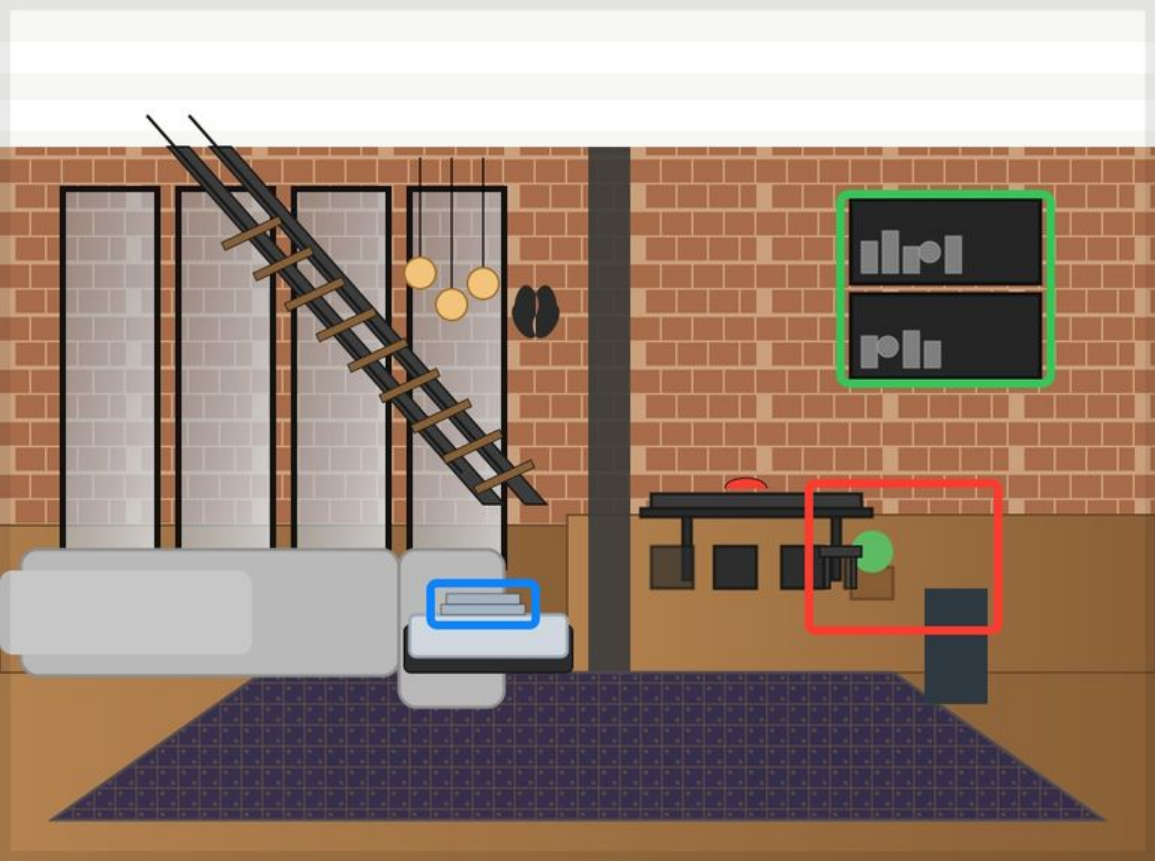} & 
\includegraphics[width=3.2cm]{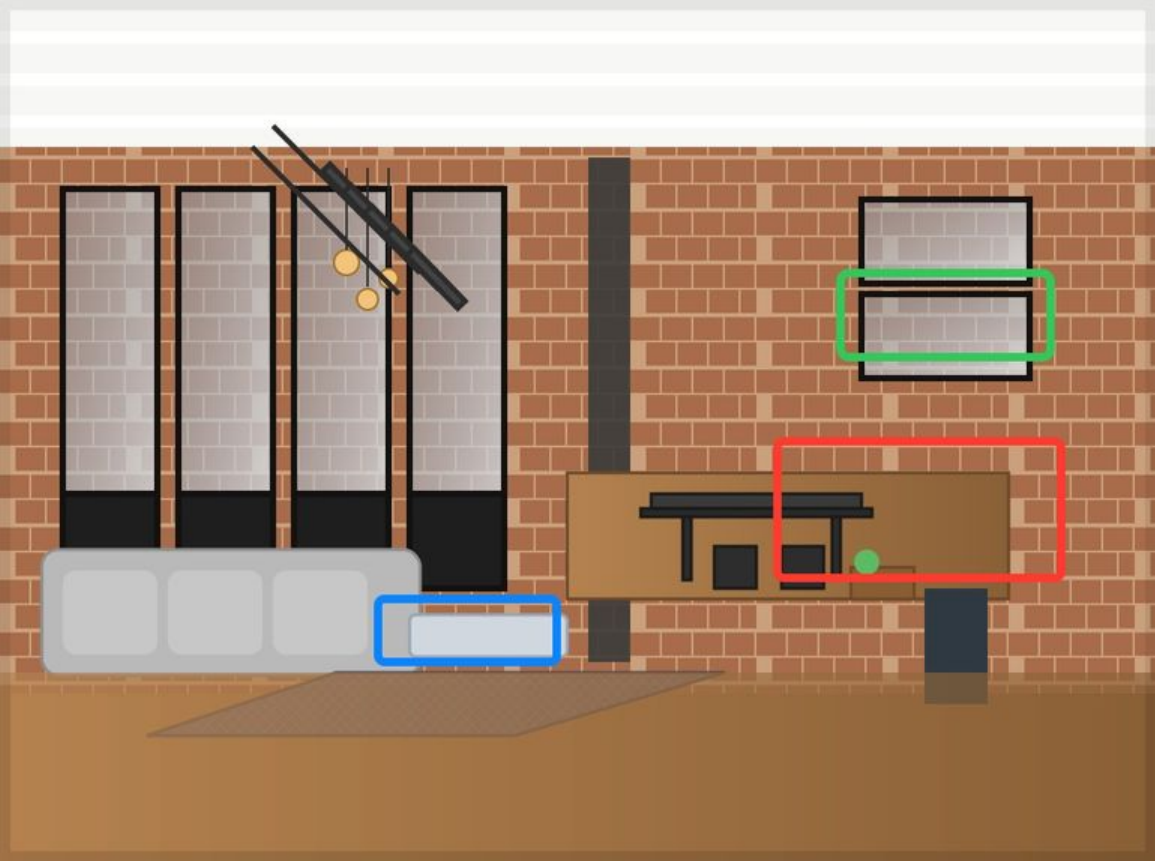} & 
\includegraphics[width=3.2cm]{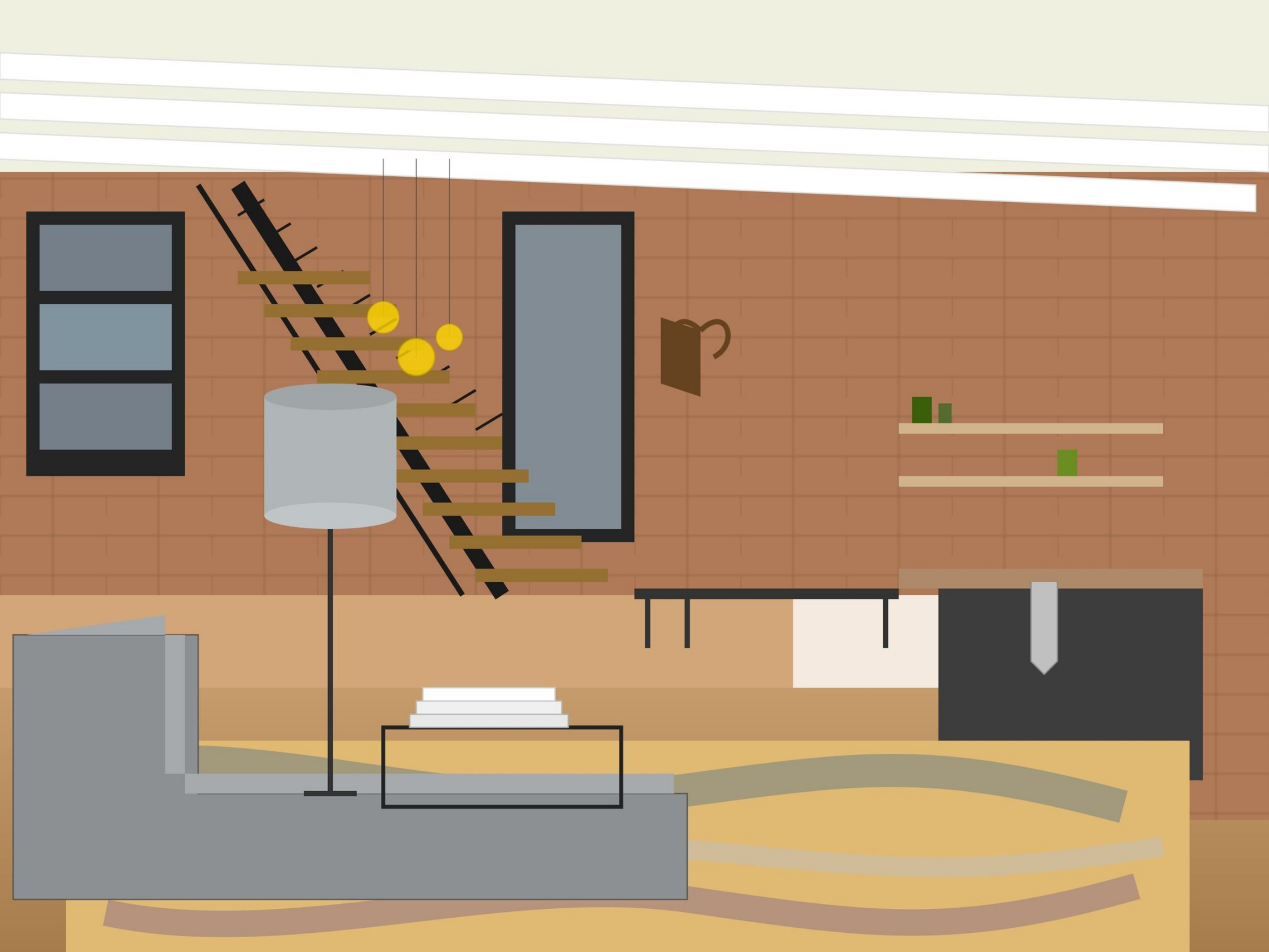} & 
\includegraphics[width=3.2cm]{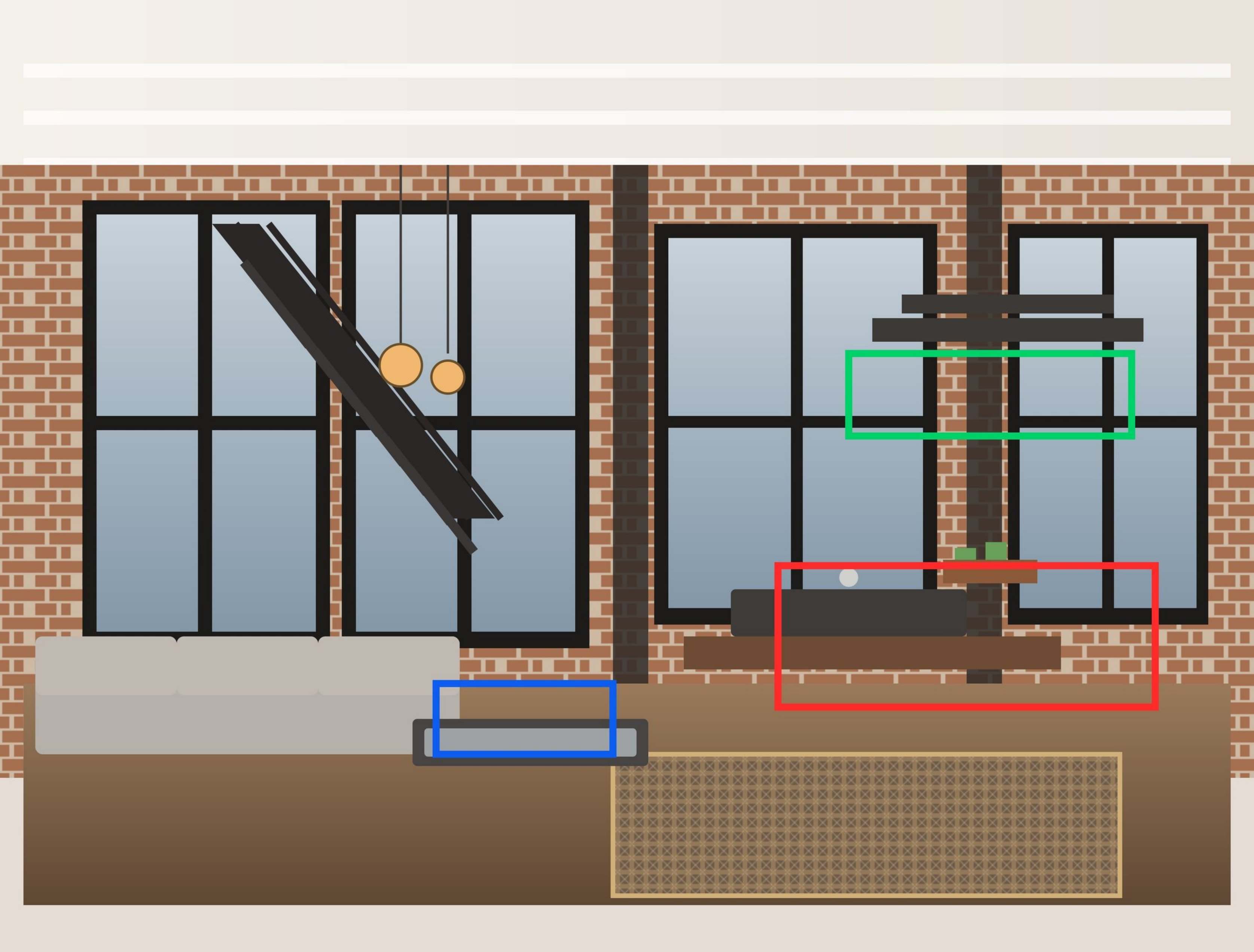} \\ 

% --- line4 ---
\cFull \quad \cFull \quad \cFull \quad \cFull & 
\cFull \quad \cFull \quad \cHalf \quad \cFull& 
\cFull \quad \cHalf  \quad \cHalf \quad \cFull& 
\cEmpty \quad \cHalf \quad \cHalf \quad \cHalf& 
\cFull \quad \cFull \quad \cHalf \quad \cHalf\\
\bottomrule
\end{tabular}
}
\vspace{0.5em}
\caption{Case omni3d\_hypersim\_140}

\end{table}

%%%%%%%%%%%%%%%%%%ade20k_668%%%%%%%%%%%%%%%%%%%
\begin{table}[!htbp]
\centering
\small
\resizebox{\textwidth}{!}{%
\begin{tabular}{C{3.2cm} C{3.2cm} C{3.2cm} C{3.2cm} C{3.2cm}}
\toprule

% --- line1 ---
% \textbf{ Original Image } & 
\multicolumn{3}{c}{\textbf{Question \& Answer}} & 
\multicolumn{2}{c}{\textbf{Key Points}} \\ 
\cmidrule(r){1-3} 
%\cmidrule(lr){2-3} 
\cmidrule(l){4-5}

\multicolumn{3}{c}{
    \begin{tabular}{p{10.0cm}} 
        \textbf{Q:} Considering the relative positions of the wall (annotated by the red box) and the door in the image provided, where is the wall (annotated by the red box) located with respect to the door? Select from the following choices. (A) left (B) right Answer with the option's letter from the given choices directly. \\ 
        \textbf{A:} B  
    \end{tabular}  
} &
\multicolumn{2}{c}{
    \begin{tabular}{l}
    1. A bounding boxe is accurately calibrated.\\
    2. Spatial relationship of the spiral staircase.\\
    3. Scene restoration fidelity.\\
    \end{tabular} 
}\\

\midrule

% --- line2 ---
\textbf{Original Image} & 
\textbf{Gemini-2.5-pro (ours)} & 
\textbf{GPT-5 (ours)} & 
\textbf{Gemini-2.5-pro (IR)} & 
\textbf{GPT-5 (IR)} \\ 
% \cmidrule(r){1-1} 
% \cmidrule(lr){2-2} 
% \cmidrule(lr){3-3} 
% \cmidrule(lr){4-4} 
% \cmidrule(l){5-5}
%\addlinespace[3pt]

% --- line3 ---
\includegraphics[width=3.2cm]{imags/pdf/appendix/ade20k_668a.pdf} & 
\includegraphics[width=3.2cm]{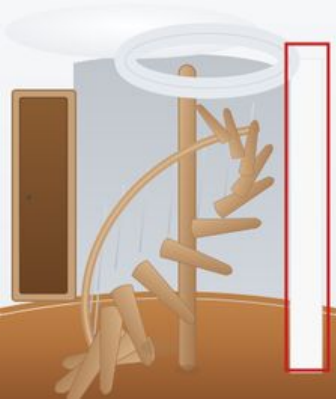} & 
\includegraphics[width=3.2cm]{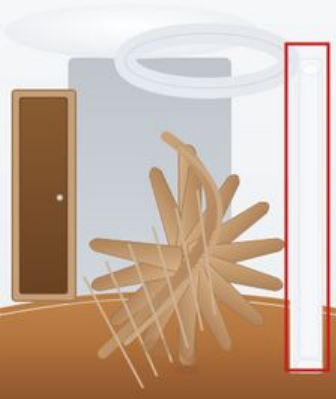} & 
\includegraphics[width=3.2cm]{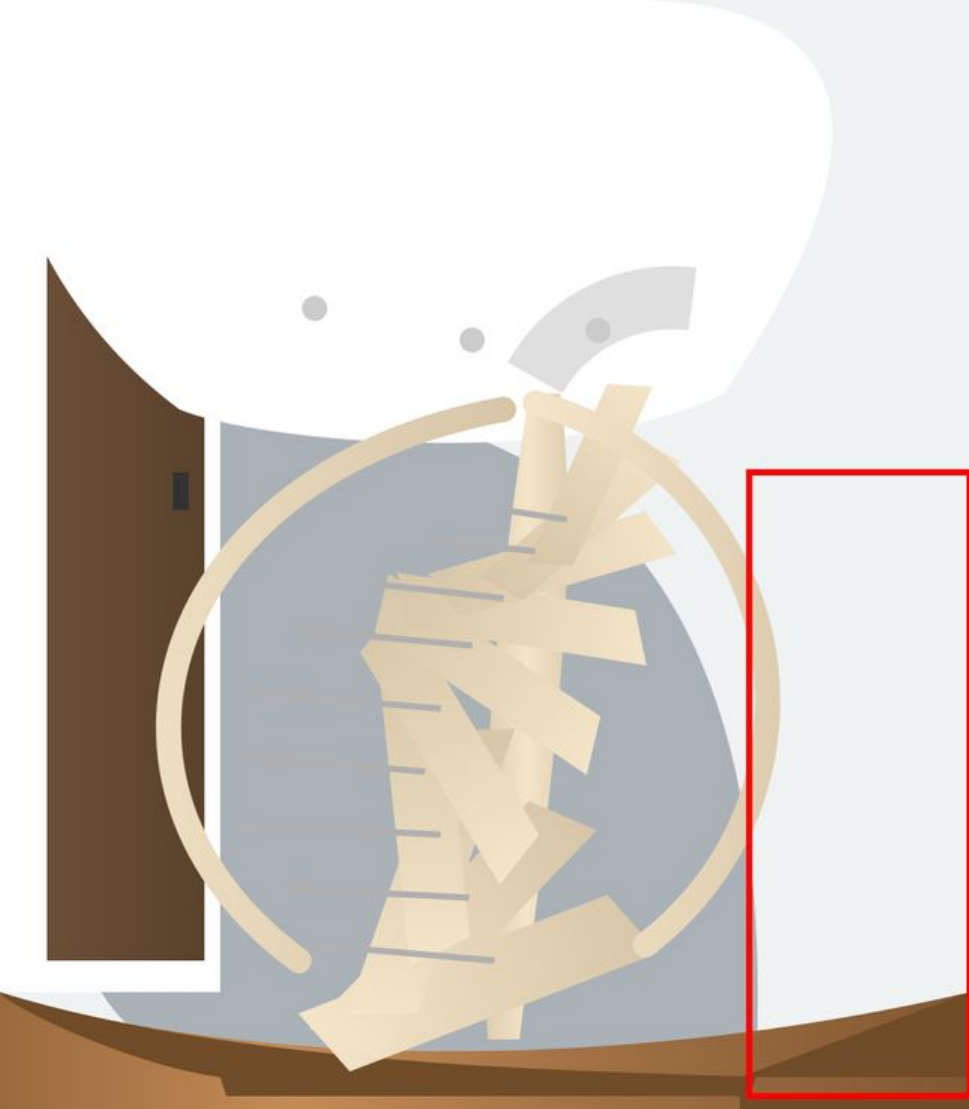} & 
\includegraphics[width=3.2cm]{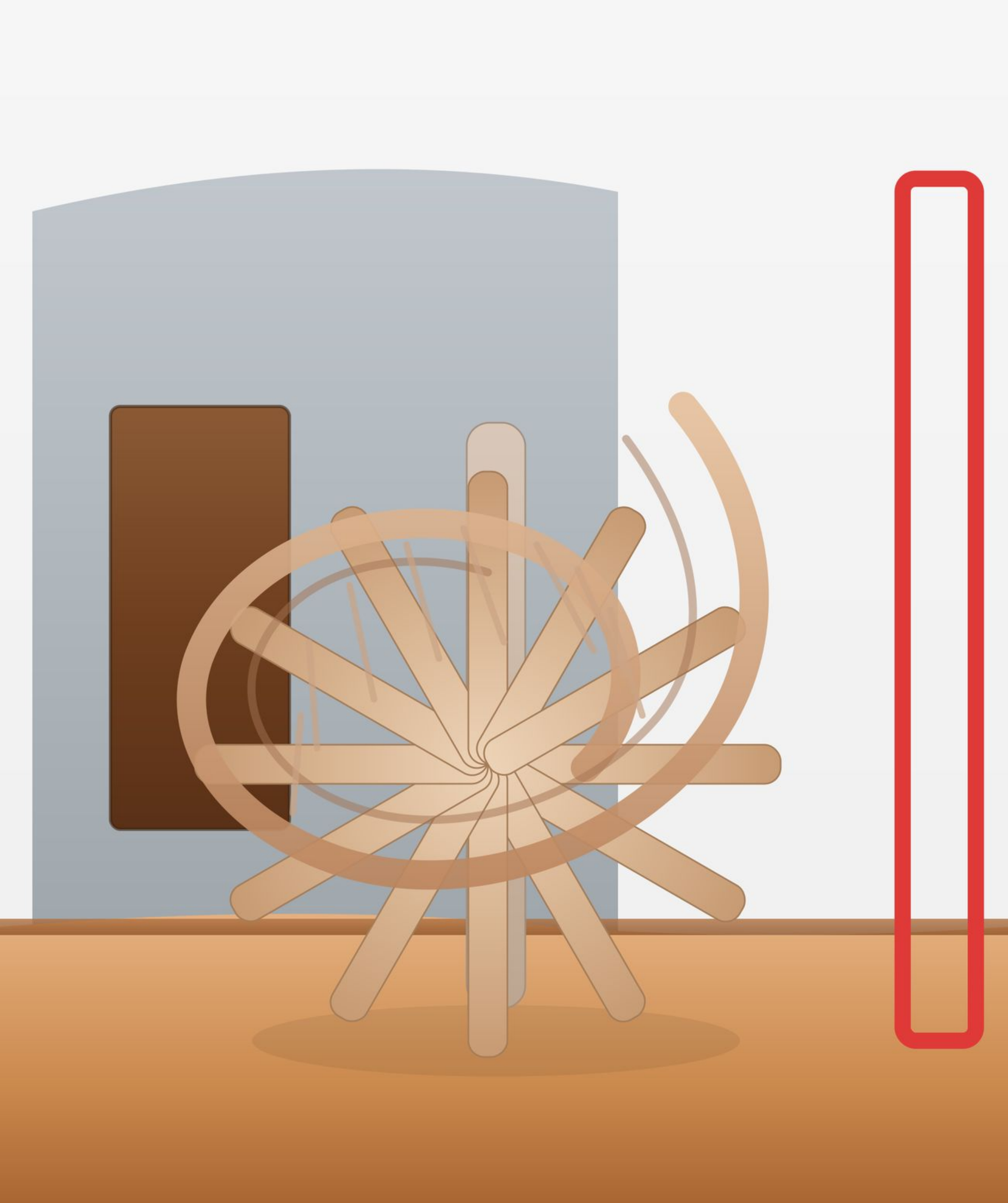} \\ 

% --- line4 ---
\cFull \quad \cFull \quad \cFull & 
\cFull \quad \cFull \quad \cFull & 
\cFull \quad \cHalf \quad \cFull & 
\cFull \quad \cHalf \quad \cHalf & 
\cFull \quad \cHalf \quad \cHalf \\
\bottomrule
\end{tabular}
}
\vspace{0.5em}
\caption{Case ade20k\_668}

\end{table}

%%%%%%%%%%%%%%%%%%omni3d_sunrgbd_109%%%%%%%%%%%%%%%%%%%
\begin{table}[!htbp]
\centering
\small
\resizebox{\textwidth}{!}{%
\begin{tabular}{C{3.2cm} C{3.2cm} C{3.2cm} C{3.2cm} C{3.2cm}}
\toprule

% --- line1 ---
% \textbf{ Original Image } & 
\multicolumn{3}{c}{\textbf{Question \& Answer}} & 
\multicolumn{2}{c}{\textbf{Key Points}} \\ 
\cmidrule(r){1-3} 
%\cmidrule(lr){2-3} 
\cmidrule(l){4-5}

\multicolumn{3}{c}{
    \begin{tabular}{p{10.0cm}} 
        \textbf{Q:} Which object is closer to the camera taking this photo, the night stand (highlighted by a red box) or the chair (highlighted by a blue box)? (A) night stand (B) chair Answer with the option's letter from the given choices directly. \\ 
        \textbf{A:} A  
    \end{tabular}  
} &
\multicolumn{2}{c}{
    \begin{tabular}{p{6.0cm}}
    1.Two bounding boxes are accurately shown.\\
    2.The objects represented by the bounding box annotations are accurate.\\
    3.Right perspective relationship of the bed frame.\\
    4.Environmental detail representation.
    \end{tabular} 
}\\

\midrule

% --- line2 ---
\textbf{Original Image} & 
\textbf{Gemini-2.5-pro (ours)} & 
\textbf{GPT-5 (ours)} & 
\textbf{Gemini-2.5-pro (IR)} & 
\textbf{GPT-5 (IR)} \\ 
% \cmidrule(r){1-1} 
% \cmidrule(lr){2-2} 
% \cmidrule(lr){3-3} 
% \cmidrule(lr){4-4} 
% \cmidrule(l){5-5}
%\addlinespace[3pt]

% --- line3 ---
\includegraphics[width=3.2cm]{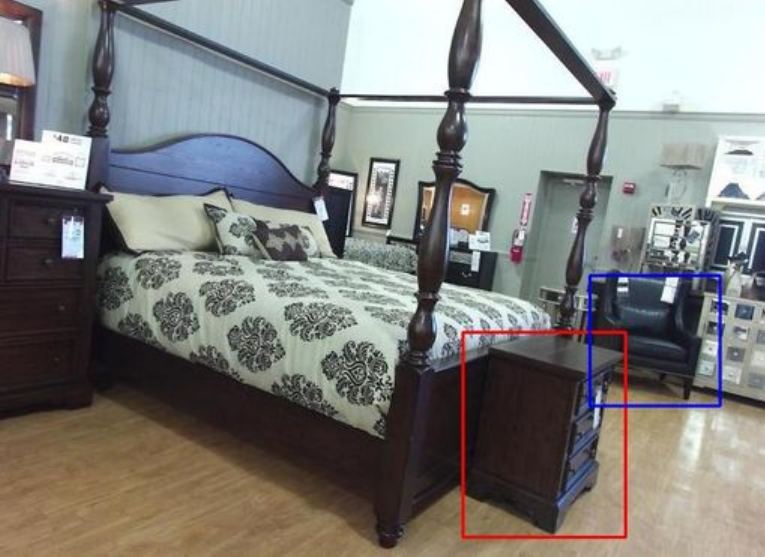} & 
\includegraphics[width=3.2cm]{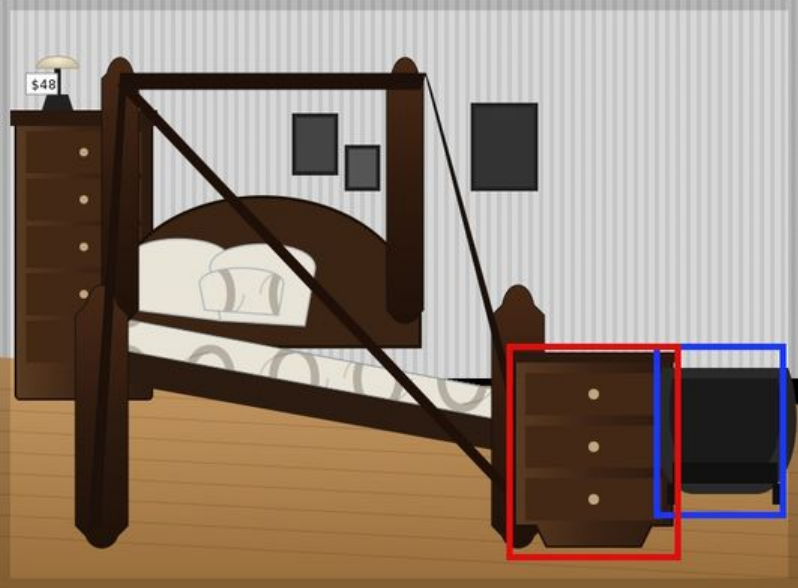} & 
\includegraphics[width=3.2cm]{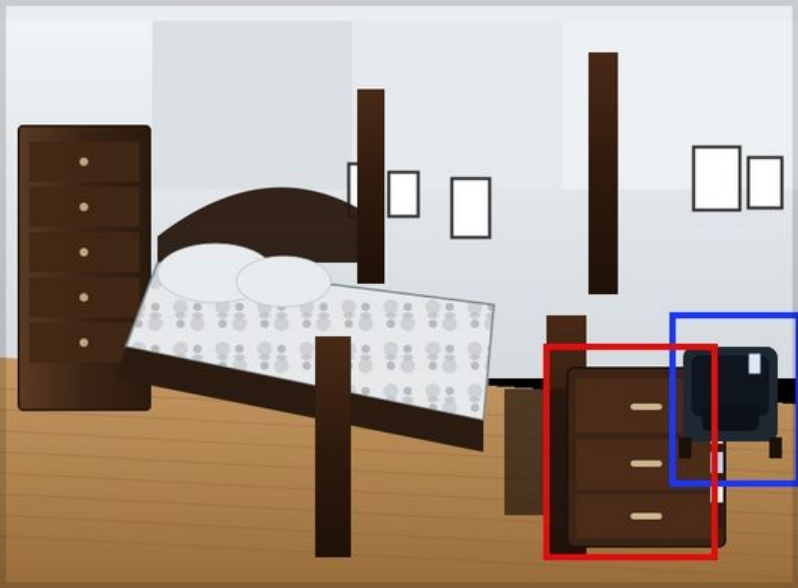} & 
\includegraphics[width=3.2cm]{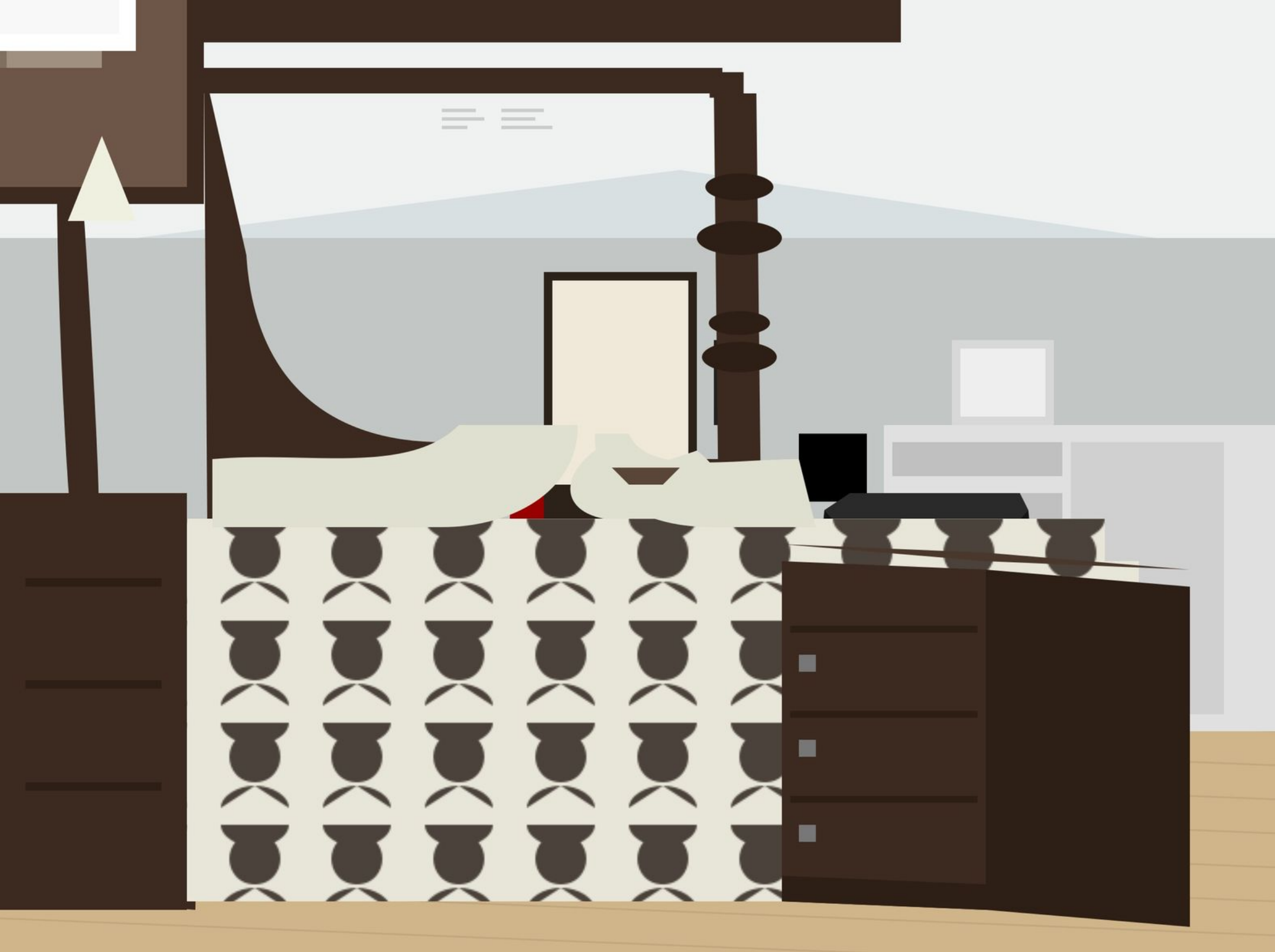} & 
\includegraphics[width=3.2cm]{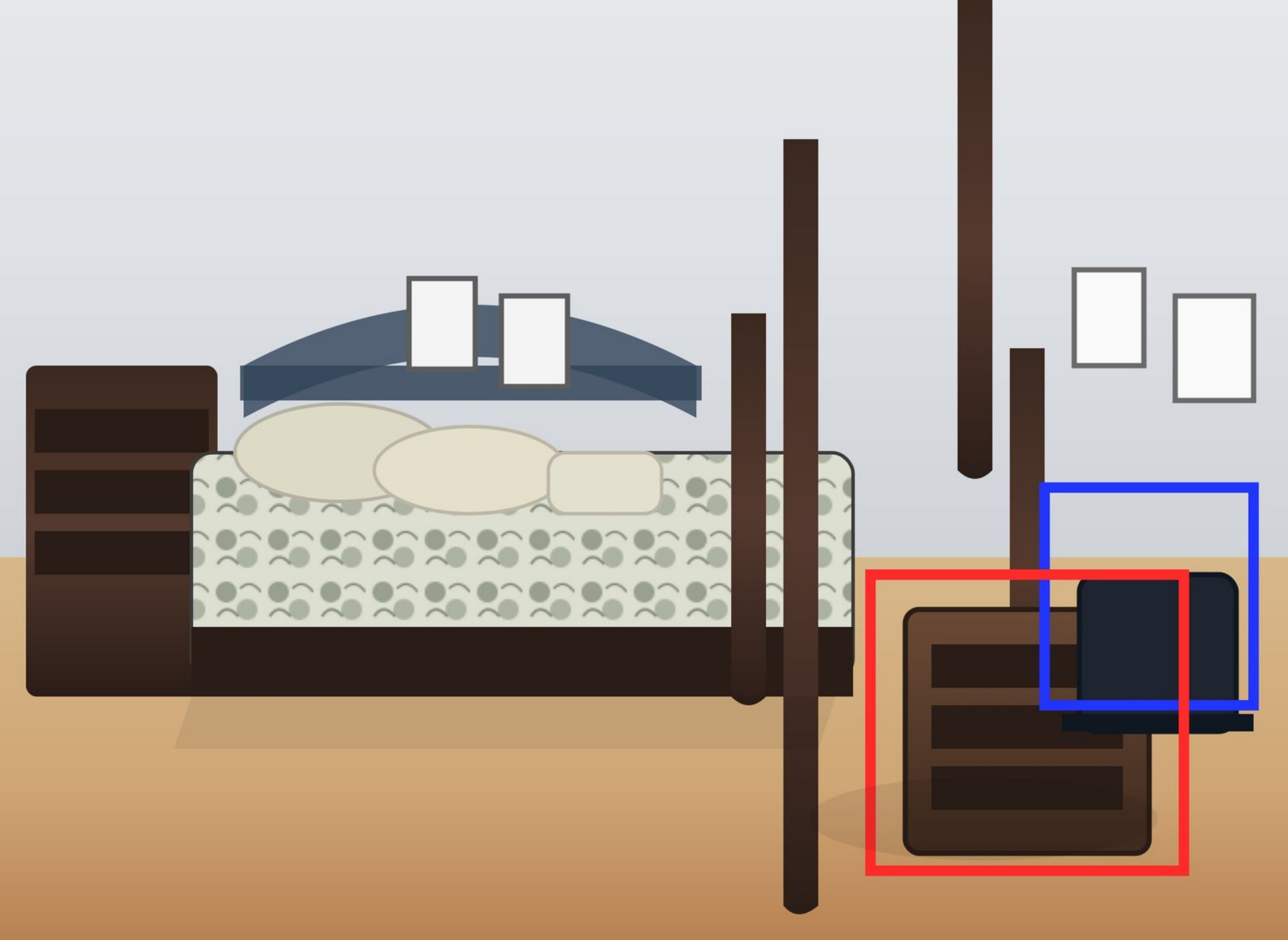} \\ 

% --- line4 ---
\cFull \quad \cFull \quad \cFull \quad \cFull & 
\cFull \quad \cFull \quad \cFull \quad \cFull & 
\cFull \quad \cFull \quad \cHalf \quad \cHalf & 
\cEmpty \quad \cHalf \quad \cEmpty \quad \cHalf & 
\cFull \quad \cFull \quad \cHalf \quad \cHalf\\
\bottomrule
\end{tabular}
}
\vspace{0.5em}
\caption{Case omni3d\_sunrgbd\_109}

\end{table}

\clearpage
\subsubsection{CRUD-Action Trajectory of Canvas-CoT}
\label{sec:traj}

\begin{figure}[h]
    \centering
    
    \begin{subfigure}{0.16\linewidth}
        \centering
        \includegraphics[width=\linewidth]{imags/pdf/appendix/logcase_a.pdf}
        \caption{Original Image}
        \label{fig:logcase_a}
    \end{subfigure}
    \hfill
    \begin{subfigure}{0.16\linewidth}
        \centering
        \includegraphics[width=\linewidth]{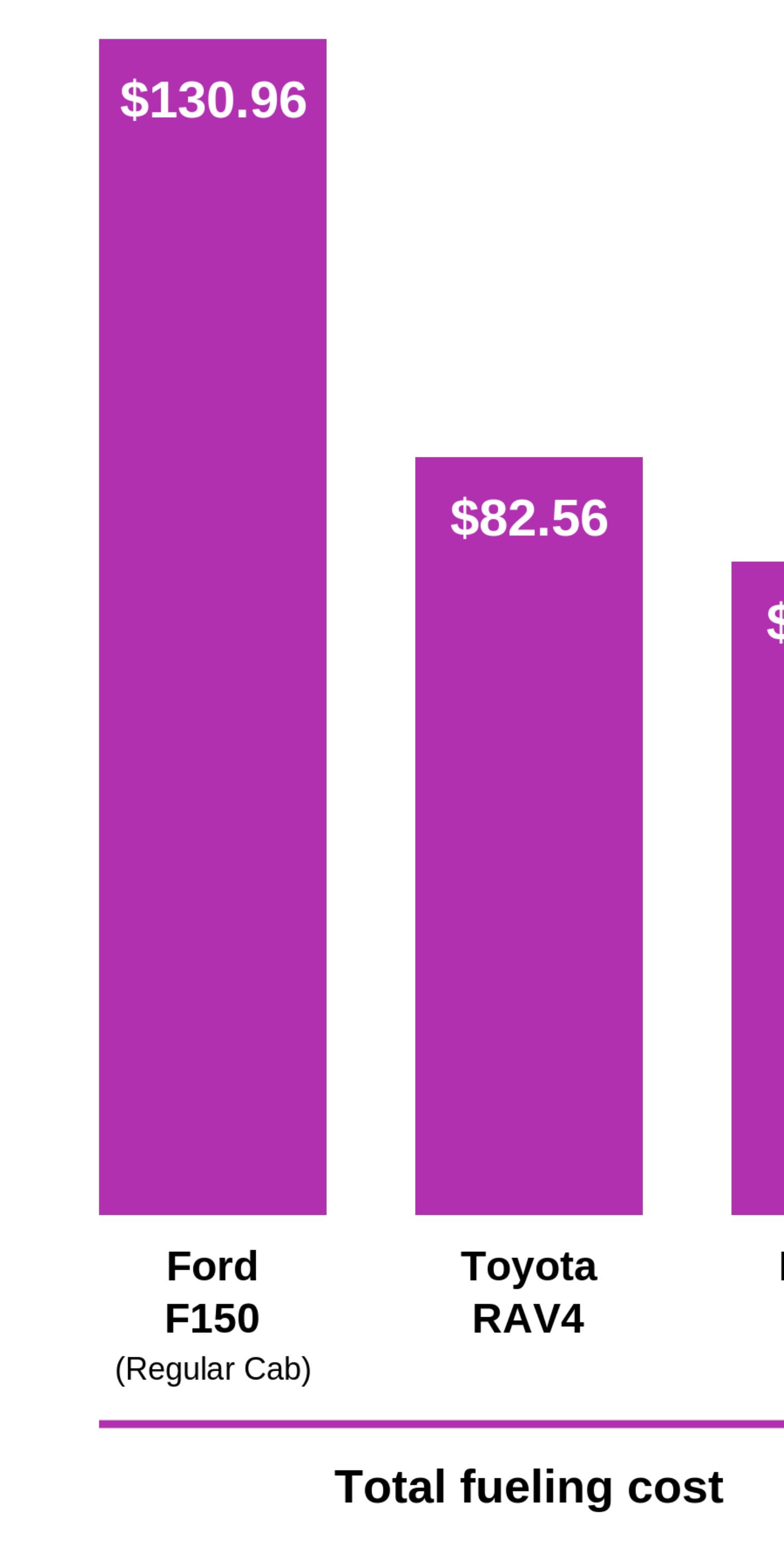}
        \caption{Image Rendered by initial SVG}
        \label{fig:logcase_b}
    \end{subfigure}
    \hfill
    \begin{subfigure}{0.16\linewidth}
        \centering
        \includegraphics[width=\linewidth]{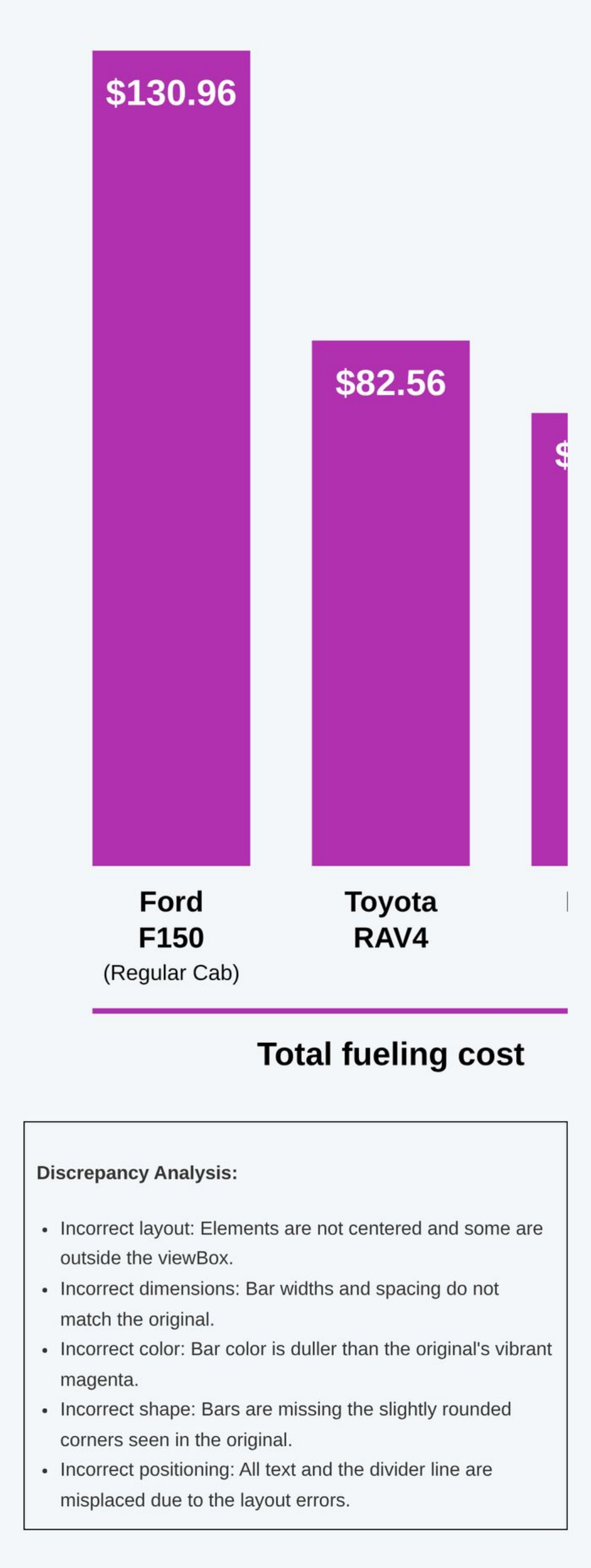}
        \caption{Image Rendered by SVG after Round-1}
        \label{fig:logcase_c}
    \end{subfigure}
    \hfill
    \begin{subfigure}{0.16\linewidth}
        \centering
        \includegraphics[width=\linewidth]{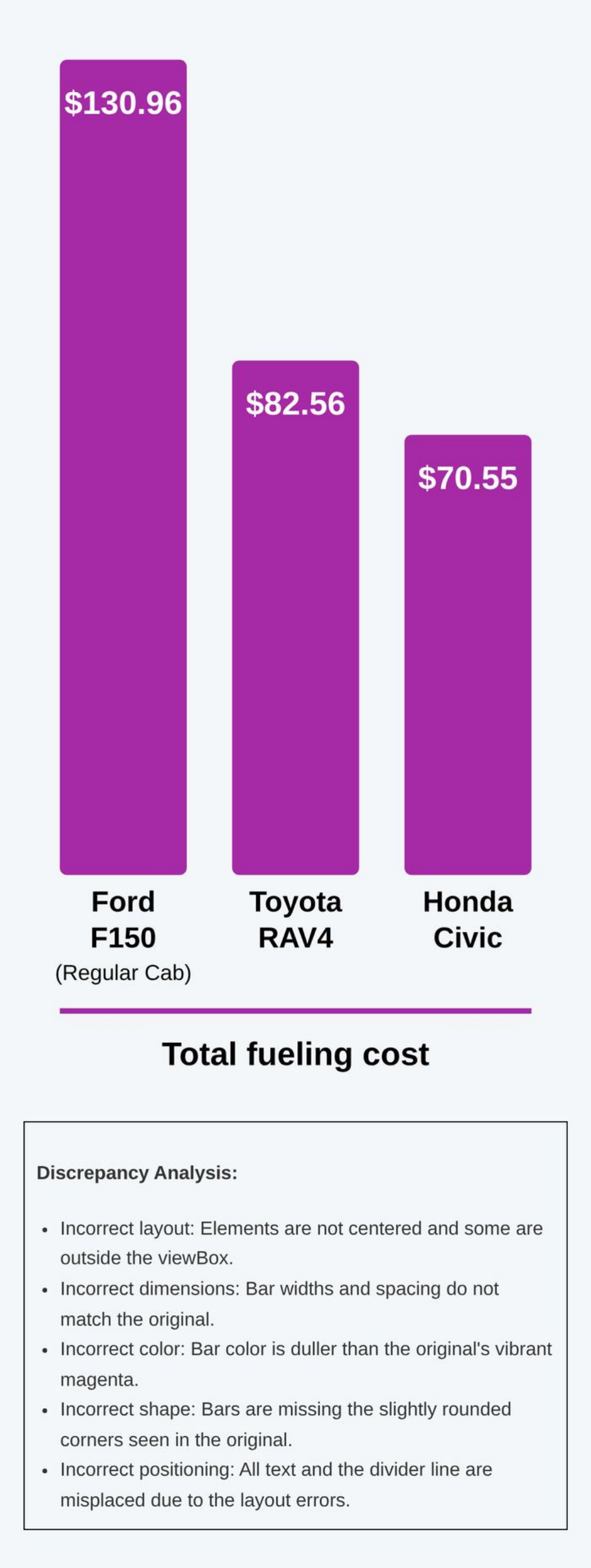}
        \caption{Image Rendered by SVG after Round-2}
        \label{fig:logcase_d}
    \end{subfigure}
    \hfill
    \begin{subfigure}{0.16\linewidth}
        \centering
        \includegraphics[width=\linewidth]{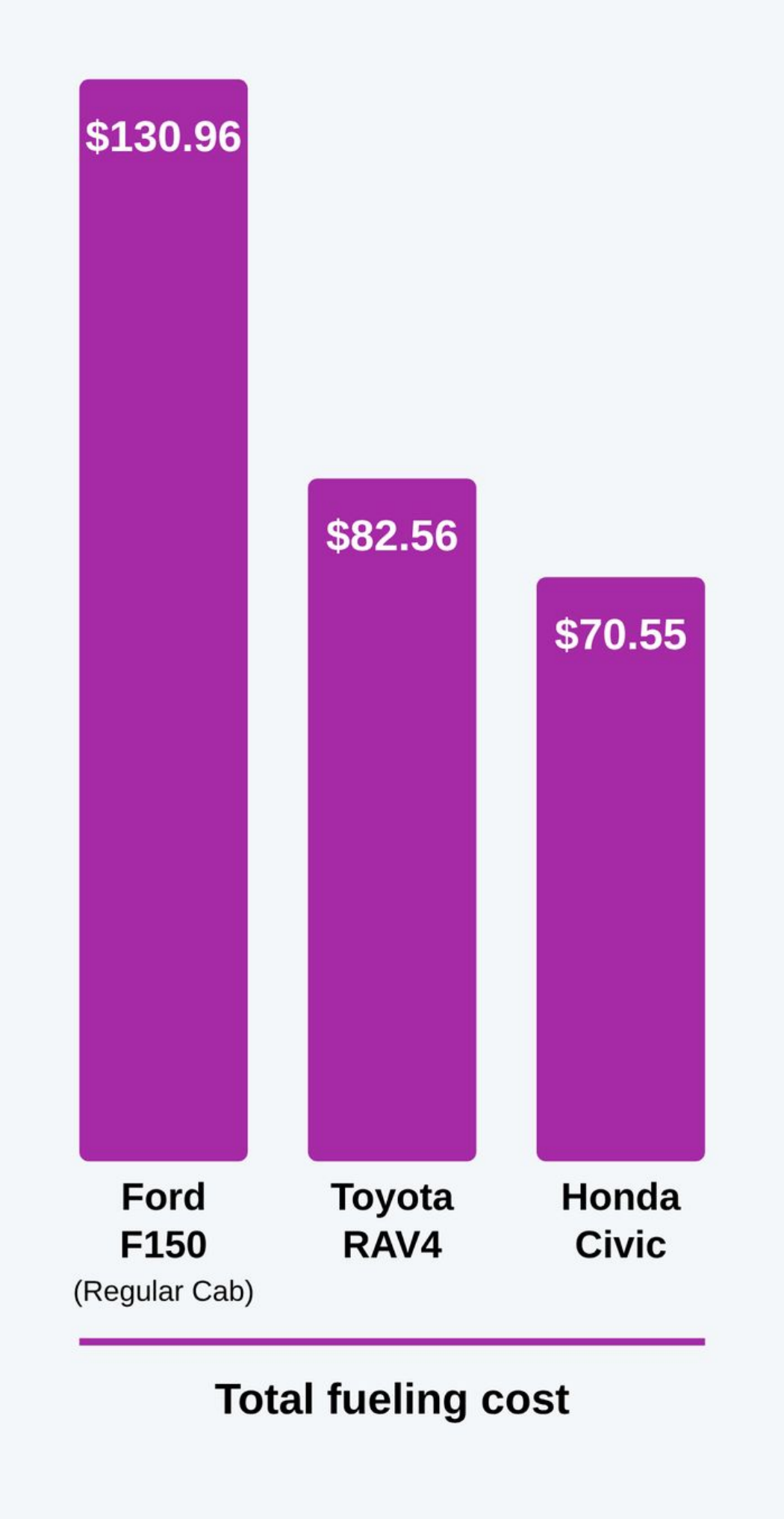}
        \caption{Image Rendered by final SVG}
        \label{fig:logcase_e}
    \end{subfigure}

    \caption{Image evolution with CRUD-Action Trajectory}
    \label{fig:logcase}
\end{figure}

Figure~\ref{fig:log} illustrates the execution flow of Canvas-CoT via a “Bar Chart Correction” task. The agent begins by identifying layout and styling discrepancies, utilizing \texttt{insert\_element} to externalize its reasoning onto the canvas. Guided by this analysis, the agent then performs a \texttt{replace\_element} operation to modify the target SVG node in-place, specifically recalculating bar coordinates and styling attributes. Finally, the agent employs \texttt{remove\_element} to delete the auxiliary reasoning nodes from the canvas. The corresponding visualization and the full trajectory are presented in Figure~\ref{fig:logcase} and Figure~\ref{fig:log}.

%%%%%%%%%%%%%%%%%%%%%log text%%%%%%%%%%%%%
% \begin{figure}[h] 
%     \centering
%         \includegraphics[width=1.0\linewidth]{imags/log-part1.png}
%         \caption{CRUD-Action Trajectory of Canvas-CoT}
%         \label{fig:log}
% \end{figure}

% \clearpage
% \begin{figure}[t]
%     \ContinuedFloat
%     \centering
%         \includegraphics[width=1.0\linewidth]{imags/log-part2.png}
%         \caption{CRUD-Action Trajectory of Canvas-CoT}
%         \label{fig:log}
% \end{figure}

% \clearpage
% \begin{figure}[t]
%     \ContinuedFloat
%     \centering
%         \includegraphics[width=1.0\linewidth]{imags/log-part3.png}
%         \caption{CRUD-Action Trajectory of Canvas-CoT}
%         \label{fig:log}
% \end{figure}

%%%%%%%%%%%%%%%%bar chart case%%%%%%%%%%%%%%%%%

%%%%%%%%%%%%%%%%%%%%%%%%bar chart anal%%%%%%
\begin{figure}[htbp]
    \centering
    
    % --- 任务描述 ---
    \begin{tcolorbox}[
        colback=white, colframe=black!60, boxrule=1pt, arc=2mm,
        title=\textbf{Task: Image-to-SVG and Chart Information Recognition},
        coltitle=white, colbacktitle=black!70, fonttitle=\bfseries
    ]
        \small\textbf{Goal:} Correct the initial SVG using structural CRUD operations to maintain a traceable and modifiable reasoning state. And answer the question: what is the average total fueling cost excluding the Ford F150?
    \end{tcolorbox}

    \vspace{-0.15cm}

    % --- 推理轨迹表格 ---
    \renewcommand{\arraystretch}{1.5}
    \begin{tabularx}{\textwidth}{|p{1.8cm}|X|}
        \hline
        \rowcolor{gray!20} \textbf{Role} & \textbf{Trajectory Step} \\ \hline

        % --- Round 1: Insert ---
        \textbf{Solver} & 
        \small\textbf{[Round 1: Externalizing Analysis via \texttt{insert\_element}]} \\
        & \small\textit{Input: System Prompt + User Prompt + Initial SVG Code + Original Image (Fig.~\ref{fig:logcase_a}) + Image Rendered by Initial SVG (Fig.~\ref{fig:logcase_b})} \\
        & \small\texttt{<think>} The user wants me to correct an SVG image to match an original image. First, I will compare the original image with the rendered SVG.
        \begin{itemize}[leftmargin=1.5em, nosep]
            \item \textcolor{red}{\textbf{Layout and Sizing:}} Elements extend beyond the \texttt{viewBox} of \texttt{0 0 300 600}. For example, the third bar at $x=280$ with $width=87$ goes up to $x=367$, which is outside the 300px width.
            \item \textcolor{red}{\textbf{Proportions:}} Relative heights are approximately correct, but absolute positions, widths, and spacing do not match the compact and centered layout of the original image.
            \item \textcolor{red}{\textbf{Color:}} The current fill is \texttt{\#b030b0} (purple), but the original image uses a more vibrant magenta \texttt{\#a629a6}.
            \item \textcolor{red}{\textbf{Shape:}} The bars in the original image have slightly rounded corners, whereas the current SVG rects are sharp-cornered.
            \item \textcolor{red}{\textbf{Text Positioning:}} Costs and labels are misplaced due to the underlying layout errors.
        \end{itemize}
        My plan is to recalculate the layout, refine visuals with rounded corners and correct colors, and replace the \texttt{main\_svg}. I will start by noting down the identified issues on the notebook. \texttt{</think>} \\
        & \small\texttt{\textbf{\textcolor{blue}{<tool\_call>}}\{"name":"insert\_element", "arguments":\{"rootId":"root", "fragment":"<div id='analysis'>... (issues 1-5) ...</div>"\}\}\textbf{\textcolor{blue}{</tool\_call>}}} \\ \hline
        
        % --- Critique 1 ---
        \textbf{Critique} & 
        \small\textit{Current State: figure~\ref{fig:logcase_c}} \\ 
        &\small\texttt{<tool\_response>} [Action Result] The second image is the result of \texttt{tool\_call insert\_element}. 
        ...(Critique Prompt) \texttt{</tool\_response>} \\ \hline

        % --- Round 2: Replace ---
        \textbf{Solver} & 
        \small\textbf{[Round 2: Update via \texttt{replace\_element}]} \\
        & \small\texttt{<think>} \textit{I have already calculated the new layout parameters: ViewBox \texttt{0 0 300 600}, Bar Width \texttt{70px}, Spacing \texttt{25px}, and Horizontal Margin \texttt{20px}. Bar X-coordinates are set to \texttt{20}, \texttt{115}, \texttt{210}, with proportional heights (\texttt{450px}, \texttt{284px}, \texttt{243px}) bottom-aligned at $y=470$. I will now use \texttt{replace\_element} to update the \texttt{main\_svg} on the notebook.} \texttt{</think>} \\
        & \small\texttt{\textbf{\textcolor{blue}{<tool\_call>}}\{"name":"replace\_element", "arguments":\{"targetId":"main\_svg", "fragment":"}
        \newline
        \small\texttt{<svg id="main\_svg" viewBox="0 0 300 600">}
        \newline
        \small\texttt{~~<rect id="bar1" fill="\#a629a6" \textcolor{red}{height="450" width="70" x="20" y="20"} rx="4" ry="4"/>}
        \newline
        \small\texttt{~~<text id="cost1" ...>\$130.96</text>... </svg>"\}\}\textbf{\textcolor{blue}{</tool\_call>}}} \\ \hline

        % --- Critique 2 ---
        \textbf{Critique} & 
        \small\textit{Current State: figure~\ref{fig:logcase_d}} \\
        & \small\texttt{<tool\_response>} [Action Result] The second image is the result of \texttt{tool\_call replace\_element}. ...(Critique Prompt) \texttt{</tool\_response>} \\ \hline

        % --- Round 3: Remove ---
        \textbf{Solver} & 
        \small\textbf{[Round 3: Remove Redundant Content via \texttt{remove\_element}]} \\
        & \small\texttt{<think>} The correction is verified. The SVG on the notebook now accurately represents the original image. I will now remove the analysis node. \texttt{</think>} \\
        & \small\texttt{\textbf{\textcolor{blue}{<tool\_call>}}\{"name":"remove\_element", "arguments":\{"targetId":"analysis"\}\}}
        \small\texttt{\textbf{\textcolor{blue}{</tool\_call>}}} \\ \hline

         % --- Critique 3 ---
        \textbf{Critique} & 
        \small\textit{Final State: figure~\ref{fig:logcase_e}} \\
        & \small\texttt{<tool\_response>} [Action Result] The second image is the result of \texttt{tool\_call remove\_element}. ...(Critique Prompt) \texttt{</tool\_response>} \\ \hline

        % --- Final Result ---
        \textbf{Final Ans.} & 
        \small\texttt{\textbf{\textcolor{asparagus}{<answer>}}}
        \small\texttt{<svg id="main\_svg" viewBox="0 0 300 600" ...>}
        \newline
        \small\texttt{~~<rect id="bar1" fill="\#a629a6" \textcolor{red}{height="450" width="70" x="20" y="20"} rx="4" ry="4"></rect>}
        % \newline
        \small\texttt{~~...(Final SVG Code)}
        % \newline
        \small\texttt{</svg>}
        \small\texttt{\textbf{\textcolor{asparagus}{</answer>}}} \\ \hline

    \end{tabularx}
    \caption{\textbf{CRUD-Action Trajectory of Canvas-CoT.}}
    \label{fig:log}
\end{figure}

\subsection{RBench-V Cases}
\label{sec:rbenchcase}
\begin{figure}[htbp]
    \centering
    
    % --- 1.任务描述 ---
    \begin{tcolorbox}[
        colback=white,
        colframe=black!60,
        boxrule=1pt,
        left=3pt, right=3pt, top=3pt, bottom=3pt,
        arc=2mm,
        title=\textbf{Task: Geometry Calculation},
        coltitle=white,
        colbacktitle=black!70,
        fonttitle=\bfseries
    ]
        \textbf{Question:} In triangle $\triangle ABC$, the bisector of angle $\angle B$ and the bisector of the exterior angle at vertex $C$ intersect at point $D$. Given that $\angle A = 27^\circ$, what is the measure of angle $\angle BDC$?
    \end{tcolorbox}

    %\vspace{0.2cm}

    % --- 2. 推理轨迹表格 ---
    \renewcommand{\arraystretch}{1.5}
    \begin{tabularx}{\textwidth}{|p{2cm}|X|}
        \hline
        \rowcolor{header_gray} \textbf{Role} & \textbf{Trajectory Step} \\ \hline

        % --- Step 1: System prompt ---
        \textbf{System} & 
        \textit{System prompt}
        \\ \hline
        % --- Step 2: User prompt ---
        \textbf{User} & 
        \textit{Question}
        \\ \hline

        % --- Step 3: Solver Initial ---
        \textbf{Solver} & 
        \small\textbf{[Round 1]} Here is a step-by-step derivation to find the measure of angle $\angle BDC$. I'll visualize the geometric setup described in the problem.
        
        \vspace{0.2em}
        \textbf{[Action]}
        
        \small\texttt{<tool\_call>\{"name":"insert\_element", "arguments":\{"rootId":"root", "fragment":"<div id="step1"...<strong style="font-size: 17px;">Problem Setup</strong>...</div>","beforeId":null \} \}</tool\_call> }
        
        \small\texttt{... ...}
        
        \small\texttt{<tool\_call>\{"name":"insert\_element", "arguments":...\}</tool\_call> }

        \vspace{0.2em}
        \small\textbf{[Note]}{The agent internally generates a diagram where the extension line incorrectly starts from vertex A.}
        \\ \hline

       % --- Step 4: Critique ---
        \textbf{Critique} & 
        \begin{tcolorbox}[
            colback=critique_bg,
            colframe=critique_frame,
            boxrule=0.8pt,
            left=2pt, right=2pt, top=2pt, bottom=2pt,
            arc=2mm,
            title=\textbf{\small [WARNING] CRITIQUE TRIGGERED: Visual-Textual Inconsistency},
            coltitle=critique_frame,
            colbacktitle=critique_bg,
            fonttitle=\bfseries\small
        ]
            \small
            \textbf{Hallucination Detect:} \texttt{True} 
            
            \textbf{Issue Identified:}
            \begin{itemize}[leftmargin=0.7em, itemsep=0.01em]
                \small\item \textbf{Original Question (Text Constraint):} "The problem is solved by considering the exterior angle at vertex C formed by extending the side BC. The textual part of the solution correctly states this: \textit{'Let's consider the exterior angle formed by extending side BC to a point Q.'}"
                \small\item \textbf{Notebook (Visual State):} "The diagram provided in the 'Problem Setup' section visually depicts the exterior angle at C as being formed by extending the side AC, indicated by the dashed line extending from the segment AC."
            \end{itemize}
            
            \small\textbf{Explanation:} The diagram contradicts the method described in the text and used for the solution. The mathematical derivation correctly follows the premise of extending side BC, but the accompanying visual aid incorrectly shows the extension of side AC.
        \end{tcolorbox}
        \\ \hline

        % --- Step 5: Solver Correction & Comparison ---
        \textbf{Solver} & 
        \small\textbf{[Self-Correction]} "The user has pointed out a discrepancy... The text correctly describes the exterior angle... but the diagram incorrectly shows a dashed line originating from vertex A."
        \newline
        \small\textbf{[Action]} \texttt{replace\_element(targetId="step1", ...)}
        
        % --- 图片 ---
        \begin{tcolorbox}[
            colback=correction_bg, 
            colframe=black!10, 
            boxrule=0.4pt, 
            left=2pt, right=2pt, top=2pt, bottom=2pt,
            arc=2mm,
            before=\vspace{-0.2mm},
            after=\vspace{-0.2mm}
            ]
            \centering
            \vspace{-0.4em}
            %\textbf{\small Visual Correction Execution:}\\
            %\vspace{0.2em}
            
            \newcommand{\imgheight}{3.5cm} % 图片高度控制
            
            \begin{minipage}[b]{0.43\linewidth}
                \centering
                % 左图：错误状态
                \includegraphics[height=\imgheight, keepaspectratio]{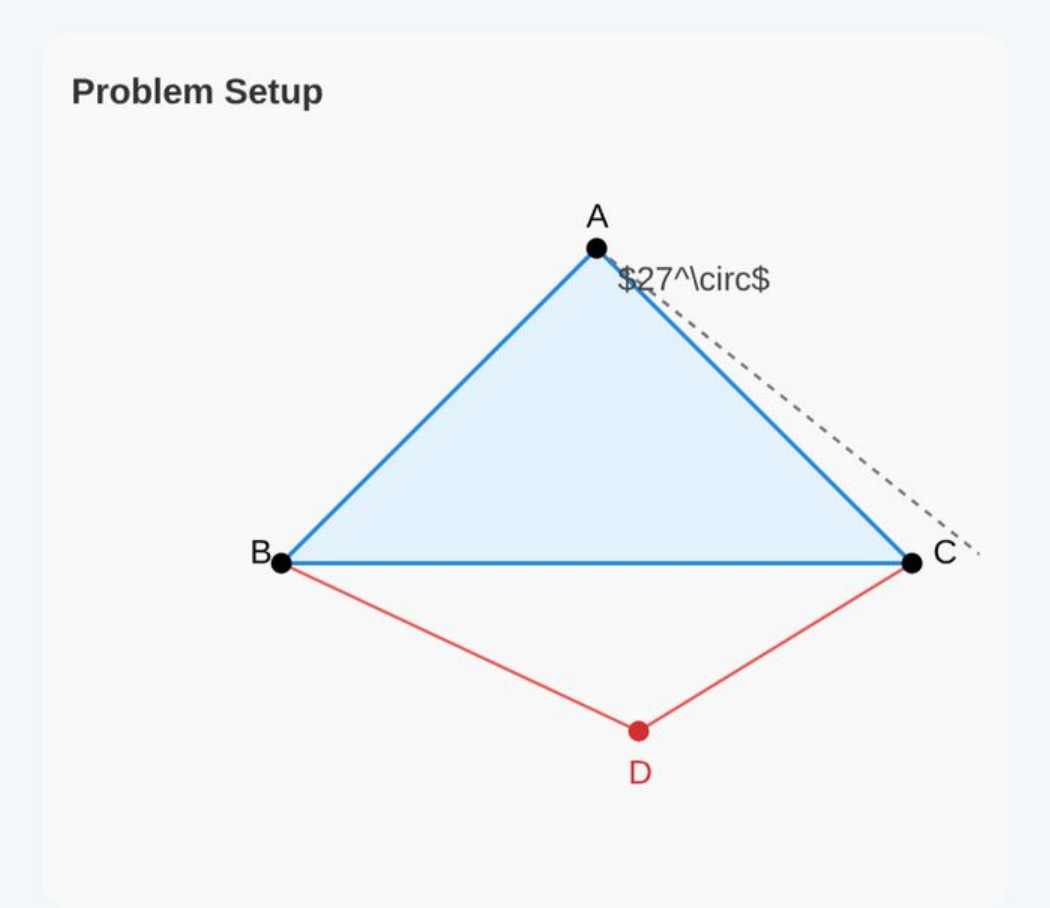}
                \captionof*{figure}{\footnotesize \textcolor{red}
                {\small\textbf{(a)Before Critique}} Incorrect Extension from A}
            \end{minipage}
            \hfill
            \begin{minipage}[b]{0.43\linewidth}
                \centering
                % 右图：正确状态
                \includegraphics[height=\imgheight, keepaspectratio]{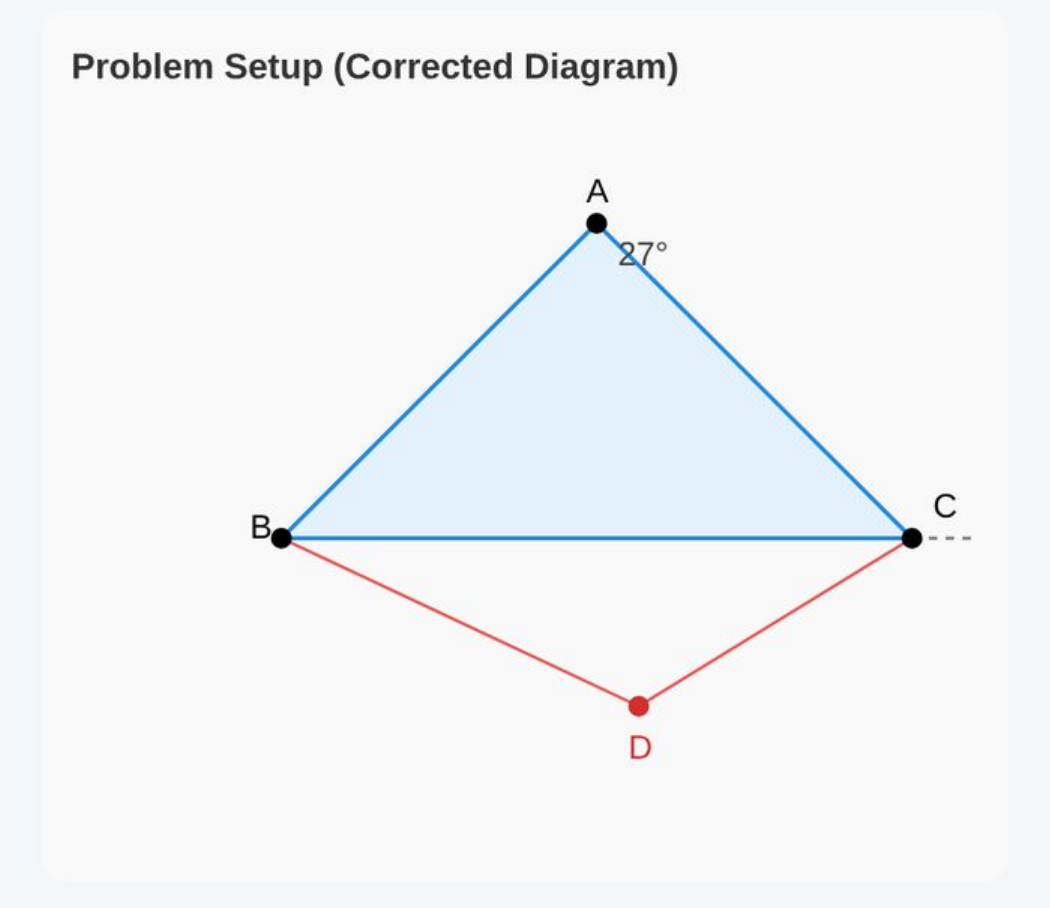}
                \captionof*{figure}{\footnotesize \textcolor{olive}{\small\textbf{(b)After Correction}} Correct Extension from B}
            \end{minipage}
        \end{tcolorbox}
        \\ \hline

        % --- Step 4: Final Derivation ---
        \textbf{Solver} & 
        \small\textbf{[Final Derivation]} 
        With the corrected setup:
        Angle sum of $\triangle BDC = 180^\circ$.
        Derived $\angle BDC = \frac{\angle A}{2}$.
        \newline
        \small\textbf{Calculation:} $\angle BDC = \frac{27^\circ}{2} = 13.5^\circ$.
        \newline
        \small\textbf{Final Output:} \texttt{<answer>}\boxed{13.5^\circ}\texttt{</answer>}
        \\ \hline

    \end{tabularx}

    \caption{\textbf{Critique-Driven Correction Process.} The Canvas-CoT agent initially hallucinates the geometric construction (extending the wrong side). The Critique Agent identifies the mismatch between the correct textual logic and the incorrect visual rendering. The agent then executes an atomic update to fix the canvas (shown in the green box) before deriving the final answer. There are also other hallucinations in the figure that need to be corrected. For brevity, only one critical critique step is shown in the final answer derivation process.}
    \label{fig:critique_process}
\end{figure}

\begin{figure}[htbp]
    \centering
    
    % --- 1. 任务描述 ---
    \begin{tcolorbox}[
        colback=white,
        colframe=black!60,
        boxrule=1pt,
        arc=2mm,
        top=2pt, bottom=2pt,
        title=\textbf{Task: Visual Reasoning \& Geometry Counting},
        coltitle=white,
        colbacktitle=black!70,
        fonttitle=\bfseries
    ]
        \begin{minipage}[c]{0.68\linewidth}
            \textbf{Question:} \\
            In the figure shown on the right, there are a total of (\quad) triangles. \\
            Solve the question according to the given picture step-by-step.
        \end{minipage}
        \hfill
        \begin{minipage}[c]{0.28\linewidth}
            \centering
            \includegraphics[height=2.0cm, width=\linewidth, keepaspectratio]{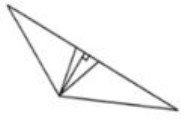}
            \vspace{-0.5em}
            \captionof*{figure}{\scriptsize \textbf{Original Input Image}}
        \end{minipage}
        
    \end{tcolorbox}

    \vspace{0.2cm}

    % --- 2. 推理轨迹表格 ---
    \renewcommand{\arraystretch}{1.5}
    \begin{tabularx}{\textwidth}{|p{2cm}|X|}
        \hline
        \rowcolor{gray!20} \textbf{Role} & \textbf{Trajectory Step} \\ \hline

        % --- Step 1: System prompt ---
        \textbf{System} & 
        \textit{System prompt}
        \\ \hline
        % --- Step 2: User prompt ---
        \textbf{User} & 
        \textit{Question + Original Input Image}
        \\ \hline

        % --- Step 3: Solver Round 1 (Hallucination) ---
        \textbf{Solver} & 
        \small\textbf{[Round 1]} I will analyze the triangle structure and count them.
        
        \vspace{0.5em}
        \textbf{[Action]}
        
        \small\texttt{<tool\_call>\{"name":"insert\_element", "arguments":\{"fragment":"<svg id='fig1'>...<line x2='140'/><line x2='260'/>...</svg>"\}\}</tool\_call>}
        
        \vspace{0.5em}
        \small\textbf{[Note]}{The agent generates a diagram with only \textbf{2} internal lines instead of 3, leading to an incorrect total count of 6.}
        \\ \hline

        % --- Step 4: Critique 1 ---
        \textbf{Critique} & 
        \begin{tcolorbox}[
            colback=critique_bg,
            colframe=critique_frame,
            boxrule=0.8pt,
            left=4pt, right=4pt, top=4pt, bottom=4pt,
            arc=2mm,
            title=\textbf{\small [WARNING] CRITIQUE: Fundamental Visual Error},
            coltitle=critique_frame,
            colbacktitle=critique_bg,
            fonttitle=\bfseries\small
        ]
            \small
            \textbf{Hallucination Detect:} \texttt{True} 
            
            \textbf{Issue Identified:} 
            The original image has \textbf{3 internal lines}, dividing the base into 5 points. The notebook claim shows only \textbf{2 lines}. This misrepresentation makes the subsequent count (6) incorrect. The correct count should be $C(5,2)=10$.
        \end{tcolorbox}
        \\ \hline

        % --- Step 5: Solver Round 2 (Structural Correction) ---
        \textbf{Solver} & 
        \small\textbf{[Self-Correction]} Clearing the canvas to redraw the correct structure with 5 points on the base (B, C, D, E, F).
        \newline
        \small\textbf{[Action]} \texttt{clear()}; \texttt{insert\_element(fragment="<svg id='fig\_correct'>...</svg>")}

        \begin{tcolorbox}[
            colback=correction_bg, 
            colframe=black!10, 
            boxrule=0.4pt, 
            arc=2mm,
            top=2pt, bottom=2pt
        ]
            \centering
            \begin{minipage}[b]{0.45\linewidth}
                \centering
                \includegraphics[height=3.3cm, keepaspectratio]{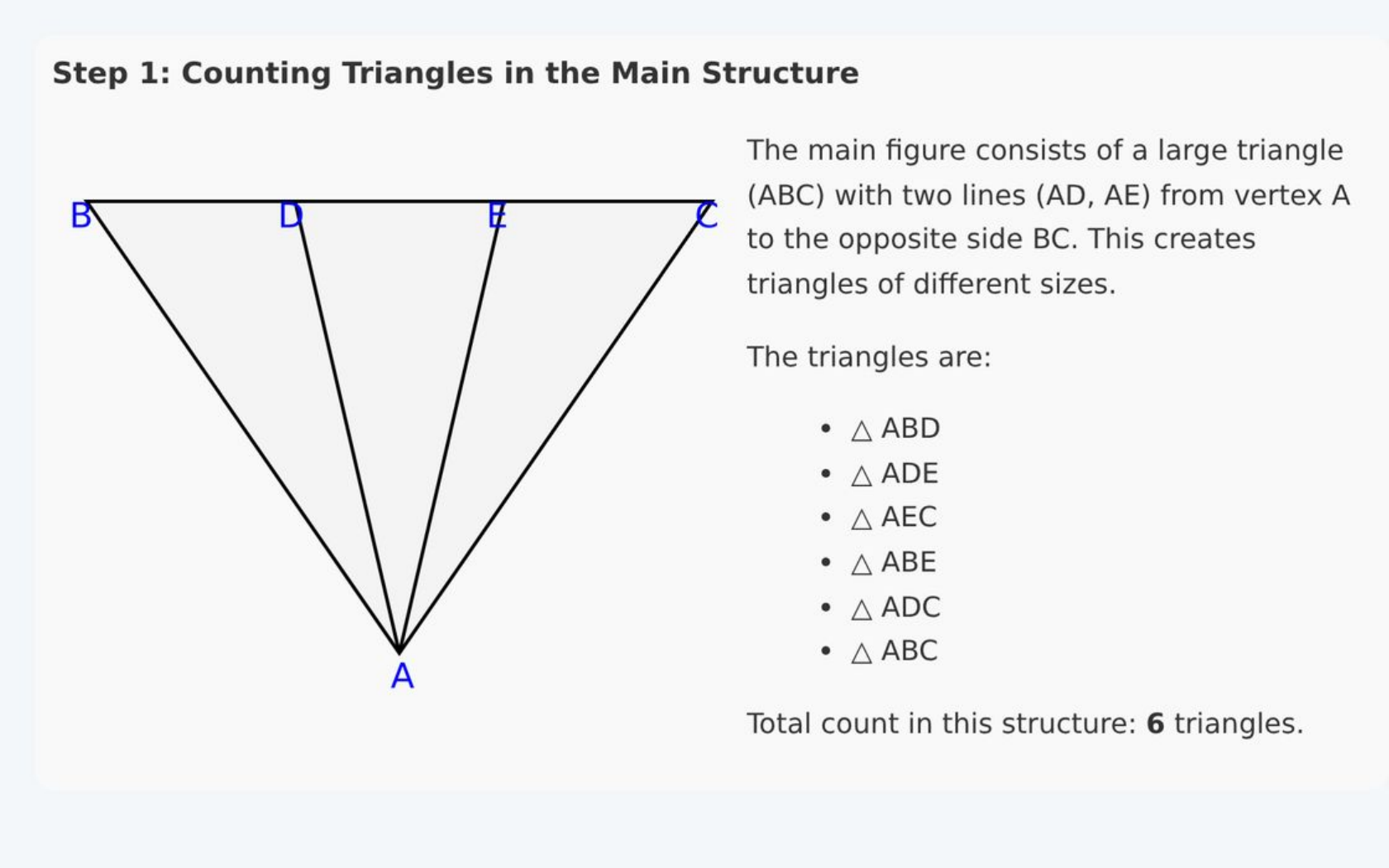}
                \captionof*{figure}{\scriptsize \textcolor{red}{\textbf{(a) Initial Hallucination}}}
            \end{minipage}
            \hfill
            \begin{minipage}[b]{0.45\linewidth}
                \centering
                \includegraphics[height=3.3cm, keepaspectratio]{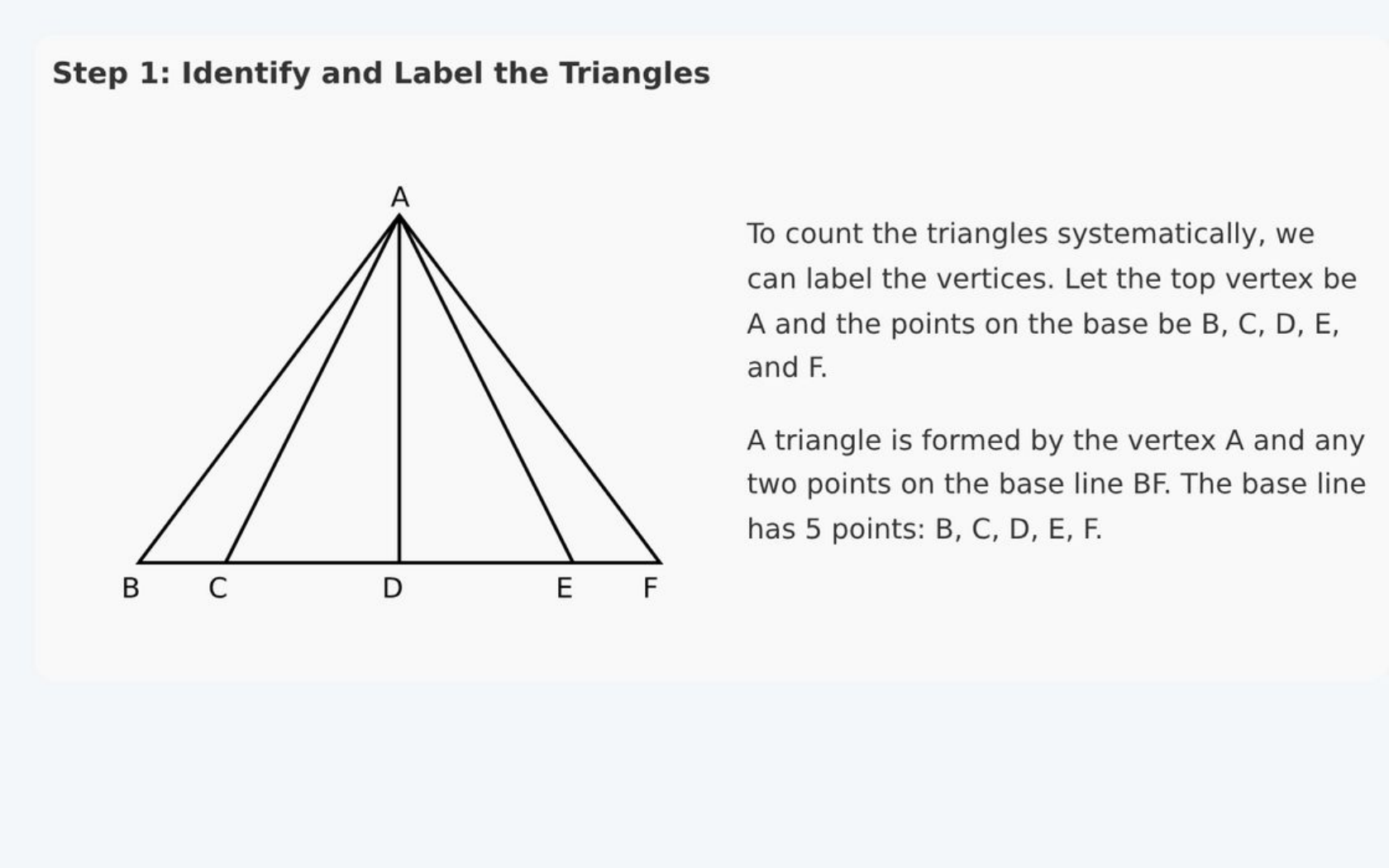}
                \captionof*{figure}{\scriptsize \textcolor{olive}{\textbf{(b) Corrected Canvas}}}
            \end{minipage}
        \end{tcolorbox}
        \\ \hline

        % --- Step 6: Critique 2 (Geometric Fidelity) ---
        \textbf{Critique} & 
        \begin{tcolorbox}[
            colback=orange!5,
            colframe=orange!60,
            boxrule=0.8pt,
            left=4pt, right=4pt, top=4pt, bottom=4pt,
            arc=2mm,
            title=\textbf{\small [WARNING] CRITIQUE: Geometric Inconsistency},
            coltitle=orange!80!black,
            colbacktitle=orange!5,
            fonttitle=\bfseries\small
        ]
            \small
            \textbf{Issue:} The notebook draws a symmetric triangle, but the original image shows a skewed triangle with an \textbf{altitude} marked by a right-angle symbol. 
            
            \textbf{Refinement:} While the count is unaffected, the visual fidelity should be improved to match the input.
        \end{tcolorbox}
        \\ \hline

        % --- Step 7: Final Execution ---
        \textbf{Solver} & 
        \small\textbf{[Final Logic]} Any two points on the base connected to the apex A form a triangle.
        
        \small\textbf{Combinatorial Calculation:}$C(5, 2) = \frac{5 \times 4}{2 \times 1} = 10$
        
        \small\textbf{Final Output:} \boxed{10}
        \\ \hline

    \end{tabularx}

    \caption{\textbf{Multi-Step Critique and Correction in Visual Reasoning.} The figure illustrates how the agent initially hallucinates a simplified version of the problem. Through iterative Critique, the agent updates its virtual notebook to reflect the correct geometric constraints and applies combinatorial logic to reach the accurate result.}
    \label{fig:geometry_counting_process}
\end{figure}

In this section, we select two representative cases in ~Figure\ref{fig:critique_process} and ~Figure\ref{fig:geometry_counting_process} to demonstrate the canvas reasoning process and the corresponding CRUD behavior trajectories of the model.

%%%%%%%%%%RBench case1%%%%%%%%%

%%%%%%%%%%%%%%%%%%%%%%%rbench case2%%%%%%%%%%%

\clearpage
\section{Prompts}
\label{sec:prompts}
% \subsection{VCode Prompts}
% \label{sec:vcodeprompts}
This section presents the specific prompts used for the VCode benchmark, organized in the order of system prompt, critique prompt, and user prompt. Here, the critique prompt is simplified to be directly applied at the end of each round to stimulate reflection.
% \label{sec:traj}

% As shown in Figure~\ref{fig:log}, to illustrate the tool execution flow of Canvas-CoT, we present the "Bar Chart Correction" task. The process consists of three distinct turns:

%  The agent first identifies layout and color errors in the initial chart. Instead of fixing them blindly, it uses \texttt{insert\_element} to write an analysis note directly onto the canvas. This action turns internal reasoning into a visible object, ensuring the model keeps track of the specific errors it needs to fix.

% Based on the analysis, the agent calculates the correct layout parameters. It then executes a single \texttt{replace\_element} operation to update the target SVG node. This demonstrates how the model performs an atomic update, instantly correcting the visual state without the need to regenerate the entire history.

% Once the corrected chart is verified against the ground truth, the agent uses \texttt{remove\_element} to delete the temporary analysis note. This step cleans up the canvas, ensuring the final output remains focused and free of auxiliary working notes.

% The visualization process is illustrated in Figure~\ref{fig:logcase}.

\begin{promptbox}
\small
\textbf{\# Objective \#}

You are a \textbf{Visual-Reasoning Agent}, solving image-to-svg problems by synchronizing a visual Chain-of-Thought on a virtual notebook. The primary goal is \textbf{100\% accuracy}.

\vspace{0.8em}
\hrule
\vspace{0.8em}

% itemsep=3pt: 项目之间留 3pt (约 1mm)
% parsep=0pt:  段落之间不留额外空隙
% topsep=3pt:  列表和上方文本空隙 

\textbf{\# Process \#}

\vspace{0.5em}
\textbf{\# Step 1: Think One Step}
\begin{itemize}[itemsep=3pt, parsep=0pt, topsep=3pt]
    \item \textbf{Action}: Thinking one more step for answering the question.
    \item \textbf{Output}: Enclose the entire thinking process within \texttt{<think>} tags.
    \item \textbf{Rule}: Do not give answer directly. (First of all, describe the image precisely.)
    \begin{itemize}[itemsep=3pt, parsep=0pt, topsep=3pt]
        \item Remember that each step should contain only a small part of the reasoning process, avoiding outputting long paragraphs of reasoning at once. For example: analyze A in one step, analyze B in one step, analyze C in one step, set up equations in one step, and perform calculations in one step.
        \item Strictly avoid delivering a lengthy explanation before presenting the notes.
    \end{itemize}
\end{itemize}

\vspace{0.5em}
\textbf{\# Step 2: Tool Call}
\begin{itemize}[itemsep=3pt, parsep=0pt, topsep=3pt]
    \item \textbf{Trigger}: Immediately after a \texttt{<think>} block is complete.
    \item \textbf{Action}: Call the appropriate \textbf{Notebook Tool} to visually record the key \textbf{evidence, data points, or intermediate results} from your thinking step. This synchronizes the internal thought process with the external visual memory.
    \item \textbf{Output}: Enclose the tool function call within \texttt{<tool\_call>} tags.
    \item \textbf{Rule}: Updates should be incremental. \textbf{Instead of only showing a final answer (e.g., '3'), first visualize the components that lead to it.} For example, if you identify three items, use \texttt{insert\_element} to list those items on the notebook \textit{before} presenting the final count.
\end{itemize}

\vspace{0.5em}
\textbf{\# Step 3: Iterate}
\begin{itemize}[itemsep=3pt, parsep=0pt, topsep=3pt]
    \item \textbf{Trigger}: Current evidence is insufficient for a \textbf{100\% accuracy} answer.
    \item \textbf{Action}: Identify missing information, reformulate queries, and Repeat \textbf{Step 1 (Think)} and \textbf{Step 2 (Tool Call)}.
    \item \textbf{Process}: Each loop should build upon the previous state on the notebook, progressively moving towards the solution.
\end{itemize}

\vspace{0.5em}
\textbf{\# Step 4: Final Answer}
\begin{itemize}[itemsep=3pt, parsep=0pt, topsep=3pt]
    \item \textbf{Trigger}: Current evidence is sufficient for a \textbf{100\% accuracy} answer.
    \item \textbf{Action}: State the final, correct answer clearly and concisely.
    \item \textbf{Output}: Enclose the final answer within \texttt{<answer>} tags.
\end{itemize}

\vspace{0.8em}
\hrule
\vspace{0.8em}

\textbf{\# Notebook \& Tools \#}

The notebook is an HTML container (\textbf{Width: 800px}, Height: Auto). You have 5 tools to manipulate it.

\vspace{0.5em}
\texttt{<tools> \{provided\_tools\} </tools>}

\vspace{0.5em}
For each function call, return a json object with function name and arguments within \texttt{<tool\_call></tool\_call>} XML tags:

\vspace{0.5em}
\texttt{<tool\_call> \{"name": <function-name>, "arguments": <args-json-object>\} </tool\_call>}

\vspace{0.5em}
\textbf{\# Notebook Tool Usage Guidelines}
\begin{itemize}[itemsep=3pt, parsep=0pt, topsep=3pt]
    \item \texttt{insert\_element}
    \begin{itemize}
        \item \textbf{You must assign a unique \texttt{id} attribute when creating an element, to facilitate future modifications, replacements, or deletions (e.g., \texttt{<rect id="r1" ...>}). Use short and unambiguous IDs.}
        \item Using SVG is recommended \textbf{only if} they require subsequent editing and have a \textbf{simple structure}. A simple structure is defined as:
        \begin{itemize}[itemsep=3pt, parsep=0pt, topsep=3pt]
            \item The diagram consists of basic shapes (e.g., rectangles, circles, triangles, lines) and is not a complex figure like a floor plan with text and auxiliary lines, or a solid geometry diagram involving spatial relationships.
            \item It is a table with a clear row and column structure where the cell content is \textbf{text-only}.
        \end{itemize}
        \item One Example: 
        
        \texttt{\{"name": "insert\_element", "arguments": \{"rootId": "root", "beforeId": null, "fragment": "<svg id='sg1' width='500' height='350' xmlns='http://www.w3.org/2000/svg'>...some SVG objects...</svg>"\}}
    \end{itemize}
    
    \item \texttt{modify\_element}
    \begin{itemize}
        \item One Example:
        
        \texttt{\{"name": "modify\_element", "arguments": \{"targetId": "r1", "attrs": \{"fill": "\#009E5F", "stroke": "black", "stroke-width": "2"\}\}\}}
    \end{itemize}

    \item \texttt{remove\_element}
    \begin{itemize}
        \item One Example: 
                
        \texttt{\{"name": "remove\_element", "arguments": \{"targetId": "r1"\}\}}
    \end{itemize}

    \item \texttt{replace\_element}
    \begin{itemize}
        \item One Example: 
                
        \texttt{\{"name": "replace\_element", "arguments": \{"targetId": "lbl", "fragment": "<text id='lbl' x='15' y='60' fill='\#1377EB'>new label for the rectangle</text>"\}}
    \end{itemize}

    \item \texttt{clear}
    \begin{itemize}
        \item This will clear all content on the notebook.
        \item One Example: \texttt{\{"name": "clear", "arguments": \{\}\}}
    \end{itemize}
\end{itemize}

\vspace{0.8em}
\hrule
\vspace{0.8em}

\textbf{\# Core Principles \#}
\begin{enumerate}[itemsep=3pt, parsep=0pt, topsep=3pt]
    \item \textbf{Visualize to Solve}: Do not rely solely on textual reasoning. Your thought process is incomplete until its key components are externalized on the notebook. Geometric, spatial, or logical relationships \textbf{must} be drawn. \textbf{Key findings, identified objects, and intermediate calculations must be recorded.}
    \item \textbf{Visual Validation}: Continuously use the visual state on the notebook to check for contradictions and verify intermediate steps.
    \item \textbf{Precision}: All coordinates and values used in tool calls must precisely match the calculations in your reasoning steps.
    \item \textbf{No content overlap}: When adding any content to the notebook, ensure careful layout to avoid overlapping or covering existing content unless they belong to the same object logically. If needed, remove or adjust old content before adding new ones.
    \item \textbf{Traceable Visual Reasoning}: The notebook is a \textbf{visual workspace for drafting and verifying your understanding}. Use it to sketch out object placements, spatial relationships, and key features as you identify them. This visual draft helps you verify your reasoning against the original image before finalizing the SVG code.
    \item \textbf{Modify Existing Elements}: Do NOT blindly append new elements. Your task is to optimize the initial SVG to precisely copy the original image. Use \texttt{modify\_element} to adjust attributes (color, position, size) or \texttt{replace\_element} to fix shapes or elements. If you identify elements in the original image that are missing from the current SVG, you could use tools (e.g., \texttt{insert\_element}) to add them. Similarly, if you find elements in the SVG that are NOT in the original image, you could use \texttt{remove\_element} to delete them.
    \begin{itemize}
        \item \textbf{IMPORTANT}: The initial SVG has \texttt{id="main\_svg"}. To add elements \textit{inside} it, use \texttt{insert\_element} with \texttt{rootId="main\_svg"}. If you omit \texttt{rootId}, the element will be placed \textit{outside} the SVG, which is usually WRONG.
    \end{itemize}
    \item \textbf{Spatial Precision}: Pay attention to the \textbf{absolute positions} and \textbf{relative alignment} of objects. If an object is misplaced, move it.
    \item The initial SVG may contain errors or inaccuracies, if you think an element is wrong, you can use \texttt{remove\_element} to delete it on the notebook.
\end{enumerate}

\vspace{0.8em}
\hrule
\vspace{0.8em}

\textbf{\# Notebook Operation Restrictions \#}

\textbf{\# Overall Layout \& Width Limitations}
\begin{itemize}[itemsep=3pt, parsep=0pt, topsep=3pt]
    \item The notebook area has a fixed width of \texttt{500px}; all internal elements must not exceed \texttt{500px} in width.
    \item Content block styles:
    \begin{itemize}
        \item \textbf{Background color}: Avoid using background colors whenever possible. If necessary, use light backgrounds to highlight specific parts. Avoid nesting multiple content blocks.
        \item \textbf{Padding}: Keep appropriate padding around text and elements; if a block has a background color, ensure at least \textasciitilde14px side padding and 10px top/bottom padding.
        \item \textbf{Corner radius}: Default corner radius for content block cards is 12px.
    \end{itemize}
    \item Typography rules:
    \begin{itemize}
        \item \textbf{Paragraphs}: Avoid using the \texttt{border} property, except for SVG graphics.
        \item \textbf{Lists}: Do not add left/right margins to \texttt{UL} or \texttt{LI} tags.
        \item Avoid using \texttt{<p>} tags. Avoid borders and shadows. Avoid using background colors for large content areas.
        \item \textbf{Corner radius}: Default is 12px for content block cards.
        \item \textbf{Spacing}: Vertical spacing between content blocks is 12px; padding is 10px top/bottom and 14px left/right.
    \end{itemize}
    \item Font rules:
    \begin{itemize}
        \item Notebook Text: Do not specify custom fonts. Use 18px bold (main title), 17px bold (subtitle), 16px (default body text), 14px (notes).
        \item SVG Content: The font restrictions above do NOT apply. You could adjust them in the SVG to match the original image if necessary.
        \item Pay attention to the width of elements in the SVG to ensure they do not exceed the canvas boundaries.
    \end{itemize}
    \item No overlapping content: \textbf{All content must fit within the notebook area, with no overlap or covering of existing elements.}
\end{itemize}

\textbf{\# Visual Semantics \#}

Use colors to represent logical states consistently.
\begin{itemize}[itemsep=3pt, parsep=0pt, topsep=3pt]
    \item \textbf{Black/Grey (\#000, \#666)}: Static constraints, given information, or background elements.
    \item \textcolor{RoyalBlue}{\textbf{Blue (\#1377EB)}}: The current element of focus, a hypothesis being tested, or an active calculation.
    \item \textcolor{Teal}{\textbf{Green (\#009E5F)}}: A verified truth, a correct intermediate step, or the final result.
    \item \textcolor{Red}{\textbf{Red (\#ED2633)}}: A detected contradiction, an error in reasoning, or an important warning.
    \item \textbf{Dashed Lines}: Auxiliary lines, imaginary paths, or construction guides.
\end{itemize}

\end{promptbox}

%%%%% Critique prompt v6%%%%
\begin{critiquepromptbox}
\small
\textbf{\texttt{<tool\_response>}}

\vspace{0.5em}
\textbf{[Action Result]} The first image is the original question. The second image is the result of tool\_call \{\}.

\vspace{0.8em}
\hrule
\vspace{0.8em}

\textbf{[Critique Task]} At the beginning of this new turn, your first step is to perform a critical review.

\begin{enumerate}[itemsep=3pt, parsep=0pt, topsep=3pt]
    \item Bring your intermediate analysis back to the original question for a reverse check.
    \item Verify the tool output image is consistent with the original image and your notes.
    \item If anything doesn't match, correct the notebook state and reasoning before finalizing.
\end{enumerate}

\vspace{0.5em}
Compare the original image (first) with the SVG-rendered image (second) and identify \textbf{SPECIFIC} differences for SVG code revision.

\vspace{0.5em}
Focus on identifying:

\begin{enumerate}[itemsep=3pt, parsep=0pt, topsep=3pt]
    \item \textbf{LOCATION-SPECIFIC DIFFERENCES:}
    \begin{itemize}[itemsep=3pt, parsep=0pt, topsep=3pt]
        \item Which areas/regions differ (top-left, center, bottom-right, etc.)
        \item Missing or extra elements in specific positions
    \end{itemize}

    \item \textbf{VISUAL ATTRIBUTE DIFFERENCES:}
    \begin{itemize}[itemsep=3pt, parsep=0pt, topsep=3pt]
        \item Color mismatches (specify which elements and what colors)
        \item Shape distortions (which shapes are wrong and how)
        \item Size/proportion issues (which elements are too big/small)
        \item Position/alignment problems
    \end{itemize}

    \item \textbf{SPECIFIC SVG REVISION SUGGESTIONS:}
    \begin{itemize}[itemsep=3pt, parsep=0pt, topsep=3pt]
        \item Which SVG elements need modification (circles, paths, rects, etc.)
        \item What attributes to change (fill, stroke, cx, cy, width, height, d, etc.)
        \item Specific color values or coordinate adjustments needed
    \end{itemize}
\end{enumerate}

\vspace{0.5em}
Format your response as actionable SVG revision instructions.

\textbf{\texttt{</tool\_response>}}
\end{critiquepromptbox}

%%%% USER PROMPT VCode %%%%
\begin{userpromptbox}
\small
You are an \textbf{SVG code specialist.}

\vspace{0.5em}
\textbf{Task:} Compare the \textbf{ORIGINAL} image (first) with the \textbf{CURRENT SVG RENDER} (second). Then modify the SVG code to better match the original.

\vspace{0.8em}
\hrule
\vspace{0.8em}

\textbf{**CRITICAL INSTRUCTION ON VISUAL FIDELITY**:}
\begin{itemize}[itemsep=3pt, parsep=0pt, topsep=3pt, leftmargin=*]
    \item Focus on the global \textbf{SIMILARITY} of the entire image.
    \item If the original image contains \textbf{bounding boxes, selection rectangles, or highlights}, treat them as \textbf{part of the visual content} to be drawn.
    \item Do NOT interpret a bounding box as a command to 'zoom in' or 'focus only on this area'.
    \item Do NOT ignore the box itself. If there is a red box in the original, your SVG must have a red box in the same location.
    \item Your goal is to make the SVG look \textbf{exactly like the original image}, including any annotations or markers present in the original.
    \item \textbf{Occluded Objects}: If objects in the original image are occluded or partially visible, reproduce them EXACTLY as they appear (i.e., partially visible).
\end{itemize}

\vspace{0.5em}
\textbf{Process:}
\begin{enumerate}[itemsep=3pt, parsep=0pt, topsep=3pt, leftmargin=*]
    \item \textbf{Analyze \& Understand}: First, use \texttt{<think>...</think>} to analyze the visual differences (missing/redundant elements, wrong shapes, incorrect colors, and relative spatial positions/alignment), and understand what each visual element represents in the original image.
    \item \textbf{Plan for Fidelity}: Decide how to represent these better use SVG.
    \item \textbf{Execute Changes}: Use the blackboard tools to MODIFY the initialized SVG on canvas, including but not limited to visually draft the corrected shapes, verify spatial constraints, verify item deletions or additions on initialized canvas. Don't write reasoning text on the canvas.
    \item \textbf{Iterative Refinement}: Perform multi-round iteration to correct the SVG image, verifying each change visually.
    \item Finally, output the COMPLETE revised SVG code wrapped in \texttt{<answer>...</answer>} tags.
\end{enumerate}

\vspace{0.5em}
\textbf{Rules:}
\begin{itemize}[itemsep=3pt, parsep=0pt, topsep=3pt, leftmargin=*]
    \item The SVG inside \texttt{<answer>} MUST start with \texttt{<svg} and end with \texttt{</svg>}.
    \item Do NOT change the canvas/viewBox/overall size unless necessary for correctness.
    \item \textbf{Modify the content on the CURRENT canvas. Do NOT create a new SVG element. Modify the existing one.}
    \item \textbf{When using \texttt{insert\_element}, ALWAYS use \texttt{rootId='main\_svg'} to insert inside the existing SVG.}
    \item If you think the image rendered by initial svg is worthless or completely wrong, you should use 'clear' to clear the canvas and generate a new svg for the whole original image to replace it, rather than call too many tools to revise the wrong one.
\end{itemize}

\vspace{0.8em}
\hrule
\vspace{0.8em}

\textbf{CURRENT SVG CODE:}

\vspace{0.5em}

\texttt{\{current\_svg\_code\}}

\vspace{1em}
Now, start your reasoning and revision:

\end{userpromptbox}

\end{document}